\RequirePackage{fix-cm}
\documentclass[twocolumn]{svjour3}          
\smartqed  
\usepackage[table]{xcolor}
\usepackage{bm}
\usepackage{bbm}
\usepackage{graphicx}
\usepackage[authoryear]{natbib}
\usepackage[lined,boxed,commentsnumbered,ruled]{algorithm2e}
\usepackage{tabu,multirow}
\usepackage[leftcaption]{sidecap}
\usepackage{amsmath}
\usepackage{amssymb}
\usepackage{color}
\usepackage{xcolor}
\definecolor{mygreen}{RGB}{34,139,34}
\usepackage{enumitem}
\usepackage{verbatim}
\usepackage{pythonhighlight}
\usepackage[colorlinks,citecolor=blue]{hyperref}

\usepackage{algorithmic}

\usepackage{graphicx}
\usepackage{subfigure}
\usepackage{float}  
\usepackage{textcomp}

\usepackage{booktabs}

  \usepackage{eso-pic}
  \usepackage{xspace}
  \makeatletter
  \DeclareRobustCommand\onedot{\futurelet\@let@token\@onedot}
  \def\@onedot{\ifx\@let@token.\else.\null\fi\xspace}

  \makeatother
  
  \usepackage{mathtools}

\begin{document}
  \sloppy
  
  \title{Self-Supervised Video Representation Learning in a Heuristic Decoupled Perspective}

  \author{ Zeen~Song \and
           Wenwen~Qiang \and
           Changwen~Zheng \and
           Hui~Xiong \and
           Gang~Hua
  }
  
 \institute{Zeen Song, Wenwen Qiang, Changwen Zheng are with
              National Key Laboratory of Space Integrated Information System, Institute of Software Chinese Academy of Sciences; University of Chinese Academy of Sciences, Beijing, China. \{songzeen, qiangwenwen, changwen\}@iscas.ac.cn.\\
           \and
           Hui Xiong is with Thrust of Artificial Intelligence, Hong Kong University of Science and Technology, Guangzhou, China; Department of Computer Science and Engineering, the Hong Kong University of Science and Technology, Hong Kong SAR, China. E-mail: xionghui@ust.hk.\\
           \and
           Gang Hua is with the Multimodal Experiences Lab, Dolby Laboratories Inc, Los Angeles, CA, USA; Institute of Artificial Intelligence and Robotics, Xi’an Jiaotong University, Xi’an 710049, P.R.China. E-mail: ganghua@gmail.com.\\
           \and
           Corresponding author: Wenwen Qiang.
}

\date{Received: date / Accepted: date}

\def\ourconv{RIConv++\xspace}
\def\smallgap{\vspace{0.05in}}
  
  \maketitle
  \begin{abstract}  
  Video contrastive learning (V-CL) has emerged as a popular framework for unsupervised video representation learning, demonstrating strong results in tasks such as action classification and detection. Yet, to harness these benefits, it is critical for the learned representations to fully capture both static and dynamic semantics. However, our experiments show that existing V-CL methods fail to effectively learn either type of feature. Through a rigorous theoretical analysis based on the Structural Causal Model and gradient update, we find that in a given dataset, certain static semantics consistently co-occur with specific dynamic semantics. 
  This phenomenon creates spurious correlations between static and dynamic semantics in the dataset. However, existing V-CL methods do not differentiate static and dynamic similarities when computing sample similarity. As a result, learning only one type of semantics is sufficient for the model to minimize the contrastive loss. Ultimately, this causes the V-CL pre-training process to prioritize learning the easier-to-learn semantics. To address this limitation, we propose \textit{Bi-level Optimization with Decoupling for Video Contrastive Learning.} (BOD-VCL).
  In BOD-VCL, we model videos as linear dynamical systems based on Koopman theory. In this system, all frame-to-frame transitions are represented by a linear Koopman operator. By performing eigen-decomposition on this operator, we can separate time-variant and time-invariant components of semantics, which allows us to explicitly separate the static and dynamic semantics in the video. 
  By modeling static and dynamic similarity separately, both types of semantics can be fully exploited during the V-CL training process.
  BOD-VCL can be seamlessly integrated into existing V-CL frameworks, and experimental results highlight the significant improvements achieved by our method. The source code is released at~\href{https://github.com/ZeenSong/Video_contrastive}{https://github.com/ZeenSong/Video\_contrastive}.


\end{abstract}

\section{Introduction}
Extracting useful feature representations from video data without label information has long been a challenging task \cite{feichtenhofer_large-scale_2021,schiappa_self-supervised_2022,kuang_video_2021,qian_spatiotemporal_2021}. Inspired by the success of self-supervised learning (SSL) in image and text processing \cite{chen_simple_2020, grill_bootstrap_2020, caron_emerging_2021, he_masked_2022}, SSL for video (V-SSL) is proposed to solve this challenge and has achieved significant success \cite{benaim2020speednet,feichtenhofer_large-scale_2021,feichtenhofer_masked_2022,qian_spatiotemporal_2021,tong_videomae_2022,wang_videomae_2023}. A typical method of V-SSL is video contrastive learning (V-CL). Its main idea lies in how to constrain the similarity between augmented samples. In simple terms, in the feature space, the constraint ensures that augmented samples derived from the same ancestor sample are clustered together, while augmented samples derived from different ancestor samples are pushed apart. V-CL has also been applied to various downstream tasks, including classification, detection, tracking, and others \cite{soomro2012ucf101, gu2018ava, dendorfer2021motchallenge}.

\begin{figure*}[htb]
    \centering
    \subfigure[]{%
        \includegraphics[width=0.45\textwidth, height=0.315\textwidth]{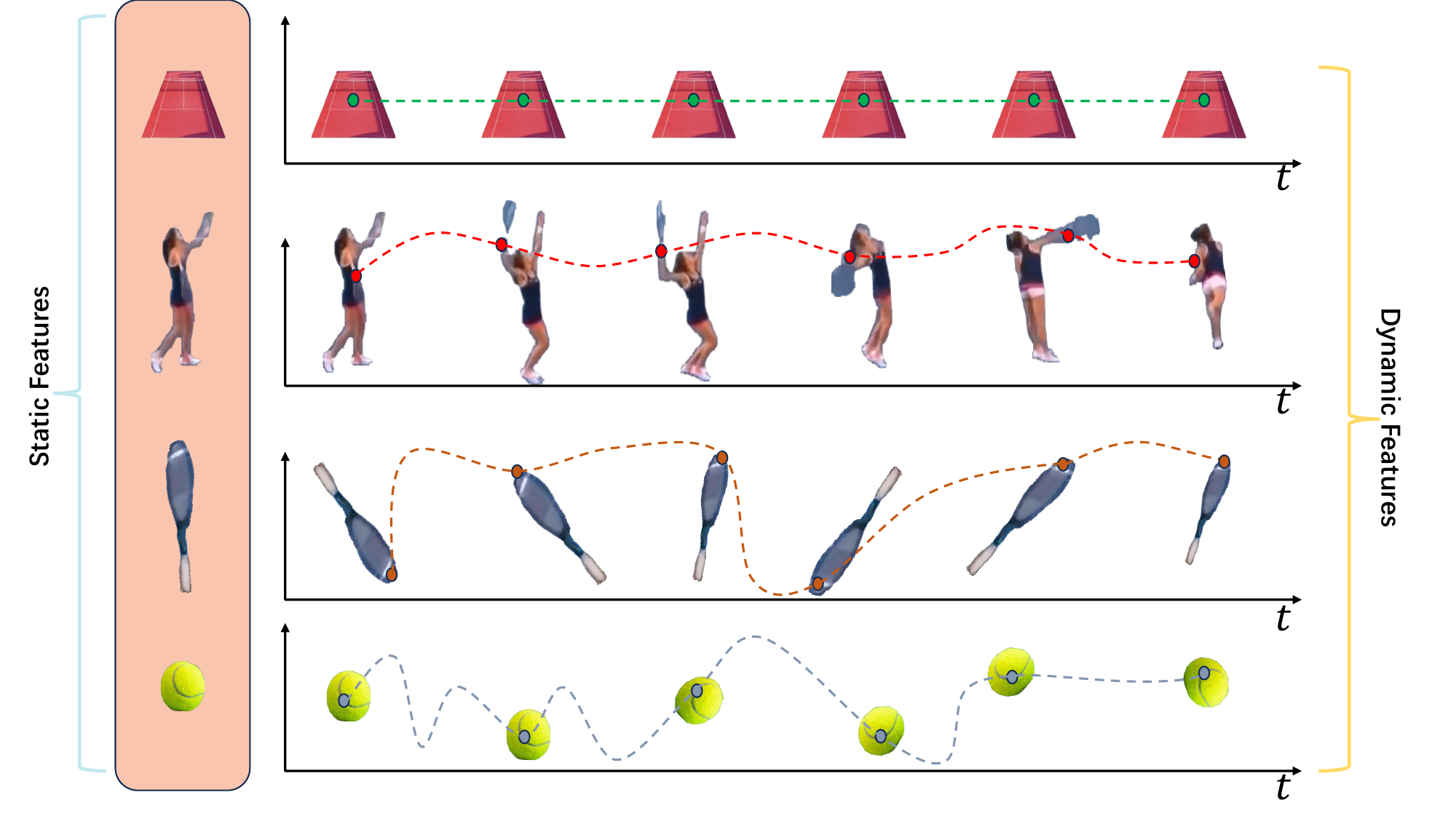}
        \label{fig:dynamic feature}
    }\hfill
    \subfigure[]{%
        \includegraphics[width=0.5\textwidth]{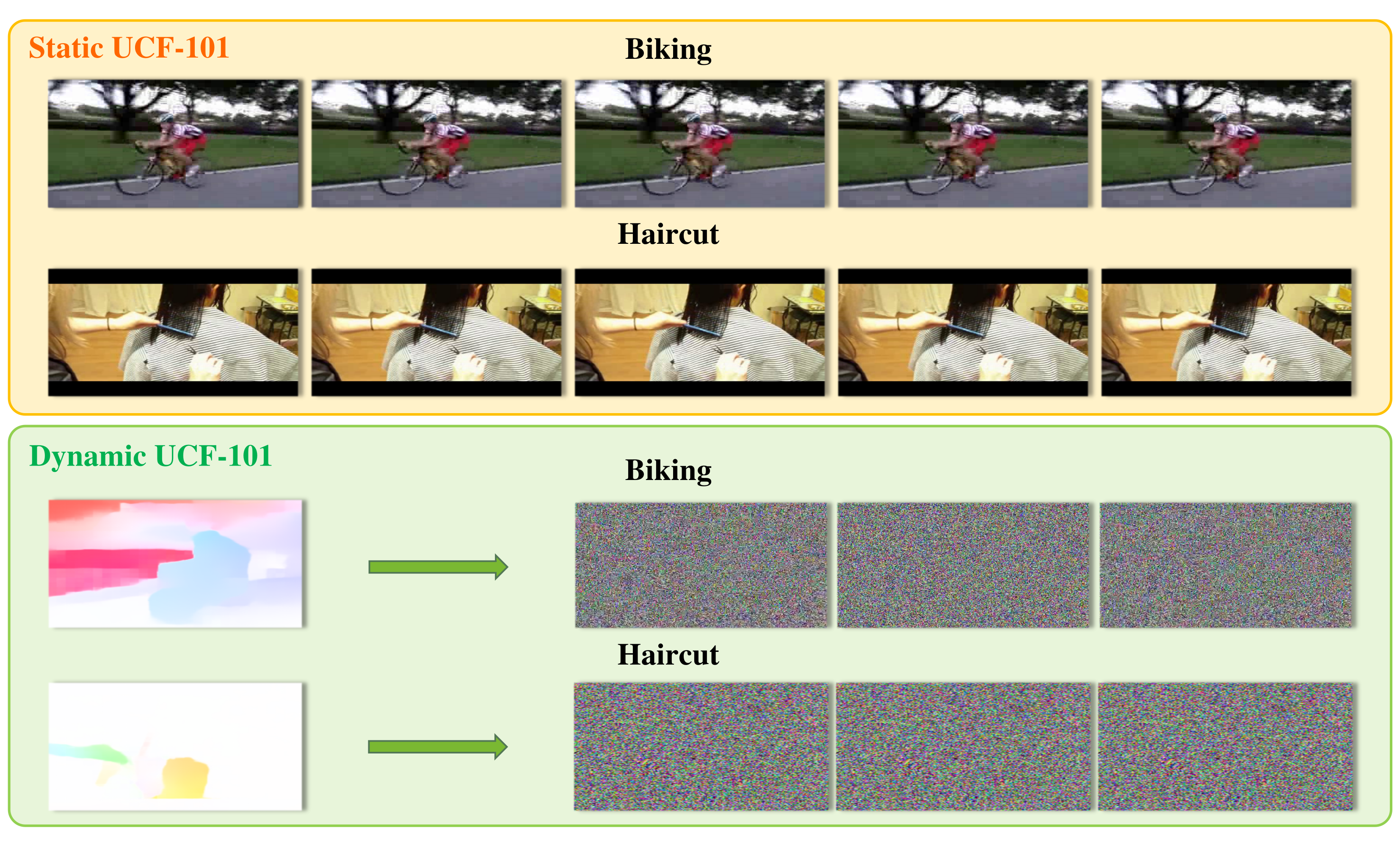}
        \label{fig:moti_exp}
    }
    \caption{(a) Visualization of static and dynamic semantics from a video capturing a tennis match. {(b) Illustration of the motivation experiments in Static UCF-101 dataset and Dynamic UCF-101 dataset.}}
\end{figure*}

Despite its broad adoption, capturing both static and dynamic semantics remains crucial for V-CL to achieve stronger performance across diverse downstream tasks \cite{severe, liRESOUNDActionRecognition2018, sarkar_uncovering_2023}. {In video data, static semantics refer to elements that remain consistent across frames, such as object appearances and scene contexts, while dynamic semantics capture temporal variations, including motion patterns and actions, as illustrated in Figure \ref{fig:dynamic feature}.} Because the modeling in V-CL is driven solely by similarity constraints, it is unclear whether this learning paradigm enables the model to encode all static and dynamic semantics. Consequently, it remains an open question \textbf{whether existing V-CL methods effectively learn representations that incorporate both semantic aspects}.

{To systematically investigate whether existing V-CL methods effectively capture both static and dynamic semantics, we design two controlled evaluation experiments using the UCF-101 \cite{soomro2012ucf101} dataset. For evaluating static semantics, we ensure identical dynamic information across all samples by duplicating a single frame from each video into a full-length sequence. For evaluating dynamic semantics, we follow Ilic et al. \cite{ilic2022appearance} to remove static information by replacing video content with noise frames warped according to the optical flows of the original videos. }
We evaluate four widely used V-CL methods \cite{feichtenhofer_large-scale_2021} and compare them with a supervised pre-training baseline, all pre-trained on Kinetics-400 \cite{kay2017kinetics} for $200$ epochs, following the linear evaluation protocol introduced in Section \ref{subsec:cl}. Results in Table \ref{tab:motivation} show that supervised pre-training consistently outperforms V-CL methods in both experiments. 
{Specifically, V-CL methods achieve, on average, $10.10\%$ lower accuracy in capturing static semantics and $8.78\%$ lower accuracy in capturing dynamic semantics compared to supervised methods. }
These observations demonstrate that current V-CL methods are less effective in representing both static and dynamic features, which may negatively impact their performance on real-world downstream tasks.

To better understand the causes behind the experimental observations described above, we conduct a theoretical analysis from a causal and gradient update perspective. Specifically, we propose a Structural Causal Model (SCM) in Figure \ref{fig:scm}. In this SCM, \(\mathbf X \) represents the video sample, \(\mathbf A \) represents the static semantics, \(\mathbf M \) represents the dynamic semantics, and \(\mathbf S \) denotes the similarity score. The causal path \(\mathbf{A} \gets \mathbf{X} \to \mathbf{M}\) indicates that both static and dynamic semantics are influenced by the video sample \(\mathbf{X}\), making \(\mathbf{X}\) a confounder and inducing a correlation between \(\mathbf{A}\) and \(\mathbf{M}\). In Proposition \ref{prop:1}, we prove that due to these correlations, gradient updates tend to favor the semantic component that is easier to optimize, leaving the harder-to-learn component insufficiently updated. Consequently, this biased learning leads to suboptimal representation quality for both static and dynamic semantics. 
In Corollary \ref{coro:1}, we further show that this issue can be mitigated by explicitly modeling static and dynamic similarities in a decoupled manner.

To effectively capture both static and dynamic semantics in a decoupled manner, we propose a novel modeling approach named \textit{Bi-level Optimization with Decoupling for Video Contrastive Learning} (BOD-VCL). Our method consists of three main components. First, we propose to estimate a linear Koopman operator specific to each video, capturing how semantic representations evolve across frames. Second, we perform an eigen-decomposition on this operator to explicitly separate static semantics from dynamic semantics within the video representations. Third, we employ a decoupled contrastive loss that independently models the similarities of these separated semantic components. Furthermore, we formulate the learning process as a bi-level optimization procedure: in the first level, we optimize a prediction loss to accurately estimate the Koopman operator; in the second level, we update the feature extractor parameters using the decoupled contrastive loss, given a reliable Koopman operator. Through iterative updates of these two optimization steps, the feature extractor progressively enhances its ability to represent video semantics, while the Koopman operator concurrently improves in distinguishing between static and dynamic semantics, producing more expressive video representations.

We empirically evaluate the effectiveness of our approach by evaluating the downstream performance on action recognition benchmark datasets, including Kinetics-400 \cite{kay2017kinetics}, UCF-101 \cite{soomro2012ucf101}, and HMDB-51 \cite{kuehne2011hmdb}. We also evaluate our results on motion-aware datasets, i.e. Something-Something v2 \cite{ssv2}, Diving48 \cite{liRESOUNDActionRecognition2018} and FineGym \cite{gym99}. Besides, we provide a case study through visualization to demonstrate the correct learning of both static and dynamic features. Through extensive experiments on downstream tasks, we demonstrate that our proposed BOD-VCL provides significant improvements. Our contributions are as follows.
\begin{itemize}
    \item Through empirical analysis, we conduct controlled experiments to evaluate the capability of pre-trained models using current V-CL methods in capturing static and dynamic semantics. From the results, we find that they are ineffective in capturing either static or dynamic semantics.
    \item We conduct a theoretical analysis to uncover the underlying causes of these experimental findings from both causal and gradient-update perspectives. Our analysis reveals that confounding effects inherent in pre-training datasets induce correlations between static and dynamic semantics. Consequently, unified similarity measures in existing V-CL methods bias models toward easier-to-learn semantics, hindering the learning of more challenging ones. We further prove that explicitly decoupling the modeling of static and dynamic similarities enables effective learning of all semantic components regardless of their learning difficulty.
    \item Motivated by our theoretical findings, we propose a novel framework, \textit{Bi-level Optimization with Decoupling for Video Contrastive Learning} (BOD-VCL), to explicitly separate and independently learn static and dynamic semantics. BOD-VCL estimates a Koopman operator to capture inter-frame evolution, applies eigen-decomposition to isolate static and dynamic components, and uses a decoupled contrastive loss to learn representation separately. The entire process is formulated as a bi-level optimization procedure, where the Koopman operator is first optimized via prediction loss and then used to guide the learning of semantic representations.
    \item Experimental results demonstrate that BOD-VCL outperforms existing V-CL methods on the benchmark as well as various motion-heavy datasets. Furthermore, ablation studies and visualization results demonstrate the effectiveness of our method in accurately capturing both static and dynamic semantics.
\end{itemize}

\section{Related Works}
\paragraph{Self-Supervised Video Representation Learning.} 
Self-supervised learning, a learning paradigm that does not require labeled data, has achieved significant success in fields such as image and text processing \cite{chen_simple_2020, grill_bootstrap_2020, zbontar_barlow_2021, devlin2018bert}. In this context, a series of works have sought to extend the success of SSL to video representation learning \cite{benaim2020speednet,feichtenhofer_large-scale_2021,feichtenhofer_masked_2022,qian_spatiotemporal_2021,tong_videomae_2022,wang_videomae_2023}. These studies construct learning objectives by designing pretext tasks that leverage the video data itself, enabling the extraction of useful features from video data without the need for labeled information \cite{schiappa_self-supervised_2022}. These features can then be applied to downstream tasks such as action recognition and action detection \cite{kay2017kinetics,gu2018ava}. V-SSL can be categorized based on the nature of its pretext tasks into several types: context-based methods, contrastive learning-based methods, and mask modeling-based methods. 

Context-based video self-supervised methods leverage inherent properties of videos, such as playback speed \cite{benaim2020speednet}, temporal order \cite{fernando2017self}, video jigsaw puzzles \cite{ahsan2019video}, and optical flow \cite{wang2019self}, to construct a series of supervisory signals. These methods aim to learn generalizable features while solving these tasks.

Contrastive-based methods of V-SSL apply spatial or temporal data augmentations to the video samples to create positive and negative sample pairs \cite{feichtenhofer_large-scale_2021,qian_spatiotemporal_2021, ranasinghe_self-supervised_2022, khorasgani_slic_2022}. The learning object is to constrain the representation of positive pairs to be more similar while ensuring the representations of negative pairs are less similar in the latent space. Different approaches to create positive and negative pairs have been proposed. Specifically, a major line of works treats the representation of different clips sampled from the same video as positive pairs while clips sampled from different videos as negative pairs \cite{feichtenhofer_large-scale_2021, qian_spatiotemporal_2021, videomoco}. TCLR \cite{dave_tclr_2022} additionally treats the overlapping clips from the same video as positive pairs. MaCLR \cite{avidan_maclr_2022} and CoCLR \cite{han_self-supervised_2021} additionally treat the representation of the optical flow and RGB video of the same video clip as positive pairs. Tubelet Contrast \cite{thoker_tubelet-contrastive_2023} additionally constructs positive samples by simulating tubelet trajectories across different videos, thereby addressing the issue of fewer action instances in pre-trained data. 

Mask modeling-based methods decompose videos into spatiotemporal blocks and employ a Transformer-based autoencoder \cite{he_masked_2022, feichtenhofer_masked_2022, tong_videomae_2022, wang_videomae_2023}. By masking a portion of the spatiotemporal blocks, the model learns to reconstruct the masked sections using the unmasked parts and the masked blocks' spatiotemporal positions. 

These existing V-SSL methods have achieved comparative results on benchmark datasets such as Kinetics-400, UCF-101, and HMDB-51. Among them, contrastive-based V-SSL has gained widespread attention in both academia and industry due to its relatively low computational cost and the absence of the need for fine-tuning. Therefore, this paper primarily focuses on V-CL methods. We also include comparisons with other types of V-SSL methods in the experimental section.

\paragraph{The Practical Limitation of V-CL.}
Although V-CL has achieved promising results on benchmark datasets, several studies have suggested that these datasets suffer from a static bias problem \cite{liRESOUNDActionRecognition2018, severe, sarkar_uncovering_2023}. This issue arises when models are able to achieve high performance by merely learning static features such as backgrounds, objects, and people, without effectively capturing dynamic features \cite{liRESOUNDActionRecognition2018, liMitigatingEvaluatingStatic2023a, qianStaticDynamicConcepts2022, wangRemovingBackgroundAdding2021}. Due to the shortcut learning effect in deep learning, models tend to focus on these static features, neglecting the extraction of temporal dynamics \cite{geirhos2020shortcut}. As a result, the accuracy reported on these datasets may not accurately reflect the true performance of the representations learned through video self-supervised learning \cite{severe}. To address this, the SEVERE benchmark evaluates existing datasets across five dimensions: label overlap, temporal awareness, point of view (PoV), action length, and environment. Additionally, Li et al. \cite{liRESOUNDActionRecognition2018} propose Diving48, a dataset that contains similar static semantics for all videos with varying diving postures. 

Some prior work has sought to explicitly constrain models to learn dynamic features in video self-supervised learning through various approaches. Notably, several studies have redefined positive and negative samples within the framework of video contrastive learning, encouraging models to prioritize similarity between sample pairs that exhibit specific dynamic characteristics. For instance, CoCLR \cite{han_self-supervised_2021} and MaCLR \cite{avidan_maclr_2022} treat RGB frames and optical flows of the same video clip as positive sample pairs. Seco \cite{yao2021seco} and DSM \cite{wang_enhancing_2020} both use correctly ordered video clips as positive pairs and incorrectly ordered clips as negative pairs.

However, the issue of bias is not necessarily contain only static bias. Another kind of bias is motion bias \cite{schuldt2004recognizing,bregonzio2009recognising}, which refers to the bias present in the dynamic features of datasets. For example, in action recognition datasets, if all videos of a specific activity, such as kicking a ball, are filmed from a parallel perspective, the motion of the person may be minimal while the background changes rapidly. In contrast, if a new video of the same activity is filmed from an overhead view, relying solely on dynamic features for recognition could lead to incorrect predictions. This issue is briefly mentioned in \cite{schuldt2004recognizing,bregonzio2009recognising}, where they illustrated that short-term motion biases could suffice for action recognition, but existing research has not thoroughly investigated it from a more fundamental perspective.

In causal inference, these biases are introduced by the confounding effect \cite{pearl2009causality, glymour_causal_2016}. In the field of deep representation learning, a series of works leveraging causal inference has been proposed \cite{yangDeconfoundedImageCaptioning2020, yueInterventionalFewShotLearning2020, qiangInterventionalContrastiveLearning2022}. In this paper, we analyze the learning of V-CL methods from a causal perspective and analyze how the confounding effect influences the V-CL learning process. 

\paragraph{Background of Koopman Theory.}
In this paper, we propose incorporating Koopman theory into video representation learning. The Koopman theory maps observable state into a high-dimensional space, transforming nonlinear dynamical systems into linear ones \cite{luenberger1979dynamic, koopman1931hamiltonian}. 

Recently, Dynamic Mode Decomposition (DMD) \cite{schmid2010dynamic} has become a standard algorithm for approximating the Koopman operator in a data-driven manner. DMD has been widely applied to various fields, including fluid dynamics \cite{schmid2010dynamic}, neuroscience \cite{brunton2016extracting}, and video processing \cite{erichson_compressed_2019}. Additionally, approximating the Koopman operator using deep neural networks has garnered increased attention in recent years \cite{takeishi2017learning,lusch2018deep,berman2022multifactor}.

In this paper, we explicitly model the evolution of frames inside videos by approximating a Koopman operator and utilize its eigenvalue decomposition to separate static and dynamic features.

\section{Problem Analysis}
In this section, we first provide a brief overview of common approaches in V-CL. We then conduct a series of experiments to investigate the ability of V-CL methods to capture both static and dynamic semantics. Finally, we provide a theoretical analysis of the observed results from a causal perspective.

\subsection{The Learning Framework of V-CL}
\label{subsec:cl}
V-CL leverages large-scale, unlabeled video datasets to pre-train a spatiotemporal feature extractor $f(\cdot)$, in a self-supervised manner. The unlabeled video dataset is denoted as \(\mathcal{D}=\{X_i\}_{i=1}^{N}\), where each \(X_i\) represents a video. Each video \(X_i\) consists of a sequence of frames, \(X_i=\{x_{i,t}\}_{t=1}^{T_i}\), where every frame \(x_{i,t}\in\mathbb{R}^{H \times W \times C}\) has a height of \(H\), a width of \(W\), and \(C\) channels. Here, \(T_i\) denotes the total number of frames in the video. Self-supervision in V-CL is achieved by generating different views from the videos via data augmentation. These data augmentation strategies typically encompass both temporal and spatial augmentations. {In particular, temporal augmentations randomly sample a subset of consecutive frames from the original video, with a specified subset length \(L\) and a stride \(\delta\). Meanwhile, spatial augmentations consistently apply transformations such as cropping, color distortion, and random horizontal flipping \cite{tian_what_2020} to each frame in the sampled subset.} In the following, we provide a detailed overview of how supervisory signals are generated in several classical V-CL approaches.

V-SimCLR \cite{feichtenhofer_large-scale_2021, chen_simple_2020} is one of the most fundamental approaches in V-CL, generating supervisory signals by constructing positive and negative sample pairs through data augmentation. Specifically, a mini-batch \(\mathcal{B} = \{X_i\}_{i=1}^{N_{bs}}\) of video data is sampled, where \(N_{bs}\) represents the number of videos in the batch. For each video \(X_i\) in the batch, \(\varrho\geq 2\) independent data augmentations are applied, yielding \(\varrho\) augmented samples denoted by \(\{\rho_{i,j}(X_i)\}_{j=1}^{\varrho}\), where \(\rho_{i,j}\) indicates the \(j\)-th augmentation of \(X_i\). These augmented samples are subsequently processed by a spatiotemporal feature extractor \(f\) and a projection head \(f_p\) to produce the embedding features \(Z_{i,j} = f_p(f(\rho_{i,j}(X_i)))\). The positive set for an embedding \(Z_{i,j}\) is defined as $\mathcal{Z}_{i,j}^{+} = \{ Z_{i,k} \mid k \in \{1, \dots, \varrho\} \setminus \{j\} \}$, which consists of the embeddings from other augmentations of the same video. Conversely, the negative set is defined as $\mathcal{Z}_i^{-} = \{ Z_{m,j} \mid m \in \{1, \dots, N_{bs}\} \setminus \{i\},\; j \in \{1, \dots, \varrho\} \}$, comprising the embeddings of all augmented samples from other videos in the mini-batch. 
For a given anchor $Z_{i,j}$, the objective is formulated as the InfoNCE loss:
\begin{equation}
\label{eq:infoNCE}
    \ell_{i,j} = -\log\frac{\sum_{Z^+\in\mathcal{Z}_{i,j}^+}\exp(\text{sim}(Z_{i,j},Z^+)/\alpha)}{\sum_{Z^\prime\in\{\mathcal{Z}_{i,j}^+,\mathcal{Z}_i^-\}}\exp(\text{sim}(Z_{i,j},Z^\prime)/\alpha)},
\end{equation}
where \(\alpha\) is a temperature hyper-parameter and \(\mathrm{sim}(u, v) = u^\top v / (\|u\| \|v\|)\) is cosine similarity. The overall loss for the mini-batch is computed as:
\begin{equation}
\label{eq:loss_batch}
    \mathcal{L} = \frac{1}{N_{bs}\cdot\varrho}\sum_{i=1}^{N_{bs}}\sum_{j=1}^{\varrho}\ell_{i,j}.
\end{equation}
Minimizing this loss encourages the feature extractor to produce embeddings that are more similar to those from the same video (positive pairs) and more dissimilar to those from different videos (negative pairs). In practice, the model parameters are updated via backpropagation and gradient descent, which iteratively refines both $f$ and $f_p$ so that they are capable of producing embeddings that more effectively distinguish between positive and negative samples.

Another commonly used V-CL method is V-MoCo \cite{feichtenhofer_large-scale_2021,he_momentum_2020}. Like V-SimCLR, it generates supervisory signals by constructing positive and negative sample pairs via data augmentation. However, instead of treating other augmented samples within the same mini-batch as negatives, V-MoCo maintains a memory bank that stores embeddings from previous iterations to serve as negative samples. Moreover, for any given anchor sample, its positive samples are generated by a target network \(f_{\theta_m}\) that shares the same architecture as \(f\) combined with \(f_p\). Specifically, for an anchor sample \(Z_{i,j}\), its positive set is defined as $\mathcal{Z}_{i,j}^{+} = \{ f_{\theta_m}(\rho_{i,k}(X_i)) \mid k \in \{1, \dots, \varrho\} \setminus \{j\} \}$. Its negative set, \(\mathcal{Z}^{-}\), is maintained as a fixed-length queue that is continuously updated with embeddings from augmented samples of previous iterations. The loss function in V-MoCo is identical to that of V-SimCLR as in Equation (\ref{eq:infoNCE}). The update rules for \(f\) and \(f_p\) are the same as in V-SimCLR, but the momentum encoder is updated according to:
\begin{equation}
\label{eq:ema}
    \theta_m \leftarrow \gamma \theta_m + (1-\gamma)\theta,
\end{equation}
where \(\gamma\) is a hyperparameter and \(\theta\) represents the parameters of \(f\) and \(f_p\).

Another distinct V-CL method is V-BYOL \cite{feichtenhofer_large-scale_2021, grill_bootstrap_2020}. Unlike V-MoCo and V-SimCLR, V-BYOL is a negative-free approach, meaning it only seeks to maximize the similarity between positive pairs without explicitly constraining the similarity between negative pairs. In addition to the feature extractor \(f\) and projection head \(f_p\), V-BYOL incorporates a target network \(f_{\theta_m}\) and further introduces an extra predictor \(f_{pre}\). In V-BYOL, the anchor sample is computed as $Z_{i,j} = f_{pre}\bigl(f_p(f(\rho_{i,j}(X_i)))\bigr)$, and its corresponding positive set is defined as $\mathcal{Z}_{i,j}^{+} = \{ f_{\theta_m}(\rho_{i,k}(X_i)) \mid k \in \{1, \dots, \varrho\} \setminus \{j\} \}$. For the anchor \(Z_{i,j}\), its loss function is:
\begin{equation}
\label{eq:byol}
    \ell_{i,j} = \sum_{Z^+\in\mathcal{Z}_{i,j}^+} \text{sim}(Z_{i,j}, Z^+),
\end{equation}
where $\text{sim}(\cdot)$ denotes the cosine similarity. The overall loss for the mini-batch can also be computed as Equation (\ref{eq:loss_batch}). The target network $f_{\theta_m}$ is optimized similarly to V-MoCo as Equation (\ref{eq:ema}).

By analyzing the Equation (\ref{eq:infoNCE})-(\ref{eq:byol}), it becomes evident that the core idea underlying these V-CL methods is to exploit the similarity relationships among augmented samples. Although various V-CL approaches exist \cite{dave_tclr_2022, qian_spatiotemporal_2021, videomoco, chen2021rspnet}, their fundamental objective remains consistent: to enable the feature extractor to learn from the similarity among augmented views. Consequently, when the pre-trained feature extractor is applied to downstream tasks (e.g., action recognition), it should generate distinct features for samples from different classes and similar features for samples within the same class. Next, we describe the evaluation protocols for V-CL methods.

To evaluate whether a model pre-trained via V-CL has learned effective video representations, the standard approach is linear-probe evaluation. In this procedure, the weights of the feature extractor \(f\) are frozen, and a linear classifier is trained on a labeled video dataset \(\mathcal{D}\)\footnote{This dataset does not need to be the same as the pre-training dataset. A common practice is to pre-train on Kinetics-400 and perform linear-probe evaluation on UCF-101 or HMDB51.}. The dataset is split into a training set and a test set: \(\mathcal{D}_{\text{train}} = \{(X_i, y_i)\}_{i=1}^{N_{\text{train}}}\) and \(\mathcal{D}_{\text{test}} = \{(X_i, y_i)\}_{i=1}^{N_{\text{test}}}\), where each video \(X_i\) is associated with a label \(y_i \in \{1, \dots, C\}\). With the pre-trained feature extractor fixed, the representation of a video is given by \(f(X_i)\). A linear classifier is learned based on these representations. And it is parameterized by weights \(W = [\boldsymbol{w}_1, \dots, \boldsymbol{w}_C] \in \mathbb{R}^{M \times C}\) and biases \(b = [b_1, \dots, b_C] \in \mathbb{R}^C\), where \(M\) is the dimension of the embedding produced by \(f\), and \(C\) is the number of classes. The classifier predicts the probability of class \(c\) for a video \(X_i\) as:  
\begin{equation}
\label{eq:downstream}
    p_{i, c} = \frac{\exp(\boldsymbol{w}_c^\top f(X_i) + b_c)}{\sum_{k=1}^C \exp(\boldsymbol{w}_k^\top f(X_i) + b_k)}, \quad \text{for } c = 1, \dots, C.
\end{equation}
The optimal \(W\) and \(b\) are obtained by minimizing the cross-entropy loss:  
\begin{equation}
\mathcal{L}(W, b) = -\frac{1}{N_{\text{train}}} \sum_{i=1}^{N_{\text{train}}} \log p_{i,y_i}.
\end{equation}
Once trained, the classifier's performance is evaluated on the test set using classification accuracy:  
\begin{equation}
\text{Acc} = \frac{1}{N_{\text{test}}} \sum_{i=1}^{N_{\text{test}}} \mathbf{1}\left[\arg\max_c p_{i, c} = y_i\right],
\end{equation}
where \(\mathbf{1}[\cdot]\) is the indicator function that returns $1$ if the predicted label \(\arg\max_c p_{i, c}\) matches the ground-truth label \(y_i\) and $0$ otherwise. A higher test accuracy indicates that the pre-trained feature extractor has successfully learned discriminative video representations that generalize well to unseen samples in downstream tasks.

\begin{table*}[htb]
    \centering
    \caption{{Top-1 and Top-5 linear evaluation accuracies of various methods on the Static and Dynamic UCF-101 dataset. "Supervised" indicates the model is trained with supervision on the K400 dataset. All other SSL models are also pre-trained on the Kinetics-400 dataset.}}
    \begin{tabular}{lcccc}
    \toprule
      \multirow{2}{*}{Method}   & \multicolumn{2}{c}{Static UCF-101} & \multicolumn{2}{c}{Dynamic UCF-101} \\
      \cmidrule(r){2-3} \cmidrule(r){4-5}
      & Top-1 ACC (\%) & Top-5 ACC (\%) & Top-1 ACC (\%) & Top-5 ACC (\%)  \\
    \midrule
      Supervised (K400)  &\bf 78.30 &\bf 94.24 &\bf 34.86 &\bf 71.62 \\
    \midrule
      V-SwAV$_{\rho=2}$ \cite{feichtenhofer_large-scale_2021} & 64.84 & 87.52 & 25.56 & 63.38   \\
      V-BYOL$_{\rho=2}$ \cite{feichtenhofer_large-scale_2021} & 70.74 & 89.14 & 27.77 & 67.94 \\
      V-SimCLR$_{\rho=2}$ \cite{feichtenhofer_large-scale_2021} & 63.52 & 84.99 & 21.54 & 60.76 \\
      V-MoCo$_{\rho=2}$ \cite{feichtenhofer_large-scale_2021} & {73.72} & {92.18} & {29.47} & {68.49} \\
    \bottomrule
    \end{tabular}
    \label{tab:motivation}
\end{table*}

\subsection{Empirical Analysis}
\label{subsec:problem}
Efficient video representations should not only capture static visual information but also effectively reflect dynamic changes within a video \cite{kong2022human,schiappa_self-supervised_2022, madan2024foundation}. Accordingly, video semantics can be categorized into static semantics and dynamic semantics. Static semantics refer to visual content that remains unchanged over time, such as the appearance of objects, characters, and backgrounds. In contrast, dynamic semantics describe patterns of change within a video, including object movements and action trajectories. We present an example of static and dynamic semantics in Figure \ref{fig:dynamic feature}. When applying video representations to downstream tasks, static semantics are used to distinguish scenes and objects, while dynamic semantics are used to differentiate actions and trends. Given the fundamental role of both static and dynamic semantics, it is necessary to assess how well current V-CL methods capture these two types of semantic information.

Therefore, we propose to evaluate the capability of V-CL pre-trained models to capture static and dynamic semantics using a controlled variable approach. 

{We employ the UCF-101 dataset \cite{soomro2012ucf101} as the evaluation benchmark and assess four representative V-CL pre-training methods: V-SwAV, V-BYOL, V-SimCLR, and V-MoCo. In addition, we also compare against a supervised pre-trained model. All models are pre-trained on the Kinetics-400 dataset \cite{kay2017kinetics} for 200 epochs.}
{The evaluation comprises two parts: static semantics and dynamic semantics. For the static semantics evaluation, we control for dynamic semantics by keeping them constant across all samples. Concretely, a single frame is extracted from each video in the UCF-101 training set and duplicated to construct a video clip, ensuring that the only distinguishing factors among samples are static semantics, since no motion is present. These clips are then fed into the pre-trained models with frozen parameters, and a linear classifier is trained on the resulting representations, as described in Section \ref{subsec:cl}. The same procedure is applied to the UCF-101 test set, and we report the Top-1 and Top-5 accuracies (\%) on the static version of the UCF-101 test set in Table \ref{tab:motivation}. Poor performance in this setting suggests that the feature extractor produces non-discriminative representations, indicating its limited capacity to capture static semantics.}

{For the dynamic semantics evaluation, static semantics are controlled by keeping them identical across all samples. To this end, we adopt the approach of Ilic et al. \cite{ilic2022appearance}, which effectively removes static semantics. The procedure involves three steps: (i) generating an initial noise frame with the same spatial resolution as the input video frames; (ii) estimating the inter-frame optical flow between consecutive frames of the original video using RAFT \cite{teed2020raft}; and (iii) recursively warping the initial noise frame with the extracted flows to produce a sequence of noise frames whose motion patterns mirror those of the original video. We then apply the same linear evaluation protocol and report the Top-1 and Top-5 accuracies (\%) on the dynamic version of the UCF-101 test set in Table \ref{tab:motivation}. Low performance in this setting indicates that the feature extractor fails to effectively capture dynamic semantics.}

As shown in Table \ref{tab:motivation}, models pre-trained using supervised learning consistently achieve higher classification accuracy than those pre-trained with V-CL across both tasks. Specifically, in the experiment for evaluating static semantics, we observe that the Top-1 accuracy of models pre-trained with V-CL is, on average, $10.10\%$ lower than that of models pre-trained with supervised learning. Since the dynamic information in all samples is controlled to be identical, the classification task relies entirely on the static semantics learned by the pre-trained model. Therefore, these results indicate that V-CL methods struggle to effectively capture static semantics in videos.  {In the dynamic semantics experiment, models pre-trained with V-CL exhibit an average accuracy $8.78\%$ lower than their supervised counterparts.} 
As the static information in all samples is controlled to be identical, the classification task primarily depends on dynamic features. This suggests that V-CL methods also struggle to capture temporal variations and motion dynamics in videos. Combining the results from both experiments, we conclude that V-CL methods are significantly less effective than supervised pre-training in capturing both static and dynamic semantics. Since downstream tasks such as action recognition and action detection typically rely on both types of semantics, the limitations of V-CL pre-trained models revealed by these experiments may negatively impact their performance in real-world applications. 

\begin{figure}
    \centering
    \includegraphics[width=.5\linewidth]{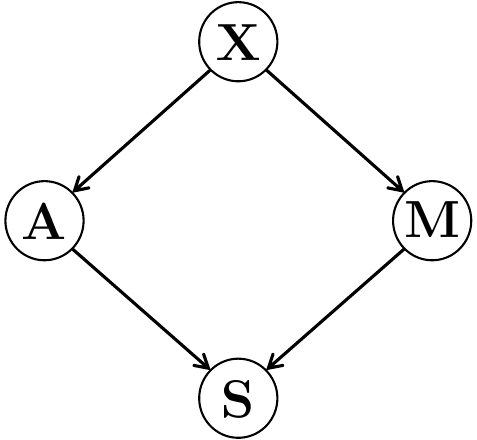}
    \caption{A graphical illustration of the proposed SCM. In this SCM, $\mathbf{X}$ represents the video samples, $ \mathbf{A} $ represents static semantics, $ \mathbf{M} $ represents dynamic semantics, and $ \mathbf{S} $ represents the similarity score. }
    \label{fig:scm}
\end{figure}

\subsection{Theoretical Analysis}
\label{subsec:causal_anal}
In the previous section, we observed that V-CL methods are less effective than supervised pre-training in capturing both static and dynamic semantics. \textbf{This raises the question: Why do V-CL methods exhibit such limitations in semantic representation?} To address this, we conduct a theoretical analysis from a causal and 
gradient optimization perspective.

\paragraph{Existance of Confounding Effect} To analyze the limitations of V-CL methods in capturing static and dynamic semantics from a causal perspective, we first construct a structural causal model (SCM), as illustrated in Figure \ref{fig:scm}. In this model, we define the following random variables: \(\mathbf X \) represents the video sample, \(\mathbf A \) represents the static semantics of a video (e.g., object appearance and scene attributes), \(\mathbf M \) represents the dynamic semantics (e.g., motion trajectories and action patterns), and \(\mathbf S \) denotes the similarity score. In the SCM, each edge represents a causal relationship between variables. The relationship \(\mathbf A \gets\mathbf X \to\mathbf M \) indicates that both static and dynamic semantics are extracted from the video sample. The relationship \(\mathbf A \to\mathbf S \gets\mathbf M \) reflects that the similarity score is computed based on both static and dynamic semantics.

In causal inference, if an external variable simultaneously affects multiple explanatory variables, this variable is referred to as the \textbf{confounding} variable \cite{pearl2009causality, glymour_causal_2016}. The existence of this confounding variable may introduce spurious correlation among the affected variables \cite{glymour_causal_2016}. Specifically, in Figure \ref{fig:scm}, the path \(\mathbf A \gets\mathbf X \to\mathbf M\) indicates that \(\mathbf X\) influences both static semantics \(\mathbf A\) and dynamic semantics \(\mathbf M\). Therefore, $\mathbf X$ is a confounding variable of $\mathbf A$ and $\mathbf M$, and:
\begin{equation}
P(\mathbf A,\mathbf M) = \mathbb{E}_{\mathbf X}[P(\mathbf A,\mathbf M|\mathbf X)] \neq P(\mathbf A)P(\mathbf M).
\end{equation}
This equation indicates that the joint distribution of the static semantic variable \(\mathbf A\) and the dynamic semantic variable \(\mathbf M\) is obtained by marginalizing over the expectation of the conditional distributions \(P(\mathbf A,\mathbf M \mid \mathbf X)\) with respect to the distribution \(P(\mathbf X)\). In practical scenarios, due to selection bias in the dataset \cite{khosla2012undoing, liRESOUNDActionRecognition2018}, certain static semantics (for example, a playground) tend to co-occur with specific dynamic semantics (for example, playing soccer). Consequently, even if these two semantic components are intrinsically independent, they exhibit statistical dependence in the observed data.

\paragraph{Impact of Confounding on Learning Semantics}
In current V-CL training processes, models typically compute a unified similarity metric without distinguishing the contributions of static and dynamic semantics, as illustrated in Equations (\ref{eq:infoNCE}) and (\ref{eq:byol}). In other words, as long as the similarity between two video representations in the embedding space is sufficiently high, the model does not differentiate whether that similarity arises from static or dynamic semantics. Moreover, it is well known that neural networks tend to adopt the simplest and most efficient shortcuts to minimize the loss function \cite{geirhos2020shortcut,beery2018recognition,hermann2023foundations}. Consequently, due to the correlations introduced by confounding effects, once a particular semantic feature (e.g., a basketball court) is learned, it becomes sufficient for the model to classify videos containing the associated action (e.g., playing basketball) as similar. {This phenomenon may cause the pre-trained model to capture only a subset, namely the easier-to-learn subset, of the full semantic information, leading to the poor performance observed in Section \ref{subsec:problem}.} \footnote{Note that we don't specify whether the easier-to-learn subset is static or dynamic. It may contain both of them.}. To illustrate this shortcut learning phenomenon formally, we provide the following analysis.

Consider a learnable feature extractor \( f \) that takes a video \( X \) as input and produces an output embedding $Z$ that captures both static and dynamic semantic information. For analytical convenience, we assume that \( f \) decomposes additively as:
\begin{equation}
f(X) = g_A(X;\theta_A) + g_M(X;\theta_M)=Z ,    
\end{equation}
where \( g_A \) and \( g_M \) denote the sub-networks (or pathways) responsible for extracting static and dynamic semantics with \(\theta_A\) and \(\theta_M\) representing their corresponding parameters, respectively. In particular, $g_A(X;\theta_A) = \bigl(a^{(1)}, \dots, a^{(M)}\bigr)$ and $ g_M(X;\theta_M) = \bigl(m^{(1)}, \dots, m^{(M)}\bigr)$. Here, \(a^{(i)}\) and \(m^{(i)}\) denote specific components of static and dynamic semantics, respectively, and $M$ denotes the total number of semantics components.

The representation $Z$ is used to compute a loss function as shown in Equation (\ref{eq:infoNCE}) or Equation (\ref{eq:byol}), which we collectively denote as \(\mathcal{L}(\theta_A,\theta_M)\). Then, following the conclusions in existing literature \cite{hermann2023foundations, pezeshki2021gradient}, we define the difficulty of learning a particular feature as follows:
\begin{definition}
\label{def:1}
(Ease of Learning) Let \(\mathcal{L}(\theta_A,\theta_M)\) denote the training loss. Define
\begin{equation}
    \Delta_{a^{(j)}}(\theta_A,\theta_M)=\Bigl\|\nabla_{\theta_{a^{(j)}}}\mathcal{L}(\theta_A,\theta_M)\Bigr\|
\end{equation}
as the gradient norm with respect to the subset of parameters \(\theta_{a^{(j)}} \subset \theta_A\) responsible for extracting \(a^{(j)}\), and similarly,
\begin{equation}
    \Delta_{m^{(j)}}(\theta_A,\theta_M)=\Bigl\|\nabla_{\theta_{m^{(j)}}}\mathcal{L}(\theta_A,\theta_M)\Bigr\|.
\end{equation}
We say that \(a^{(j)}\) is \emph{easier to learn} than \(m^{(j)}\) at \((\theta_A,\theta_M)\) if
\begin{equation}
    \Delta_{a^{(j)}}(\theta_A,\theta_M)>\Delta_{m^{(j)}}(\theta_A,\theta_M).
\end{equation}
\end{definition}

Building upon these definitions, we propose the following proposition:
\begin{proposition}[Incomplete Semantic Learning with Unified Loss]
\label{prop:1}
Assume that for each \(j \in \{1, \dots, k\}\), the pair \((a^{(j)}, m^{(j)})\) is confoundedly correlated, meaning that \(a^{(j)}\) and \(m^{(j)}\) provide redundant information about the similarity structure \(S\). Suppose further that, during training (i.e., along the gradient descent iterations), for each \(j\), one component of the pair \((a^{(j)}, m^{(j)})\) is consistently easier to learn than its counterpart, as defined in Definition~\ref{def:1}. Specifically, for each \(j\), either \(\Delta_{a^{(j)}}(\theta_A, \theta_M) > \Delta_{m^{(j)}}(\theta_A, \theta_M)\) or \(\Delta_{m^{(j)}}(\theta_A, \theta_M) > \Delta_{a^{(j)}}(\theta_A, \theta_M)\) holds consistently throughout training.

Then, when the loss function \(\mathcal{L}\) measures similarity using both \(A\) and \(M\) via the representation \(Z = g_A(X; \theta_A) + g_M(X; \theta_M)\), the learned representation converges as \(t \to \infty\) to a solution where, for each \(j\), the contribution from the harder semantic component vanishes. That is, for each \(j\), either
\begin{equation}
    \begin{split}
        &\lim_{t \to \infty} g_M^{(j)}(X; \theta_M^{(t)}) \approx g_M^{(j)}(X; \theta_M^{(0)}) \quad \text{or} \\
        &\lim_{t \to \infty} g_A^{(j)}(X; \theta_A^{(t)}) \approx g_A^{(j)}(X; \theta_A^{(0)}),
    \end{split}
\end{equation}
depending on which component is harder to learn. Consequently, rather than jointly encoding both \(a^{(j)}\) and \(m^{(j)}\) for each semantic dimension \(j\), the final representation encodes only the easier semantic component.
\end{proposition}
Proof is provided in Appendix \ref{app:proof}. Proposition \ref{prop:1} demonstrates that if confounding effects induce a correlation between static semantics and dynamic semantics and if the similarity in V-CL training is computed using both types of semantics simultaneously, then during optimization, the model tends to optimize only the parameters corresponding to the easier-to-learn semantics. This occurs because the correlation between the two types of semantics allows the optimization of the easier-to-learn component to sufficiently reduce the loss function, leaving the parameters associated with the harder-to-learn semantics under-optimized. Consequently, the final model is only capable of extracting the easier-to-learn semantics, which leads to poor performance on downstream tasks that require discrimination based solely on either static or dynamic semantic information.

\paragraph{Decoupled Loss for Comprehensive Learning}
Through the analysis above, we observe that the primary cause behind Proposition \ref{prop:1} is that, when computing similarity, the easier-to-learn semantic is sufficient to minimize the loss function, thereby impeding the learning of the more difficult semantic. Consequently, if losses corresponding to static similarity and dynamic similarity could be computed separately, the parameter update processes for the two types of semantics would not interfere with each other. We provide a detailed analysis of this phenomenon in the following Corollary.

\begin{corollary}[Complete Representation with Decoupled Losses]
\label{coro:1}
Under the same structural assumptions as in Proposition~\ref{prop:1}, suppose that the loss function is decoupled so that similarity is measured independently on the static and dynamic features. The overall loss as
\begin{equation}
    \mathcal{L}_{\text{sep}}(\theta_A,\theta_M) = \mathcal{L}_A(\theta_A) + \mathcal{L}_M(\theta_M),
\end{equation}
where \(\mathcal{L}_A(\theta_A)\) and \(\mathcal{L}_M(\theta_M)\) are loss functions that measure similarity solely on the static or dynamic representations, respectively. Then, gradient descent will optimize $\theta_A$ and $\theta_M$ independently. Consequently, for each $j$, both
\begin{equation}
    \Delta_{a^{(j)}}^{(A)} = \Bigl\|\nabla_{\theta_{a^{(j)}}}\mathcal{L}_A\Bigr\| \quad \text{and} \quad \Delta_{m^{(j)}}^{(M)} = \Bigl\|\nabla_{\theta_{m^{(j)}}}\mathcal{L}_M\Bigr\|
\end{equation}
will be preserved without interference from each other. Consequently, the final representation effectively encodes both the easier semantic and the hard semantic.
\end{corollary}
Proof is provided in Appendix \ref{app:proof_2}. Corollary \ref{coro:1} demonstrates that if the static similarity loss and dynamic similarity loss can be computed separately, then even if one type of semantic is inherently more difficult to learn, their respective learning processes will not interfere with one another. Consequently, a feature extractor trained under such a regime is expected to achieve superior performance on downstream tasks. The key challenge in this modeling approach, however, lies in how to determine and separate static and dynamic semantics from the actual output of the feature extractor.

\begin{figure*}[htb]
    \centering
\includegraphics[width=\linewidth]{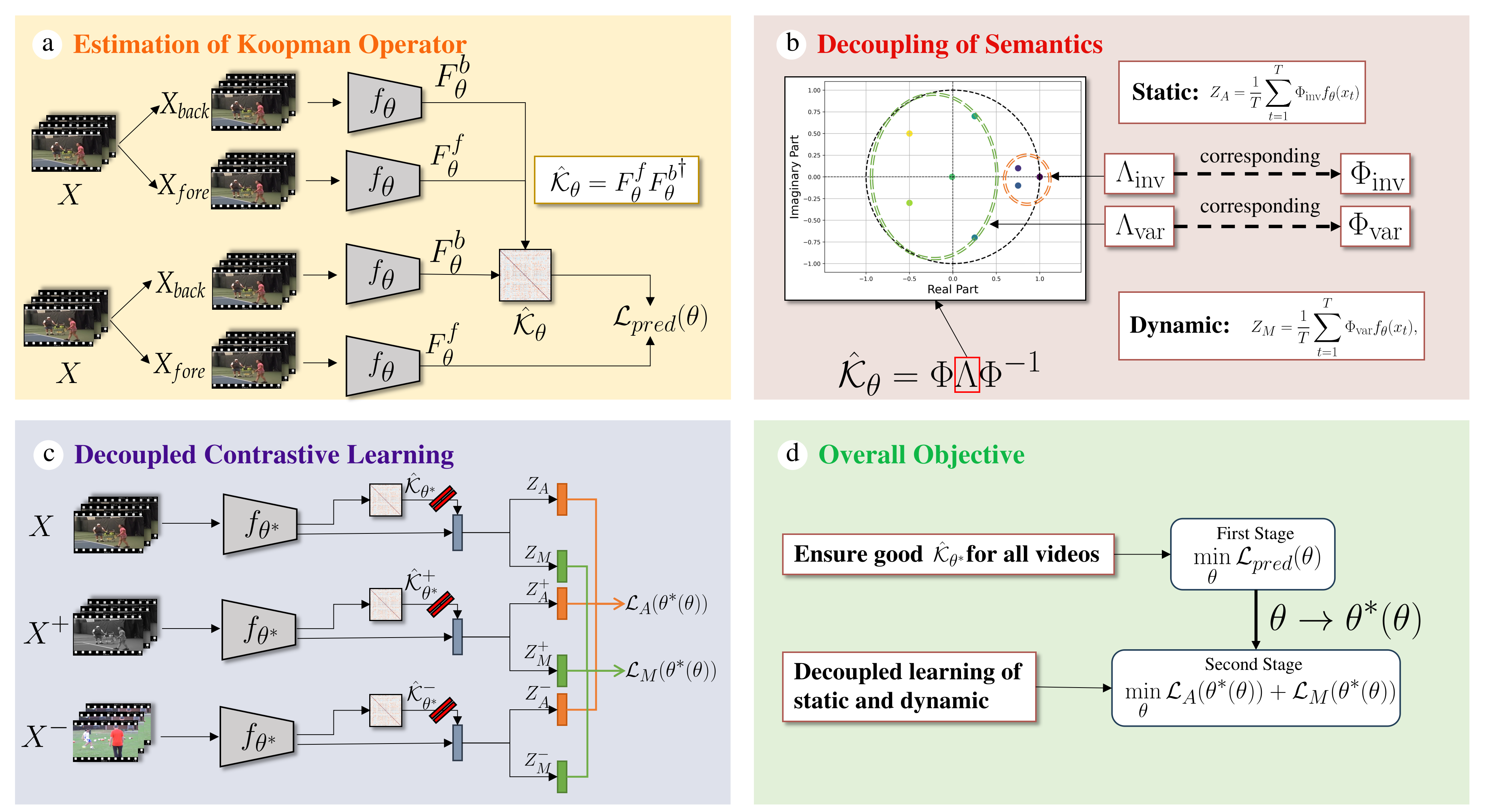}
    \caption{Overview of the proposed BOD-VCL framework. (a) The process of the estimation of $\hat{\mathcal{K}}_\theta$. $\mathcal{L}_{pred}(\theta)$ ensure the estimated $\hat{\mathcal{K}}_\theta$ precisely describe inter-frame evolution. $X_{back}$ denotes the first $T-1$ frames while $X_{fore}$ denotes the subsequent $T-1$ frames of video $X$.  (b) Eigen-decomposition is performed on \(\hat{\mathcal{K}}_\theta\) to separate eigenvalues into time-invariant and time-variant subsets, corresponding to static and dynamic semantics, respectively. The associated eigenvectors are then used to extract static and dynamic features independently.  (c) The separated static and dynamic semantic representations are utilized to independently optimize contrastive objectives, ensuring effective learning of both components.  (d) The overall optimization is formulated as a bi-level process. In the first stage, the prediction loss \(\mathcal{L}_{pred}\) is minimized to obtain a reliable Koopman operator \(\hat{\mathcal{K}}_\theta\). In the second stage, given this operator, the decoupled contrastive losses \(\mathcal{L}_A\) and \(\mathcal{L}_M\) are optimized to obtain better static and dynamic representations.  }
    \label{fig:framework}
\end{figure*}

\section{Methodology}
\label{sec:method}
In this section, we first present the motivation behind our approach. Then, we provide a detailed formal definition of static and dynamic semantics. Next, we describe the specific steps of the proposed BOD-VCL method. Finally, we introduce the overall objective function.

\subsection{Motivation}
Based on the theoretical analysis in Subsection \ref{subsec:causal_anal}, we conclude that the confounding effects inherent in the dataset lead to suboptimal learning of both static and dynamic features when an undifferentiated learning strategy is used. Proposition \ref{prop:1} suggests that in existing V-CL methods, the simultaneous use of static and dynamic semantics during similarity computation can cause the model to favor the easier-to-learn semantics, thereby neglecting the harder-to-learn ones. To address this issue, Corollary \ref{coro:1} suggests that it is necessary to develop an approach that decouples the learning process for static and dynamic semantics. However, there are two key challenges: (1) clearly defining and distinguishing between the static and dynamic semantics present in video data, and (2) devising a method to effectively separate these two types of semantic information during feature extraction. 

To address these challenges, we first revisit the semantics of video from a dynamic system perspective. Building on this perspective, we propose \textit{Bi-level Optimization with Decoupling for Video Contrastive Learning} (BOD-VCL), a novel modeling approach designed to effectively separate static and dynamic semantics and learn them in a decoupled manner. 

\subsection{Rethinking Static and Dynamic Semantics}
\label{subsec:rethinking}
In this subsection, we analyze the composition of static semantics and dynamic semantics from a frame-evolution perspective. {Specifically, we represent a video as $X = [x_1, \dots, x_T]$, a sequence of frames where each $x_t$ denotes an individual frame. In video $X$, the transition from one frame, $x_t$, to the next, $x_{t+1}$, can be described by the following discrete-time dynamic system:
\begin{equation}
\label{eq:ode}
    x_{t+1} = F(x_t), \quad \forall x_t, x_{t+1} \in X.
\end{equation}
Here, $F(\cdot)$ is the transition function and provides a compact characterization of the temporal dependencies within the observed video sequence. From the perspective of Equation (\ref{eq:ode}), a video can be treated as a system evolving frame by frame over time, where each frame is a result of applying \( F(\cdot) \) to the preceding one. Viewing the video in this manner enables a systematic examination of its temporal evolution and provides a foundation for decomposing its semantic components. A classical example of such a transition operator is optical flow, which explicitly models pixel-level correspondences between consecutive frames.}

The transition function \(F(\cdot)\) inherently determines how the semantics of each frame evolve over time. Specifically, we consider an idealized setting where an ideal feature extractor \( f^* \) maps any given frame \( x_t \) into a lower-dimensional semantic vector \(\boldsymbol{z}_t\). This semantic representation can be partitioned into two components: the invariant component \(\boldsymbol{z}^{\text{inv}}_t\) and the variant component \(\boldsymbol{z}^{\text{var}}_t\). {The invariant component \(\boldsymbol{z}^{\text{inv}}_t\) captures semantic attributes such as object appearance and background features that remain unchanged under \( F(\cdot) \) (i.e., \(\boldsymbol{z}^{\text{inv}}_{t+1} = \boldsymbol{z}^{\text{inv}}_t\)). In contrast, the variant component \(\boldsymbol{z}^{\text{var}}_t\) corresponds to dynamic semantics such as object position and motion patterns that are influenced by \( F(\cdot) \) and evolve over time (i.e., \(\boldsymbol{z}^{\text{var}}_{t+1} \neq \boldsymbol{z}^{\text{var}}_t\)).}

In practice, the representation of a video is typically obtained by aggregating frame-level feature vectors produced by a feature extractor (e.g., through average pooling \cite{schiappa_self-supervised_2022,kong2022human, karpathy2014large}). Accordingly, we define the static semantics of a video as the average of the extracted time-invariant components across all frames, capturing attributes that remain consistent throughout the video:  $Z_\text{static} = \frac{1}{T} \sum_{t=1}^{T} \boldsymbol{z}_t^\text{inv}$. Similarly, we define the dynamic semantics as the averaged time-variant components across frames, representing temporal variations such as motion patterns and actions:  $Z_\text{dynamic} = \frac{1}{T} \sum_{t=1}^{T} \boldsymbol{z}_t^\text{var}$. 

However, in the practical V-CL process, although the representation of each frame is also aggregated into a single vector through average pooling \cite{feichtenhofer_large-scale_2021, feichtenhofer_slowfast_2019, qian_spatiotemporal_2021, severe, sarkar_uncovering_2023}, each dimension of the frame-level representation often simultaneously encodes both time-invariant and time-varying attributes \cite{berman2022multifactor}. As a result, it becomes challenging to determine which components correspond to the static semantics of the video and which correspond to the dynamic semantics. 

In summary, if we can precisely differentiate which semantic components remain invariant over time and which do not, we can distinguish between the static and dynamic semantics of the videos. 

\subsection{Main Components}
\label{subsec:KMD}
To effectively distinguish between static and dynamic semantics in videos, it is essential to model the inter-frame transition function \( F(\cdot) \). However, the major challenge lies in that the transition function \( F(\cdot) \) in real-world videos is typically highly nonlinear \cite{birkhoff1927dynamical}, making it difficult to utilize $F(\cdot)$ to find the time-invariant components and the time-variant inside the video.

A key observation is that if the transition function is a linear function, then the separation of static and dynamic semantics becomes significantly more tractable. Specifically, linear functions allow us to leverage eigenvalue decomposition to naturally separate these two components \cite{schmid2010dynamic, williams_datadriven_2015, brunton_modern_2021}. However, since video transitions are inherently nonlinear, this approach cannot be directly applied.  

Koopman theory provides a theoretical foundation for addressing this issue by enabling the transformation of nonlinear dynamical systems into linear ones \cite{koopman1931hamiltonian, brunton_modern_2021}. Specifically, Koopman theory states that any nonlinear dynamical system can be mapped into a high-dimensional space, where the originally nonlinear system can be represented as a linear dynamical system. The transition function in this linear system is also called a Koopman operator. Building on this insight, we propose to first estimate a Koopman operator from video data and then utilize the learned Koopman operator to separate static and dynamic semantics through eigenvalue decomposition. With the separated semantics, it is feasible to calculate static and dynamic similarities independently.

\paragraph{Estimation of Koopman Operator}
In this subsection, we propose a method based on Koopman theory that transforms the non-linear transitions between consecutive frames into linear dynamics. 

The central idea behind Koopman theory is that the nonlinear transformation $F(\cdot)$ in the original space $\mathcal{X}$ is equivalent to a linear transformation in another infinite dimensional space $\mathcal{U}$. Specifically,  a function $\mathcal{G}(\cdot):\mathcal{X}\to\mathcal{U}$ transforms a video frame $x_t \in \mathcal{X}$ into a point in the space $\mathcal{U}$. The components of $\mathcal{G}(x_t)$ are given by a set of infinitely many observation functions $g_i$, i.e., $ \mathcal{G}(x_t) = [g_1(x_t), g_2(x_t), \dots, g_i(x_t), \dots ]^\top$. These observation functions $g_i: \mathcal{X} \to \mathbb{R}$ can be considered to capture different properties of $\mathcal{X}$. Koopman \cite{koopman1931hamiltonian} proved that, by appropriately selecting these functions, the non-linear transition function between consecutive frames in the space $\mathcal{X}$ can be represented by a linear transformation $\mathcal{K}: \mathcal{U} \to \mathcal{U}$ in the space $\mathcal{U}$. In other words, Equation (\ref{eq:ode}) can be equivalently rewritten as: 
\begin{equation}\label{qww121} 
\mathcal{K} \mathcal{G}(x_t) = \mathcal{G}(F(x_t)) = \mathcal{G}(x_{t+1}). 
\end{equation}
The linear operator $\mathcal{K}$ above is also called the Koopman operator. In this case, we can interpret $\mathcal{K}$ as governing the evolution of frames in the video.

Despite that $\mathcal{K}$ is by definition infinite dimensional, recent lines of work often interested in finding a finite-dimensional approximation of it using neural networks \cite{azencot_forecasting_2020, schmid2010dynamic,takeishi2017learning,lusch2018deep}. In other words, instead of finding infinitely many observable functions of $g_i$, these works are only interested in finding a finite subset of them that spans a Koopman invariant subspace \cite{takeishi2017learning,brunton_modern_2021}. 
{Inspired by these works, we propose to approximate a mapping to a Koopman invariant subspace using a feature extractor \( f_\theta \), parameterized by a neural network. Specifically, given a video clip \( X \), the encoder produces a sequence of per-frame features from the same forward pass, where we use the notation \( f_\theta(x_t)\in\mathbb{R}^M \) to denote the feature corresponding to the $t$-th frame. Here, $M$ is the feature dimension. We then construct $F^b_\theta = [f_\theta(x_1), f_\theta(x_2), \dots, f_\theta(x_{T-1})]\in\mathbb{R}^{M\times (T-1)}, \quad
F^f_\theta = [f_\theta(x_2), f_\theta(x_3), \dots, f_\theta(x_T)]\in\mathbb{R}^{M\times (T-1)},$ which represent the feature representations of the initial \(T-1\) frames and their subsequent \(T-1\) frames, respectively.}
{Then, we can estimate a Koopman operator $\hat{\mathcal{K}}_\theta \in \mathbb{R}^{M \times M}$ by computing the least-squares regression solution between consecutive frame features:
\begin{equation}
\label{eq:K_target}
\begin{split}
    \hat{\mathcal{K}}_\theta 
    &= \arg\min_K \sum_{t=0}^{T-1} \big\|K f_{\theta}(x_t) - f_\theta(x_{t+1})\big\|_2^2  \\
    &= F^f_\theta \, {F^b_\theta}^\dagger,
\end{split}
\end{equation}
where $\|\cdot\|_2$ denotes the Euclidean norm and $^\dagger$ is the Moore–Penrose pseudoinverse. Importantly, according to Equation \ref{eq:K_target}, $\hat{\mathcal{K}}_\theta$ can be computed with $F^f_\theta \, {F^b_\theta}^\dagger$ as the closed-form.}

Ideally, $f_\theta$ should map the video segment \( X \) into a feature space where the optimal linear transition matrix \( \hat{\mathcal{K}}_\theta \) can reliably predict the next-frame representations from the current ones. This feature space is also known as the Koopman invariant subspace \cite{bruntonKoopmanInvariantSubspaces2016, takeishi2017learning}. In other words, for each frame \(x_t\), applying \(\hat{\mathcal{K}}_\theta\) to \(f_\theta(x_t)\) should ideally yield \(f_\theta(x_{t+1})\). Consequently, we propose to optimize the following objective function:
\begin{equation}
\label{eq:dynamic}
\scalebox{.95}{
$
\begin{split}
    &\min_\theta \mathcal{L}_{pred}(\theta), \\
    &\text{s.t } \mathcal{L}_{pred}(\theta)=\mathcal{L}_\text{MSE}({F^f_\theta},\hat{\mathcal{K}}_\theta {F^b_\theta}) + \upsilon \Vert \hat{\mathcal{K}}_\theta^\top\hat{\mathcal{K}}_\theta-I \Vert_F^2.
\end{split}
$
}
\end{equation}
where \( \mathcal{L}_{\text{MSE}} \) denotes the mean squared error loss, enforcing the alignment between the predicted and actual next-frame representations. {The term $\upsilon \, \| \hat{\mathcal{K}}_\theta^\top \hat{\mathcal{K}}_\theta - I \|_F^2$ serves as an orthogonality regularizer, where $\upsilon$ is the corresponding Lagrange multiplier. This constraint enforces $\hat{\mathcal{K}}_\theta$ to be approximately orthogonal and thus full-rank. In particular, it avoids the case where the feature extractor $f_\theta$ collapses all frames $F^b$ and $F^f$ to constant vectors, ensuring that distinct frame features remain distinguishable after the linear mapping.}

Minimizing Equation (\ref{eq:dynamic}) ensures the feature extractor maps each frame into a subspace where $\hat{\mathcal{K}}_\theta$ can effectively transform $x_t$ to $x_{t+1}$. In other words, we transform the non-linear transitions between frames into linear ones.

\begin{figure*}[htb]
    \centering
    \subfigure[Eigenvalues in the complex plane.]{%
        \includegraphics[width=0.45\textwidth]{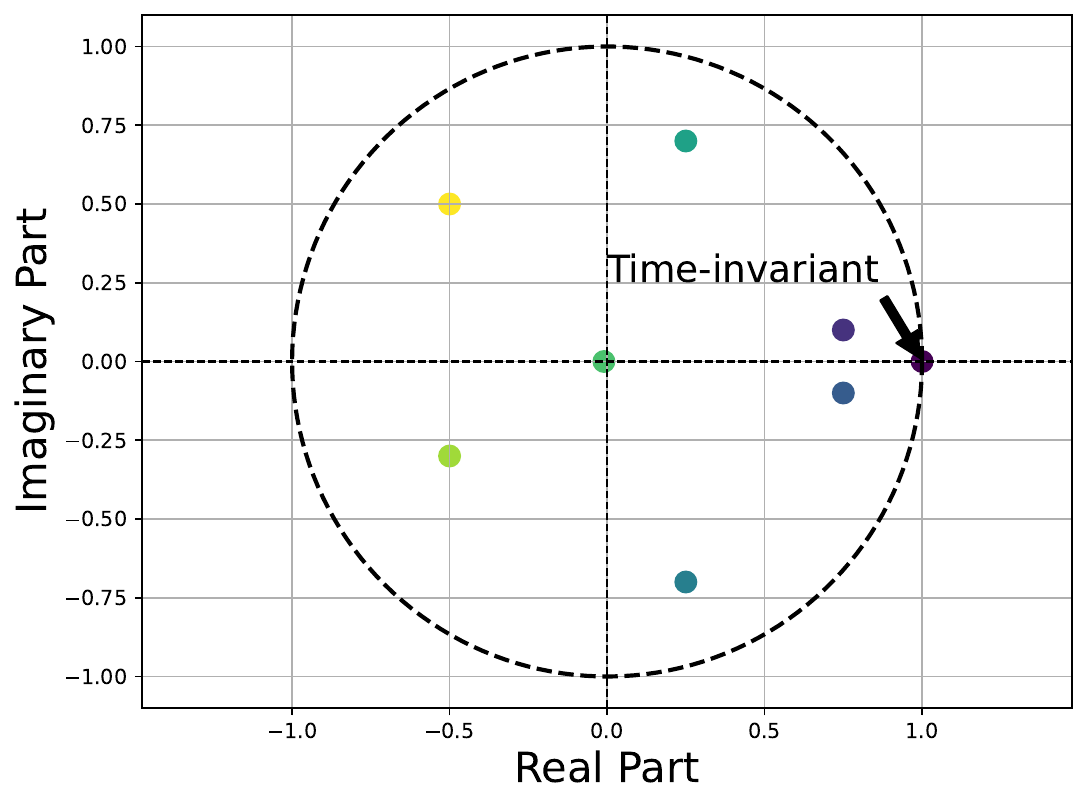}
        \label{fig:eigen_complex}
    }
    \subfigure[Amplitudes over time.]{%
        \includegraphics[width=0.45\textwidth]{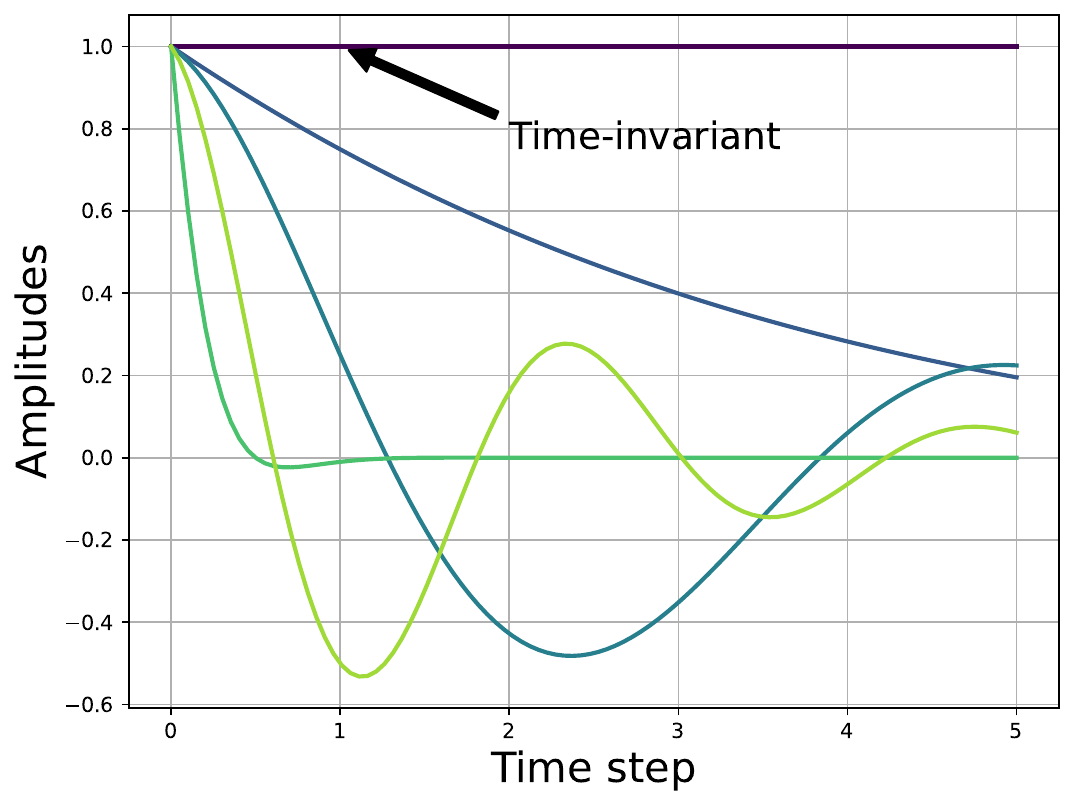}
        \label{fig:eigen_evolution}
    }
    \caption{(a) The illustration of different eigenvalues $\lambda_m$ in the complex plane. The eigenvectors near $ (1 + 0j) $ are related to the time-invariant semantics, while others are related to time-variant semantics. (b) Temporal evolution of feature amplitudes $\varphi_m^\top f_\theta(x_{t+k})$ corresponding to different $\lambda_m$ over time. Specifically, when $\lambda_m\approx1$, $\varphi_m^\top f_\theta(x_{t+k})$ remains nearly constant for different value of $k$.}
    \label{fig:eigen_explain}
\end{figure*}

\paragraph{Decoupling of Semantics}
\label{subsec:stratify}
By optimizing Equation (\ref{eq:dynamic}), we ensure that for any given video, the optimal transition matrix \( \hat{\mathcal{K}}_\theta \) can be obtained via Equation (\ref{eq:K_target}). Although the learned \(\hat{\mathcal{K}}_\theta\) can be considered to characterize how semantics evolves between frames, it does not by itself distinguish time-invariant and time-variant components. To explicitly separate these two types of semantics, we leverage the eigen-decomposition of \(\hat{\mathcal{K}}_\theta\).

This decomposition is expressed as $ \hat{\mathcal{K}}_\theta = \Phi \Lambda \Phi^{-1} $, where $ \Phi = [\varphi_1, \dots, \varphi_M] \in \mathbb{C}^{M \times M} $ contains the eigenvectors, and $ \Lambda = \operatorname{diag}([\lambda_1, \dots, \lambda_M]) \in \mathbb{C}^{M \times M} $ is the diagonal matrix of eigenvalues. The notation ``$ \operatorname{diag} $" indicates that the eigenvalues are arranged along the diagonal of the matrix. Since $ \hat{\mathcal{K}}_\theta $ is not necessarily a symmetric matrix, both $ \Phi $ and $ \Lambda $ are in the complex space $ \mathbb{C} $. Then, according to Equation (\ref{qww121}), we have:
\begin{equation}
\label{eq:evd_koop}
\begin{split}
    f_\theta(x_{t+1}) &= \hat{\mathcal{K}}_\theta f_\theta(x_t), \\
    f_\theta(x_{t+1}) &=\Phi \Lambda \Phi^{-1} f_\theta(x_t), \\
    f_\theta(x_{t+2}) &= \Phi \Lambda \Phi^{-1} \Phi \Lambda \Phi^{-1} f_\theta(x_t), \\
    &\vdots \\
    f_\theta(x_{t+k}) &= \Phi \Lambda^{k}\Phi^{-1} f_\theta(x_t).
\end{split}
\end{equation}
According to Equation (\ref{eq:evd_koop}), we observe that by projecting $ f_\theta(x_t) $ into the eigenvector space using $ \Phi^{-1} $, any future state $ f_\theta(x_{t+k}) $ can be obtained by applying $ \Lambda $ with $k$ times to this projection and then mapping back using $ \Phi $. Since $ \Lambda $ is a diagonal matrix containing the eigenvalues and $ \Phi $ consists of the corresponding linearly independent eigenvectors, we can say that after eigenvalue decomposition, the dynamics represented by $ \hat{\mathcal{K}}_\theta $ can be expressed as independent motions governed by $ \Lambda $ along the basis vectors provided by $ \Phi $. 

Now, let us take a closer look at a specific eigenvalue \(\lambda_m\) from the set \(\Lambda\). Since \(\lambda_m \in \mathbb{C}\) is a complex number, it can be expressed as \(\lambda_m = Re_m + j Im_m\), where \(Re_m\) and \(Im_m\) denote its real and imaginary parts respectively, and \(j\) represents the imaginary unit. Based on the eigen-decomposition definition, we have the relation \(\varphi_m^\top\hat{\mathcal{K}}_{\theta}=\lambda_m\varphi_m^\top\). 
Substitute this into Equation (\ref{eq:evd_koop}), we obtain \(\varphi_m^\top f_\theta(x_{t+k})=\lambda^k_m\varphi^\top_m f_\theta(x_t)\). Consequently, if \(\lambda_m \approx 1+0j\), the projection of \(f_\theta(x_t)\) onto \(\varphi_m\) remains nearly constant over time, i.e., \(\varphi_m^\top f_\theta(x_{t+k})\approx \varphi_m^\top f_\theta(x_t)\). This indicates that eigenvalues close to \(1+0j\) correspond to the time-invariant semantics of the video. We also provide an illustration of this property in Figure \ref{fig:eigen_explain}.

Building upon this phenomenon, we can now separate the time-invariant and time-variant parts of a video. Specifically, we partition the eigenvalues of \( \hat{\mathcal{K}}_\theta \) based on their distance from \( 1 + 0j \). Define \( \Lambda_{\text{inv}} \) as the diagonal matrix that retains eigenvalues satisfying \( |\lambda - 1| < \xi \) while setting others to zero, and \( \Lambda_{\text{var}} \) as the diagonal matrix that retains eigenvalues with \( |\lambda - 1| \geq \xi \) while setting others to zero. 
 
Next, as discussed in Subsection \ref{subsec:rethinking}, the static semantics of a video can be obtained by aggregating the time-invariant components across the entire sequence, while the dynamic semantics are derived by aggregating the time-variant components. Formally, this process can be expressed as:  
\begin{equation}
\label{eq:separate}
\begin{split}
Z_{A} = \frac{1}{T} \sum\limits_{t=1}^{T} \Phi\Lambda_{\text{inv}}\Phi^{-1} f_{\theta}(x_t),\\
Z_{M} = \frac{1}{T} \sum\limits_{t=1}^{T} \Phi\Lambda_{\text{var}}\Phi^{-1} f_{\theta}(x_t),
\end{split}
\end{equation}  
where $Z_A$ and $Z_M$ are the static and dynamic feature representation of video $X$, respectively. Up to this point, we have successfully decomposed video representations into static and dynamic components by modeling the Koopman operator and leveraging eigenvalue-based separation. With this foundation in place, the approach outlined in Corollary \ref{coro:1} now becomes feasible, allowing us to independently learn static and dynamic similarities, thereby addressing the issue present in Proposition \ref{prop:1}.

\paragraph{Decoupled Contrastive Learning}
According to Proposition \ref{prop:1}, the conventional V-CL training paradigm, which does not differentiate between static and dynamic semantics, leads to a bias where the model prioritizes the easier-to-learn semantics. Corollary \ref{coro:1} suggests that this issue can be mitigated by separately modeling static and dynamic similarity losses, ensuring that both types of semantics are effectively learned. In Equation (\ref{eq:separate}), we proposed the method for separate static and dynamic semantics, which serves as the foundation for defining independent loss functions. Building upon this formulation, if the original V-CL framework employs an InfoNCE loss, its decoupled version can be formulated as follows:
\begin{equation}
\label{eq:decoupled_CL}
\scalebox{.73}{
$
\begin{split}
&\min_\theta  \mathcal{L}_{A}(\theta) + \mathcal{L}_{M}(\theta), \\
    &\text{s.t. }\mathcal{L}_A(\theta)=-\sum_{i=1}^{N_{bs}}\sum_{j=1}^\varrho\log\frac{\sum_{Z_A^+\in\mathcal{Z}_{{i,j}_A}^+}\exp(\text{sim}(Z_{{i,j}_A},Z_A^+)/\alpha)}{\sum_{Z_A^\prime\in\{\mathcal{Z}_{{i,j}_A}^+,\mathcal{Z}_{i_A}^-\}}\exp(\text{sim}(Z_{{i,j}_A},Z_A^\prime)/\alpha)},\\
    &\mathcal{L}_M(\theta)=-\sum_{i=1}^{N_{bs}}\sum_{j=1}^\varrho\log\frac{\sum_{Z_M^+\in\mathcal{Z}_{{i,j}_M}^+}\exp(\text{sim}(Z_{{i,j}_M},Z_M^+)/\alpha)}{\sum_{Z_M^\prime\in\{\mathcal{Z}_{{i,j}_M}^+,\mathcal{Z}_{i_M}^-\}}\exp(\text{sim}(Z_{{i,j}_M},Z_M^\prime)/\alpha)},\\
\end{split}
$
}
\end{equation}
where \( Z_{{i,j}_A}, Z_A^+, \mathcal{Z}_{{i,j}_A}^+, \mathcal{Z}^-_{i_{A}} \) and \( Z_{{i,j}_M}, Z_M^+, \mathcal{Z}_{{i,j}_M}^+, \mathcal{Z}^-_{i_{M}} \) represent the static and dynamic components of \( Z_{{i,j}}, Z^+, \mathcal{Z}_{{i,j}}^+, \mathcal{Z}^-_{i} \), respectively. Other V-CL objectives can be adapted in a similar manner to incorporate this decoupled learning framework.

\subsection{Overall Objective}
\label{subsec:objective}
In the previous subsection, we proposed two distinct objective functions: Equation (\ref{eq:dynamic}) and Equation (\ref{eq:decoupled_CL}). Specifically, Equation (\ref{eq:dynamic}) aims to ensure that the Koopman operator \(\hat{\mathcal{K}}_\theta\) accurately captures the temporal evolution between video frames, whereas Equation (\ref{eq:decoupled_CL}) ensures that the feature extractor \(f_\theta\) effectively captures both static and dynamic semantics within the videos. Analyzing these two objectives, we observe a notable interdependency: the feature extractor \(f_\theta\) can correctly identify and separate semantic components only if the Koopman operator \(\hat{\mathcal{K}}_\theta\) reliably models inter-frame transitions; conversely, accurately modeling these transitions with the Koopman operator inherently depends upon having meaningful and semantically rich feature representations. This mutual dependency motivates our adoption of a bi-level optimization strategy, wherein the feature extractor and the Koopman operator are optimized jointly in a mutually reinforcing manner.

To address this mutual dependency, we formulate the overall optimization objective as a bi-level optimization problem, defined as follows:
\begin{equation}
\label{eq:bi-level}
\begin{split}
    &\min_\theta \mathcal{L}_{A}(\theta^*(\theta)) + \mathcal{L}_{M}(\theta^*(\theta)), \\
    &\text{s.t. } \theta^*(\theta)= \arg\min_\theta \mathcal{L}_{pred}(\theta).
\end{split}
\end{equation}
where $\mathcal{L}_{A}(\theta^*(\theta))$ and $\mathcal{L}_{M}(\theta^*(\theta))$ represents the loss is calculated based on the optimal parameters of $\mathcal{L}_{pred}(\theta)$, $\theta^*(\theta)$ represents that the optimal parameters are a function of the original parameters.

The learning process consists of two stages. In the first stage, we have:
\begin{equation}
\label{eq:first-level}
\theta^* \gets \theta - \gamma \nabla_{\theta} \mathcal{L}_{pred}(\theta)
\end{equation}
where $\gamma$ is the learning rate. This stage ensures that the estimated $\hat{\mathcal{K}}_{\theta^*}$ can capture all the inter-frame evolution. In the second stage, we have:
\begin{equation}\label{qw12345}
    \theta \gets \theta - \gamma \nabla_{\theta}\left[ \mathcal{L}_{A}(\theta^*) + \mathcal{L}_{M}(\theta^*)\right].
\end{equation}
Building upon an effective $\hat{\mathcal{K}}_{\theta^*}$, this stage ensures that the feature extractor effectively and independently learns static and dynamic semantics from video. 

Through iterative updates of these two optimization steps, the feature extractor gradually improves its capacity to represent meaningful video semantics, while the learned Koopman operator simultaneously becomes more proficient at distinguishing between static and dynamic components. This optimization strategy enables the model to progressively obtain clearer and more accurate representations of both static and dynamic semantics.
\textbf{Note that all notations used in this paper and their specific meanings are displayed in tabular form in the Appendix.}

\begin{table*}[t]
    \centering
    \caption{{Top-1 and Top-5 linear evaluation accuracies of the Video CL methods and the proposed BOD-VCL on the Static UCF-101 and Dynamic UCF-101 datasets. "Supervised (K400)" indicates the model is pre-trained with supervision on the Kinetics-400 dataset. All other SSL models are also pre-trained on the Kinetics-400 dataset.}}
    \begin{tabular}{lcccc}
    \toprule
      \multirow{2}{*}{Method}   & \multicolumn{2}{c}{Static UCF-101} & \multicolumn{2}{c}{Dynamic UCF-101} \\
      \cmidrule(r){2-3} \cmidrule(r){4-5}
      & Top-1 ACC (\%) & Top-5 ACC (\%) & Top-1 ACC (\%) & Top-5 ACC (\%)  \\
    \midrule
      Supervised (K400)  &\bf 78.30 & 94.24 &\bf 34.86 & 71.62 \\
    \midrule
      V-SwAV$_{\rho=2}$ \cite{feichtenhofer_large-scale_2021} & 64.84 & 87.52 & 25.56 & 63.38   \\
      V-BYOL$_{\rho=2}$ \cite{feichtenhofer_large-scale_2021} & 70.74 & 89.14 & 27.77 & 67.94 \\
      V-SimCLR$_{\rho=2}$ \cite{feichtenhofer_large-scale_2021} & 63.52 & 84.99 & 21.54 & 60.76 \\
      V-MoCo$_{\rho=2}$ \cite{feichtenhofer_large-scale_2021} & {73.72} & {92.18} & {29.47} & {68.49} \\
        \midrule
V-SwAV$_{\rho=2}$ + BOD-VCL &67.57 (\textcolor{mygreen}{$\uparrow 2.73$})&90.73 (\textcolor{mygreen}{$\uparrow 3.21$})&28.28 (\textcolor{mygreen}{$\uparrow 2.72$})&67.92 (\textcolor{mygreen}{$\uparrow 4.54$})\\
V-BYOL$_{\rho=2}$ + BOD-VCL &73.72 (\textcolor{mygreen}{$\uparrow 2.98$})&92.80 (\textcolor{mygreen}{$\uparrow 3.66$})&31.29 (\textcolor{mygreen}{$\uparrow 3.52$})&69.73 (\textcolor{mygreen}{$\uparrow 1.79$})\\
V-SimCLR$_{\rho=2}$ + BOD-VCL &66.05 (\textcolor{mygreen}{$\uparrow 2.53$})&88.51 (\textcolor{mygreen}{$\uparrow 3.52$})&24.72 (\textcolor{mygreen}{$\uparrow 3.18$})&62.27 (\textcolor{mygreen}{$\uparrow 1.51$})\\
V-MoCo$_{\rho=2}$ + BOD-VCL &76.67 (\textcolor{mygreen}{$\uparrow 2.95$})& \textbf{95.25} (\textcolor{mygreen}{$\uparrow 3.07$})&32.47 (\textcolor{mygreen}{$\uparrow 3.00$})& \textbf{73.38} (\textcolor{mygreen}{$\uparrow 4.89$})\\

    \bottomrule
    \end{tabular}
    \label{tab:eval_causal}
\end{table*}

\begin{table*}[t]
    \centering
    \caption{{Finetuning Results (average of 3 splits) for action classification on UCF101 and HMDB51. Self-supervised pretraining is done on Kinetics 400 datasets. All results without special notations are from the original article. Results with "$^\dagger$" are reproduced according to the open-source repositories. Results with "$^*$" utilize the ImageNet1K pre-trained weight. "-" means not mentioned in the original paper. As for modality, "V" indicates RGB video, "A" indicates audio, and "F" indicates optical flow. $\varrho$ is the number of positive samples.}}
    \label{tab:ucf}
    \resizebox{.9\linewidth}{!}{
    \begin{tabular}{c|c|c|c|c|c|c|c|c}
    \toprule
       Method  & Resolution & Frames  & Architecture & Param. & Epochs & Modality & UCF101 & HMDB51 \\
    \midrule
    SpeedNet \cite{benaim2020speednet} & 224$\times$224 & 16 & S3D-G & 9.1M & - &  V & 81.1 & 48.8 \\
    CoCLR \cite{han_self-supervised_2021} & 128$\times$128 & 16 & S3D-G  & 9.1M & 300 &  V + F & 90.6 & 62.9 \\
    V-BYOL$^\dagger_{\varrho=4}$  & 112$\times$112 & 16 & S3D-G  & 9.1M & 200 & V & 87.8 & 69.8 \\
    \midrule
        V-BYOL$_{\varrho=4}$ + BOD-VCL  & 112$\times$112 & 16 & S3D-G  & 9.1M & 200 & V &\bf 91.3 (\textcolor{mygreen}{$\uparrow 3.5$}) &\bf 71.4 (\textcolor{mygreen}{$\uparrow 1.6$}) \\
   \midrule
        VTHCL \cite{yang_video_2020} & 224$\times$224 & 8 & R3D-18 & 13.5M & 200 &  V & 80.6 & 48.6 \\
        TCLR \cite{dave_tclr_2022} & 112$\times$112 & 16 & R3D-18 & 13.5M & 100 &  V & 85.4 & 55.4 \\
        VideoMoCo \cite{videomoco} & 112$\times$112 & 16 & R3D-18 & 13.5M & 200 &  V & 74.1 & 43.6 \\
        SLIC \cite{khorasgani_slic_2022} & 128$\times$128 & 32 & R3D-18 & 13.5M & 150 & V & 83.2 & 52.2 \\
        MACLR \cite{avidan_maclr_2022} & 112$\times$112 & 32 & R3D-18 & 13.5M & 600 & V & 91.3 & 62.1 \\
        V-BYOL$^\dagger_{\varrho=4}$  & 112$\times$112 & 16 & R3D-18 & 13.5M & 200 & V & 88.3 & 69.3 \\
    \midrule
    V-BYOL$_{\varrho=4}$ + BOD-VCL & 112$\times$112 & 16 & R3D-18 & 13.5M & 200 & V &\bf 92.1 (\textcolor{mygreen}{$\uparrow 3.3$}) &\bf 71.2 (\textcolor{mygreen}{$\uparrow 1.9$})\\
    \midrule
        PacePred \cite{wang_self-supervised_2020} & 112$\times$112 & 16 & R(2+1)D-18 & 15.4M & 18 & V & 77.1 & 36.6 \\
        PacePred$^\dagger$ \cite{wang_self-supervised_2020} & 112$\times$112 & 16 & R(2+1)D-18 & 15.4M & 200 & V & 84.3 & 55.2 \\
        VideoMoCo \cite{videomoco} & 112$\times$112 & 16 & R(2+1)D-18 & 15.4M & 200 & V & 78.7 & 49.2 \\
        TCLR \cite{dave_tclr_2022} & 112$\times$112 & 16 & R(2+1)D-18 & 15.4M & 100 & V & 88.2 & 60.0 \\
        TCLR$^\dagger$  & 112$\times$112 & 16 & R(2+1)D-18 & 15.4M & 200 & V & 90.8 & 64.3 \\
        GDT \cite{GDT} & 112$\times$112 & 32 & R(2+1)D-18 & 15.4M & 200 & V + A & 89.3 & 60.0 \\
        V-BYOL$^\dagger_{\varrho=4}$  & 112$\times$112 & 16 & R(2+1)D-18  & 15.4M & 200 & V & 90.1 & 69.4 \\
    \midrule
    V-BYOL$_{\varrho=4}$ + BOD-VCL & 112$\times$112 & 16 & R(2+1)D-18 & 15.4M & 200 & V &\bf 93.0 (\textcolor{mygreen}{$\uparrow 1.9$}) &\bf 72.1 (\textcolor{mygreen}{$\uparrow 2.7$}) \\
    \midrule
    SeCo \cite{yao2021seco} & 112$\times$112 & 50 & R50+TSN & 25M & 400 &  V & 88.3 & 55.6 \\
    DSM \cite{wang_enhancing_2020} & 112$\times$112 & 16 & C3D & 27.7M & 200 & V & 70.3 & 40.5\\
    \midrule
        CVRL \cite{qian_spatiotemporal_2021} & 224$\times$224 & 32 & R3D-50 & 31.8M & 800 & V & 92.2 & 66.7 \\
        MACLR \cite{avidan_maclr_2022} & 224$\times$224 & 32 & R3D-50 & 31.8M & 600 & V & 94.0 & 67.4 \\
        MACLR \cite{avidan_maclr_2022} & 224$\times$224 & 32 & R3D-50 & 31.8M & 600 & V + F & 94.2 & 67.2 \\ 
        V-SimCLR$_{\varrho=2}$ \cite{feichtenhofer_large-scale_2021} & 224$\times$224 & 8 & R3D-50 & 31.8M & 200 & V & 88.9 & 67.2$^\dagger$ \\ 
        V-SwAV$_{\varrho=2}$ \cite{feichtenhofer_large-scale_2021} & 224$\times$224 & 8 & R3D-50 & 31.8M & 200 & V & 87.3 & 68.3$^\dagger$ \\ 
        V-MoCo$_{\varrho=4}$ \cite{feichtenhofer_large-scale_2021} & 224$\times$224 & 8 & R3D-50 & 31.8M & 200 & V & 93.5 & 71.6$^\dagger$ \\
        V-BYOL$_{\varrho=4}$ \cite{feichtenhofer_large-scale_2021} & 224$\times$224 & 8 & R3D-50 & 31.8M & 200 & V & 94.2 & 72.1 \\
        V-BYOL$_{\varrho=4}$ \cite{feichtenhofer_large-scale_2021} & 224$\times$224 & 16 & R3D-50 & 31.8M & 200 & V & 95.5 & 73.6 \\
    \midrule
        V-SimCLR$_{\varrho=2}$ + BOD-VCL & 224$\times$224 & 8 & R3D-50 & 31.8M & 200 & V & 90.3 (\textcolor{mygreen}{$\uparrow 1.4$}) & 70.4 (\textcolor{mygreen}{$\uparrow 3.2$})\\ 
        V-SwAV$_{\varrho=2}$ + BOD-VCL & 224$\times$224 & 8 & R3D-50 & 31.8M & 200 & V & 92.0 (\textcolor{mygreen}{$\uparrow 4.7$})& 72.6 (\textcolor{mygreen}{$\uparrow 4.3$})\\ 
        V-MoCo$_{\varrho=4}$ + BOD-VCL & 224$\times$224 & 8 & R3D-50 & 31.8M & 200 & V & 94.6 (\textcolor{mygreen}{$\uparrow 1.1$}) & 75.1 (\textcolor{mygreen}{$\uparrow 3.5$})\\
        V-BYOL$_{\varrho=4}$ + BOD-VCL & 224$\times$224 & 8 & R3D-50 & 31.8M & 200 & V & 95.3 (\textcolor{mygreen}{$\uparrow 1.1$}) & 74.5 (\textcolor{mygreen}{$\uparrow 2.4$})\\ 
        V-BYOL$_{\varrho=4}$ + BOD-VCL & 224$\times$224 & 16 & R3D-50 & 31.8M & 200 & V &\bf 96.5 (\textcolor{mygreen}{$\uparrow 1.0$}) &\bf 76.0 (\textcolor{mygreen}{$\uparrow 2.4$})\\ 
    \midrule
        VideoMAE \cite{tong_videomae_2022} & 224$\times$224 & 16 & ViT-B & 87M & 800 &  V & 96.1 & 73.3 \\
        MotionMAE \cite{yang_self-supervised_2022} & 224$\times$224 & 16 & ViT-B & 87M & 800 &  V & 96.1 & 73.3 \\
        MME \cite{sun_masked_2023} & 224$\times$224 & 16 & ViT-B & 87M & 800 &  V & 96.5 & 78.0 \\
        MGM \cite{MGM} & 224$\times$224 & 16 & ViT-B & 87M & 800 &  V  & 91.9 & 69.7 \\
        V-BYOL$^\dagger_{\varrho=4}$ & 224$\times$224 & 16 & ViT-B & 87M & 800 & V & 95.2 & 76.8 \\
    \midrule
        V-BYOL$_{\varrho=4}$ + BOD-VCL & 224$\times$224 & 16 & ViT-B & 87M & 800 & V &\bf 96.7 (\textcolor{mygreen}{$\uparrow 1.5$}) &\bf 78.3 (\textcolor{mygreen}{$\uparrow 1.5$}) \\ 
    \bottomrule
    \end{tabular}
    }
    \vspace{-0.1cm}
\end{table*}

\begin{table}[htb]
    \centering
    \caption{{The top-1 accuracy (\%) of action classification on motion-aware datasets Something-Something v2 and FineGym. The results are either from SEVERE \cite{severe} or the original paper. The results with $^\dagger$ are reproduced with open-source repositories. $\varrho$ is the number of positive samples.}}
    \label{tab:ssv2}
    \resizebox{.9\linewidth}{!}{
    \begin{tabular}{c|c|c|c}
    \toprule
    Method & backbone & SSv2 & Gym99 \\
    \midrule
    GDT \cite{GDT} & R(2+1)D-18 & 58.0 & 90.5 \\
    TCLR \cite{dave_tclr_2022} & R(2+1)D-18 & 59.8 & 91.6 \\
    TubeCo \cite{thoker_tubelet-contrastive_2023} & R(2+1)D-18 & 60.2 & 92.8 \\
    VideoMoCo \cite{videomoco} & R(2+1)D-18 & 59.0 & 90.3\\
    V-BYOL$^\dagger_{\varrho=4}$ & R(2+1)D-18 & 56.7 & 88.6\\
    \midrule
    V-BYOL$_{\varrho=4}$ + BOD-VCL & R(2+1)D-18 &\bf 61.4 (\textcolor{mygreen}{$\uparrow 4.7$}) &\bf 92.7 (\textcolor{mygreen}{$\uparrow 4.1$}) \\
    \midrule
    SVT \cite{ranasinghe_self-supervised_2022} & ViT-B & 59.2 & 62.3\\
    V-BYOL$^\dagger_{\varrho=4}$ & ViT-B & 58.7 & 61.4\\
    \midrule
    V-BYOL$_{\varrho=4}$ + BOD-VCL & ViT-B &\bf 63.9 (\textcolor{mygreen}{$\uparrow 5.2$}) &\bf 64.5 (\textcolor{mygreen}{$\uparrow 3.1$}) \\
    \bottomrule
    \end{tabular}
    }
    \label{tab:my_label}
\end{table}

\begin{table}[htb]
    \centering
    \caption{{Linear evaluation results on the Kinetics-400 dataset, where $\varrho$, $L$, and $\delta$ denote the number of positive samples, clip length, and stride, respectively.}}
    \resizebox{0.8\linewidth}{!}{
    \begin{tabular}{c|c|c|c|c}
    \toprule
    Method & $\varrho$ & $L$ & $\delta$ & Top-1 Acc. (\%) \\
    \midrule
    supervised & - & 8 & 8 & 74.7 \\
    \midrule
    V-SimCLR \cite{feichtenhofer_large-scale_2021} & 2 & 8 & 8 & 60.5 \\
    V-BYOL \cite{feichtenhofer_large-scale_2021} & 2 & 8 & 8 & 65.8 \\
    V-MoCo \cite{feichtenhofer_large-scale_2021} & 2 & 8 & 8 & 65.8 \\
    V-SwAV \cite{feichtenhofer_large-scale_2021} & 2 & 8 & 8 & 61.6 \\
    \midrule
    V-SimCLR + BOD-VCL & 2 & 8 & 8 & 63.8 (\textcolor{mygreen}{$\uparrow 3.3$}) \\
    V-BYOL + BOD-VCL & 2 & 8 & 8 & 67.9 (\textcolor{mygreen}{$\uparrow 2.1$}) \\
    V-MoCo + BOD-VCL & 2 & 8 & 8 & \bf 68.6 (\textcolor{mygreen}{$\uparrow 2.9$})\\
    V-SwAV + BOD-VCL & 2 & 8 & 8 & 64.2 (\textcolor{mygreen}{$\uparrow 2.6$}) \\
    \midrule
    V-MoCo \cite{feichtenhofer_large-scale_2021} & 4 & 8 & 8 & 67.8 \\
    V-BYOL \cite{feichtenhofer_large-scale_2021} & 4 & 8 & 8 & 68.9 \\
    V-MoCo \cite{feichtenhofer_large-scale_2021} & 2 & 16 & 4 & 67.6 \\
    V-BYOL \cite{feichtenhofer_large-scale_2021} & 4 & 16 & 4 & 71.5 \\
    \midrule
    V-MoCo + BOD-VCL & 4 & 8 & 8 & 70.2 (\textcolor{mygreen}{$\uparrow 2.4$})\\
    V-BYOL + BOD-VCL & 4 & 8 & 8 & 71.7 (\textcolor{mygreen}{$\uparrow 2.8$}) \\
    V-MoCo + BOD-VCL & 2 & 16 & 4 & 69.4 (\textcolor{mygreen}{$\uparrow 1.9$})\\
    V-BYOL + BOD-VCL & 4 & 16 & 4 & \bf 73.5 (\textcolor{mygreen}{$\uparrow 2.0$}) \\
    \bottomrule
    \end{tabular}
    }
    \label{tab:k400}
\end{table}

\section{Experiments}
\label{sec:exp}
\subsection{Experimental Setup}
\label{subsec:setup}
\textbf{Datasets}. We use five action recognition datasets: Kinetics400 \cite{kay2017kinetics}, UCF101 \cite{soomro2012ucf101}, HMDB51 \cite{kuehne2011hmdb}, Something-Something v2 \cite{ssv2}(SSv2) and FineGym \cite{gym99} (Gym99). Specifically, Kinetics400 is a large-scale dataset with 24K realistic action videos from 400 action categories, which is one of the largest and most widely used datasets for video pre-trained models. UCF101 and HMDB51 are two standard datasets for evaluating the performance of action recognition. SSv2 and FineGym are two motion-aware datasets that encompass a wider variety of action labels. We also use one action detection dataset AVA \cite{gu2018ava}. AVA is a large-scale video dataset specifically designed for spatiotemporally localized action detection.

\noindent\textbf{Networks}. We implement the feature extractor $f_\theta$ with Slow-R50 \cite{feichtenhofer_slowfast_2019} by default. We also integrate BOD-VCL with the best performance V-CL method, V-BYOL on other networks, namely S3D-G \cite{s3d-g}, R3D-18\cite{R3d18}, R(2+1)D-18 \cite{r2+1d}, and ViT-B \cite{timesformer}, for a fair comparison with prior works.

\noindent\textbf{Pretraining Setting}. We follow the pretraining procedure in \cite{feichtenhofer_large-scale_2021} and pre-train the backbone networks on the Kinetics 400 \cite{kay2017kinetics} dataset. We integrate our method with 4 video contrastive learning methods, namely V-MoCo, V-BYOL, V-SimCLR, and V-SwAV. The projection MLP output dimensions are set to 128 \cite{he_momentum_2020,chen_simple_2020, caron_unsupervised_2021, grill_bootstrap_2020}. The default epoch for pre-training is set to 200, and the total training hours for V-SimCLR + BOD-VCL, V-BYOL + BOD-VCL, V-MoCo + BOD-VCL, and V-SwAV + BOD-VCL are 24.6, 30.8, 22.5, and 27.2 hours, respectively, on 16 NVIDIA V100 GPUs. The projection head $f_p$ for V-MoCo, V-SimCLR, and V-SwAV is a 3-layer MLP with a hidden dimension of 2048. The projection and prediction MLPs for V-BYOL have 2 layers with a hidden dimension of 4096. The temperature parameter $\alpha=0.1$ for V-SimCLR, V-MoCo, and V-SwAV. We apply SGD with LARS \cite{you2017large} as the optimizer. The synchronized batch normalization is used for all methods except V-MoCo.

\subsection{Evaluation on Static and Dynamic Semantics}
We analyze whether our proposed BOD-VCL effectively captures both the static and dynamic semantics of videos. Specifically, following the experimental setup described in Section \ref{subsec:problem}, {we validate BOD-VCL on the Static UCF-101 and Dynamic UCF-101 datasets.} We integrate BOD-VCL with four V-CL methods: V-SwAV, V-BYOL, V-SimCLR, and V-MoCo. The experimental results are presented in Table \ref{tab:eval_causal}.

\noindent\textbf{Results}. 
{From the results presented in Table \ref{tab:eval_causal}, we observe that V-MoCo achieves accuracy comparable to supervised methods on both the Static UCF-101 and Dynamic UCF-101 datasets. Additionally, other V-CL methods, when pre-trained using the BOD-VCL approach, demonstrate an average improvement of 2.8\% in Top-1 ACC and 3.4\% in Top-5 ACC on Static UCF-101, as well as an average improvement of 3.1\% in Top-1 ACC and 3.2\% in Top-5 ACC on Dynamic UCF-101. These findings indicate that our proposed BOD-VCL method can enhance the learning of both static and dynamic features. Therefore, we can conclude that our approach effectively mitigates the issue present in Proposition \ref{prop:1}.}

\subsection{Action Recognition}
\noindent\textbf{Evaluation Protocols}. After pre-training, in accordance with the standard evaluation protocols \cite{feichtenhofer_large-scale_2021, qian_spatiotemporal_2021, dave_tclr_2022, severe}, we perform a supervised fine-tuning of the feature extractor on downstream datasets. During this process, the entire set of weights for the feature extractor is trained with supervision, complemented by an additional classification layer. The results of UCF101 and HMDB51 are presented in Table \ref{tab:ucf}. To ensure a fair comparison, we evaluate our method on commonly utilized networks and provide details regarding the input resolution and the number of frames used. The results of Something-Something v2 and FineGym are presented in Table \ref{tab:ssv2} and follow the same setup as in the SEVERE benchmark \cite{severe}. 
We then provide linear evaluation results on the Kinetics-400 dataset, where the weights of the feature extractor are frozen, and an additional linear classification layer is trained on top. The results with different positive samples, clip lengths, and strides are presented in Table \ref{tab:k400}. During inference, we uniformly sample three clips from a video and average the softmax scores for the prediction.

\noindent\textbf{Results}. 
From Table \ref{tab:ucf}, we can observe that under the same number of epochs and the same resolution and frame length, our proposed BOD-VCL consistently improves the accuracies of the V-CL methods and outperforms the prior works under the same experimental setting. Our methods improved the performance of V-CL methods on UCF101 with an average increase of 2.1\% and an average improvement of 2.7\% on HMDB51. From Table \ref{tab:ssv2}, we can observe that the original V-BYOL performs poorly compared to prior works, while V-BYOL + BOD-VCL significantly improves the performance of V-BYOL on both SSv2 and Gym99. Specifically, when using ViT-B, the V-BYOL + BOD-VCL improved the performance of V-BYOL in SSv2 by 5.6\%. This demonstrates the effectiveness of our proposed method when learning dynamic semantics. From Table \ref{tab:k400}, we can observe that under different settings, our proposed V-CL + BOD-VCL consistently improves the performance of the corresponding V-CL method by an average of 2.5\%.

\subsection{Action Detection}
\label{sec:detect}

\textbf{Dataset} We evaluate the performance of BOD-VCL on the AVA \cite{gu2018ava} dataset, an action detection dataset taken from 437 movies, with bounding box annotations for spatiotemporal localization of human actions.

\noindent\textbf{Implementation and evaluation protocol}. We follow the standard protocol \cite{feichtenhofer_large-scale_2021,feichtenhofer_slowfast_2019} to train a detector that is similar to Faster R-CNN \cite{ren2015faster} with minimal modifications adapted for video. The pre-trained network is used as the backbone. The region-of-interest (RoI) features \cite{girshick2015fast} are extracted at the last layers of the backbone network and extend into 3D by replicating it along the temporal axis. The region proposals are computed with an off-the-shelf person detector, which is pre-trained with Detectron \cite{Detectron2018}. We report the detection performance with mean Average Precision (mAP) over 60 classes with a frame-level IoU threshold of 0.5.

\begin{table}[htb]
    \centering
    \caption{{Action detection results on AVA datasets}}
    \begin{tabular}{c|c|c}
    \toprule
       Method  & mAP & (+BOD-VCL) \\
    \midrule
       V-SimCLR  & 17.6 & 21.2 (\textcolor{mygreen}{$\uparrow 3.6$}) \\
       V-BYOL & 23.4 &\bf 26.2 (\textcolor{mygreen}{$\uparrow 2.8$}) \\
       V-MoCo & 20.3 & 23.5 (\textcolor{mygreen}{$\uparrow 3.2$}) \\
       V-SwAV & 18.2 & 21.6 (\textcolor{mygreen}{$\uparrow 3.4$}) \\
    \bottomrule
    \end{tabular}
    \label{tab:ava}
\end{table}

\noindent\textbf{Results} The results of action detection is illustrated in Table \ref{tab:ava}. From the results, our proposed BOD-VCL improved the performance of V-CLs by an average of 3.2\%. These results demonstrate that the representations learned through our method enable the detector to better capture the information of physical entities and to detect the types of action. This suggests that our model has learned to capture both dynamic and static semantics.

\begin{figure*}
    \centering
    \subfigure[]{
        \includegraphics[width=0.2\textwidth]{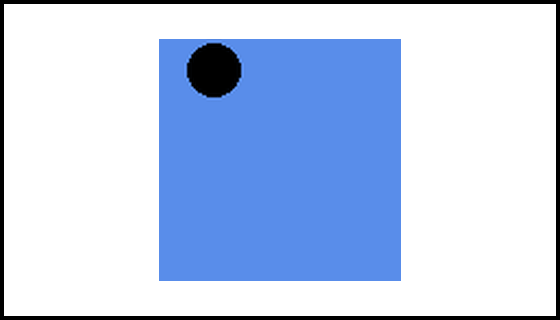}
    }
    \subfigure[]{
        \includegraphics[width=0.2\textwidth]{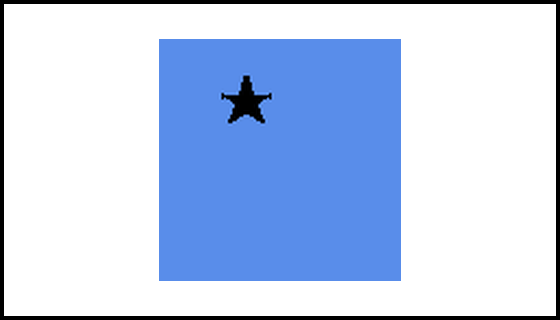}
    }
    \subfigure[]{
        \includegraphics[width=0.2\textwidth]{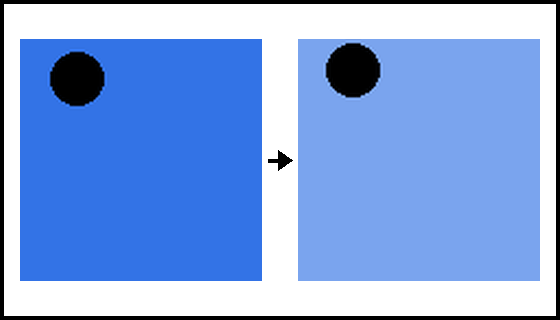}
    }
    \subfigure[]{
        \includegraphics[width=0.2\textwidth]{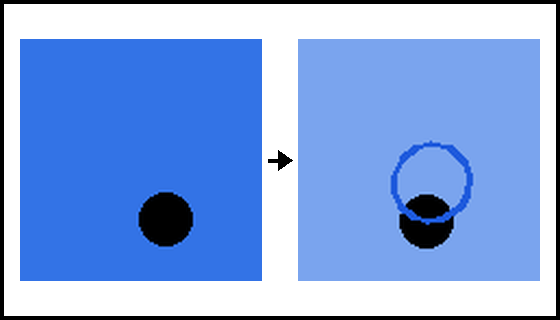}
    }
    \subfigure[]{
        \includegraphics[width=0.2\textwidth]{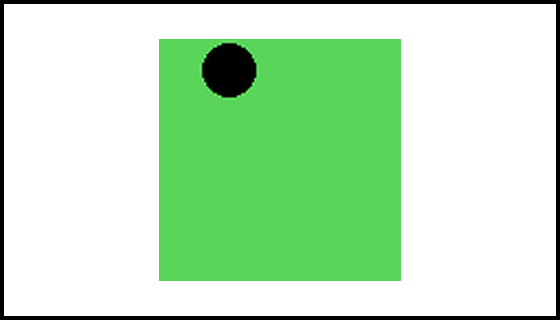}
    }
    \subfigure[]{
        \includegraphics[width=0.2\textwidth]{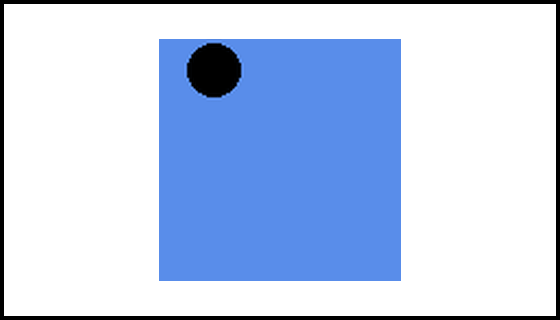}
    }
    \subfigure[]{
        \includegraphics[width=0.2\textwidth]{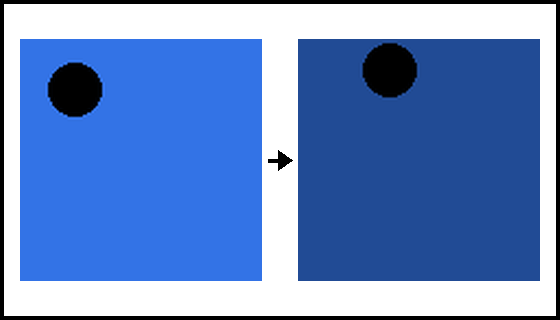}
    }
    \subfigure[]{
        \includegraphics[width=0.2\textwidth]{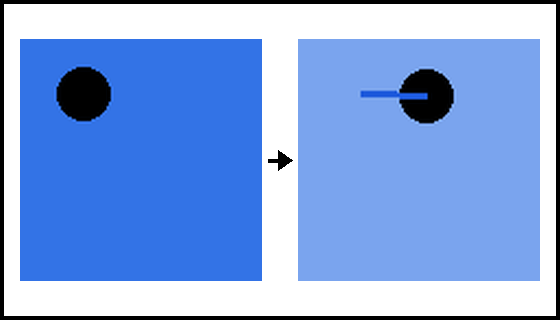}
    }
    \caption{{Examples from the constructed synthetic dataset. (a) and (e) represent simple static semantics defined by background color (blue and green). (b) and (f) show complex static semantics defined by foreground shape (pentagon and circle). (c) and (g) illustrate simple dynamic semantics defined by background brightness change (brightening and darkening). (d) and (h) correspond to complex dynamic semantics defined by foreground motion trajectory (rotation and linear motion).}}
    \label{fig:synth_examples}
\end{figure*}

\begin{figure*}
    \centering
    \subfigure[]{
    \includegraphics[width=0.3\textwidth]{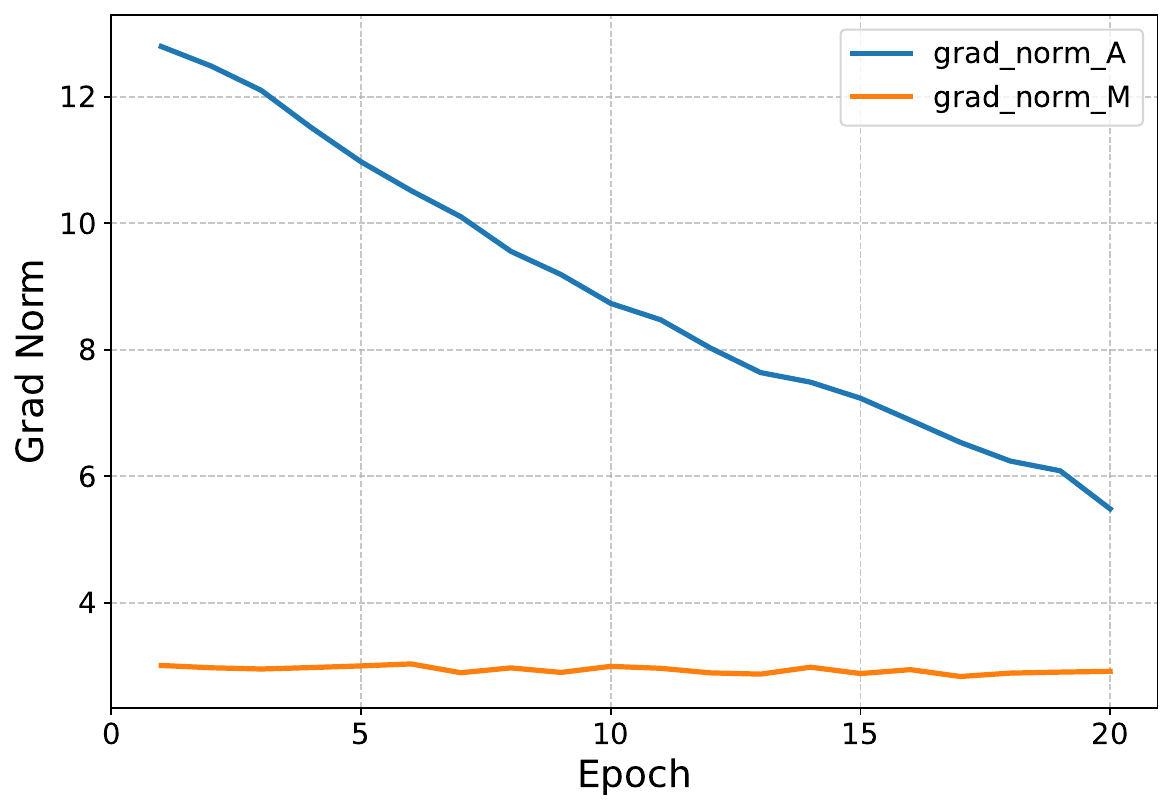}
    \label{fig:grad_1}
    }
    \subfigure[]{
    \includegraphics[width=0.3\textwidth]{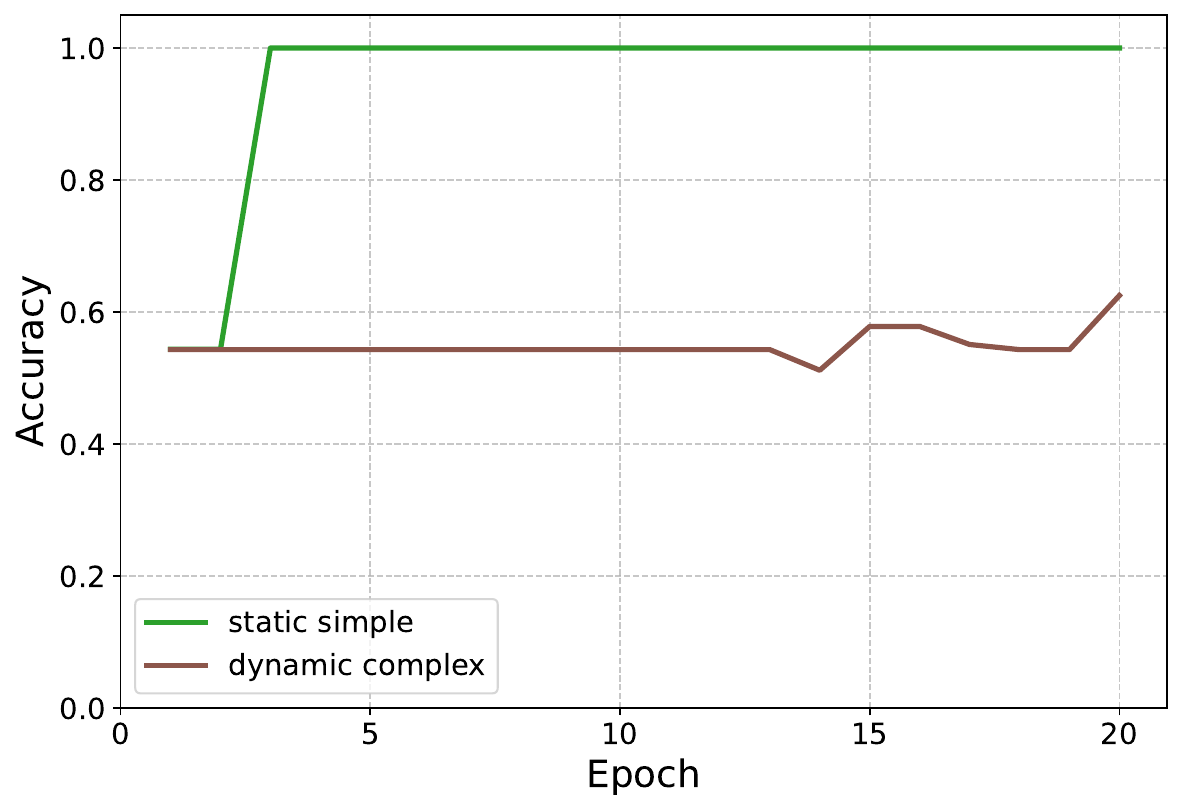}
    \label{fig:acc_1}
    }
    \subfigure[]{
    \includegraphics[width=0.3\textwidth]{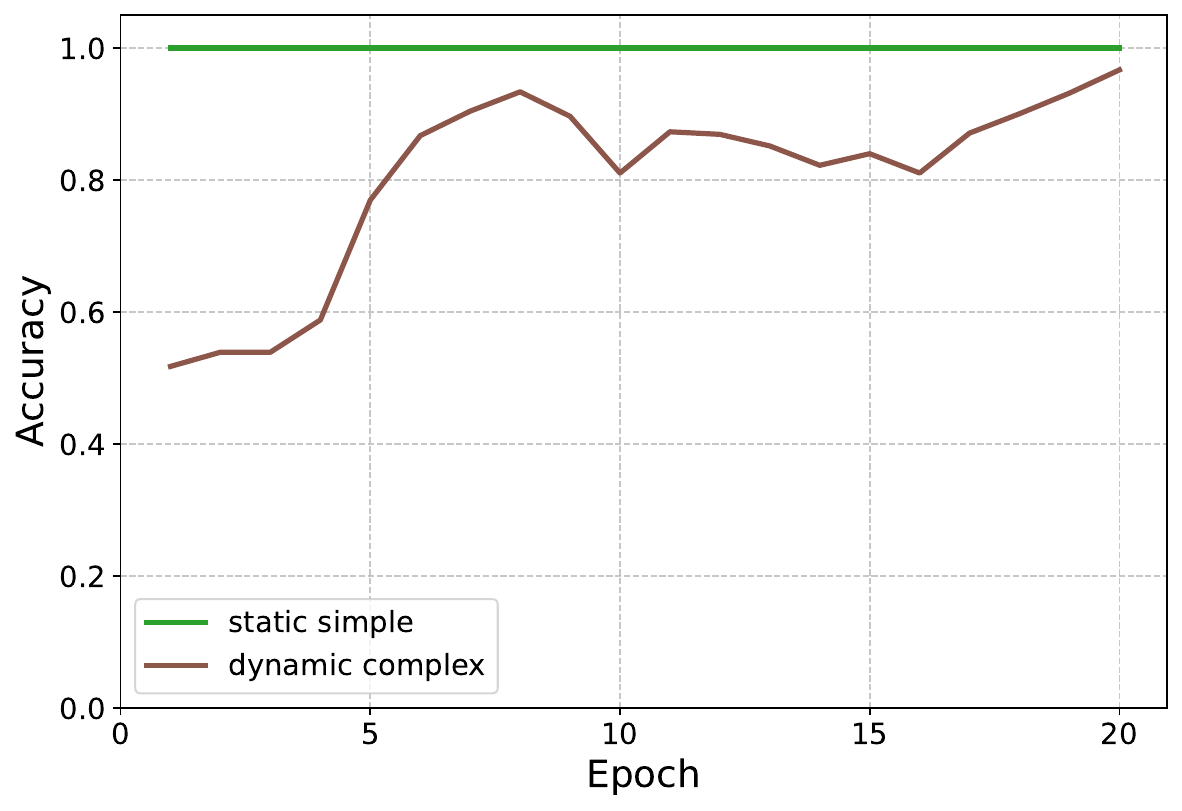}
    \label{fig:acc_1_dec}
    }
    \subfigure[]{
    \includegraphics[width=0.3\textwidth]{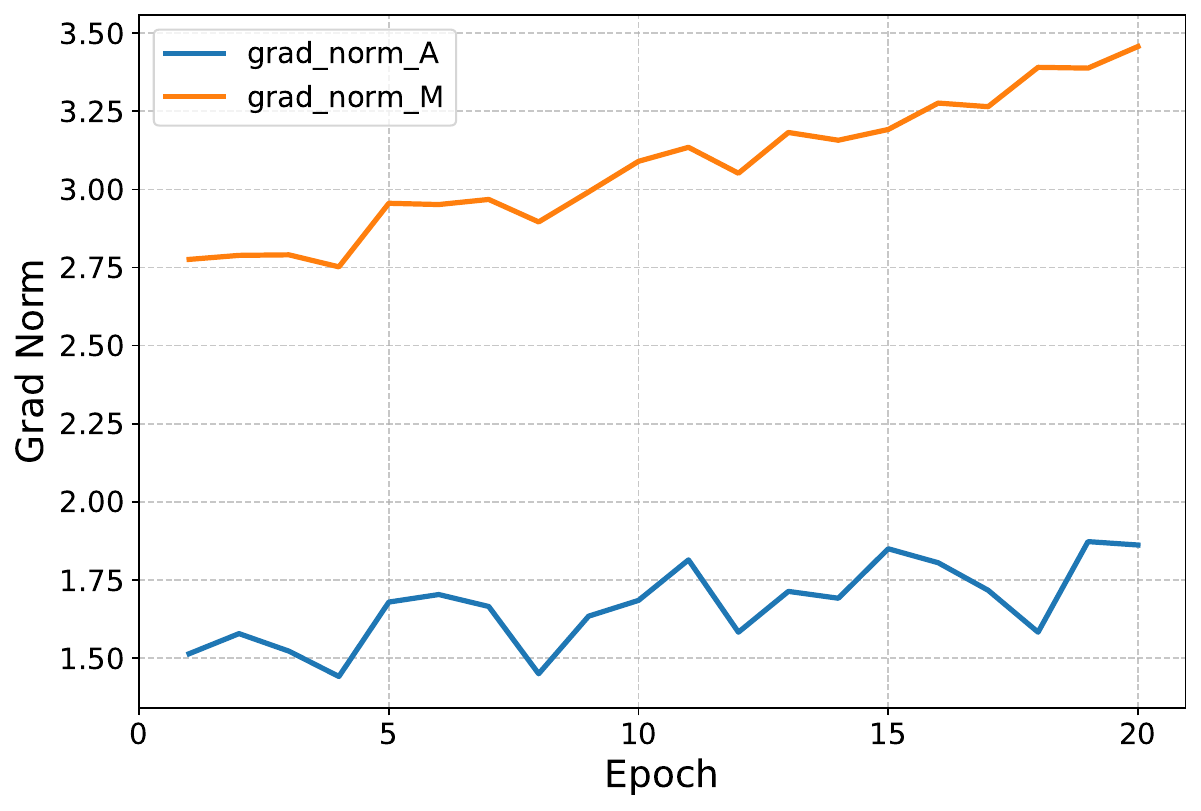}
    \label{fig:grad_2}
    }
    \subfigure[]{
    \includegraphics[width=0.3\textwidth]{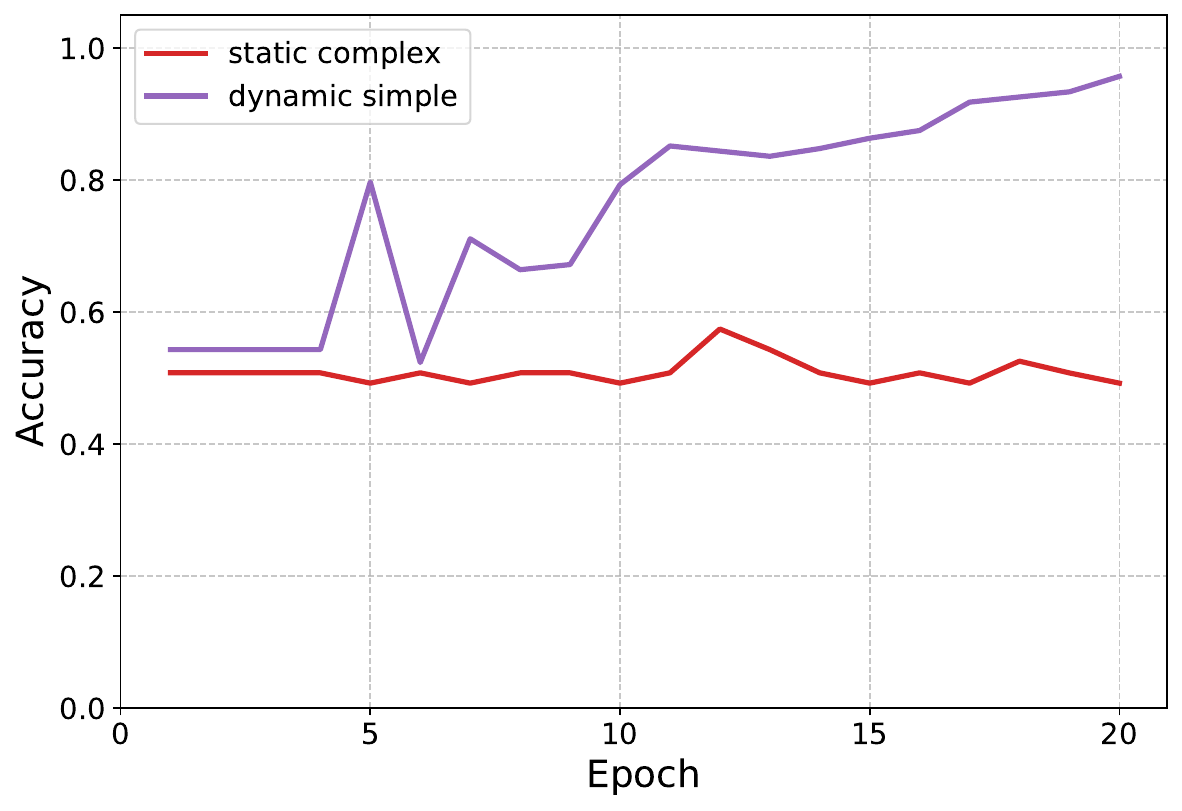}
    \label{fig:acc_2}
    }
    \subfigure[]{
    \includegraphics[width=0.3\textwidth]{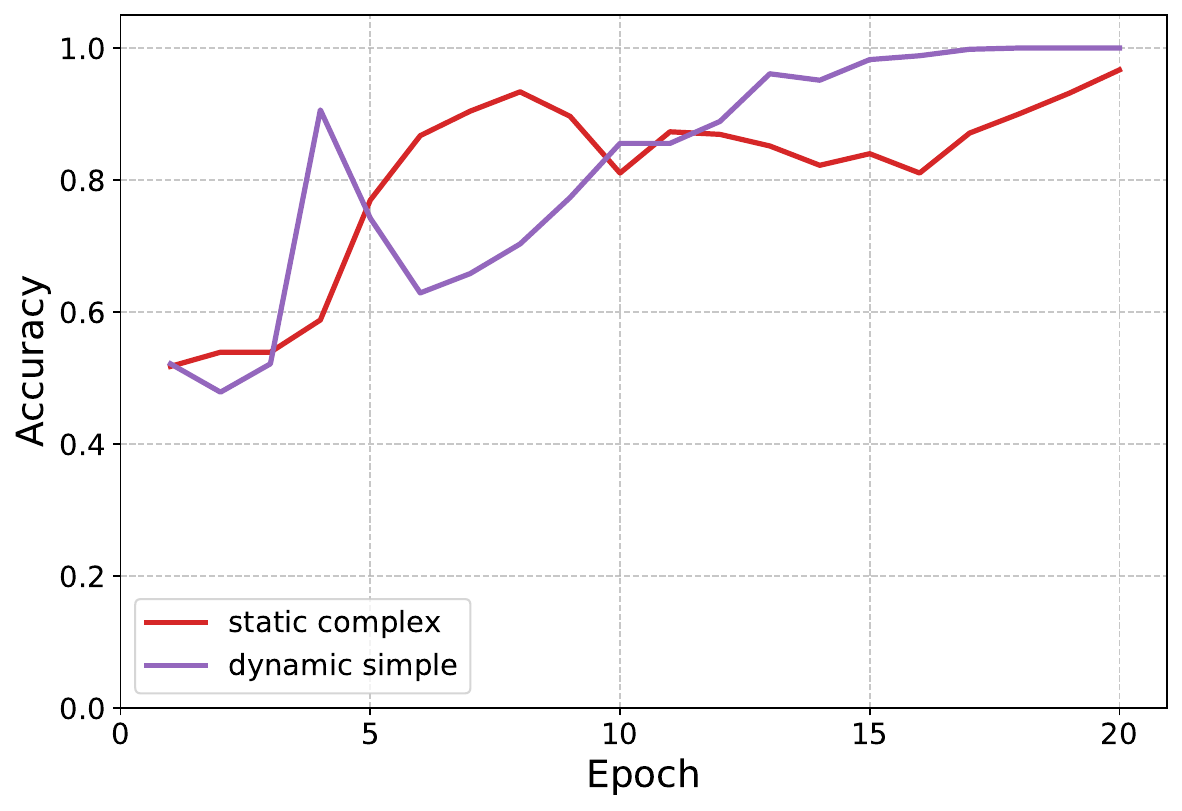}
    \label{fig:acc_2_dec}
    }
    \caption{{Gradient norms and linear evaluation accuracies under different correlation settings. (a) Gradient norms when simple static semantics are correlated with complex dynamic semantics. (b) Linear evaluation accuracy in the same setting. (c) Accuracy in the same setting when a decoupled loss is applied. (d) Gradient norms when simple dynamic semantics are correlated with complex static semantics. (e) Linear evaluation accuracy in the same setting. (f) Accuracy in the same setting with a decoupled loss.}}
    \label{fig:synth_results}
\end{figure*}

\begin{figure}
    \centering
    \includegraphics[width=0.9\linewidth]{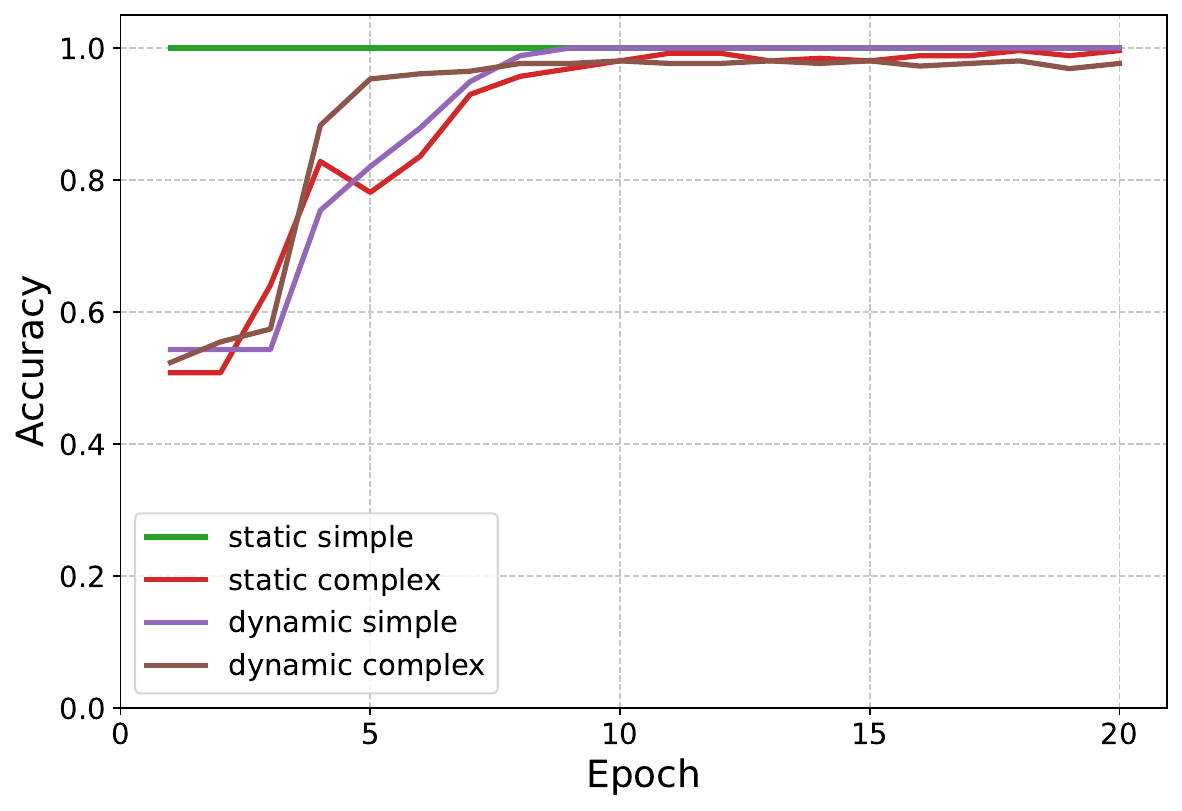}
    \caption{{Linear evaluation accuracies when no semantic correlations are present.}}
    \label{fig:acc_no_corr}
\end{figure}

\subsection{{Validation on Synthetic Dataset}}

\paragraph{{Experimental Setup.}}  {To empirically validate the theoretical conclusions established in Proposition~1 and Corollary~1, we design a controlled set of experiments based on a synthetic dataset. Direct verification on large-scale datasets such as Kinetics-400 is infeasible, since it is not possible to distinguish which parts of the raw video correspond to simple or complex semantics, nor to explicitly identify the parameters \(\theta_A\) and \(\theta_M\). To overcome these challenges, we construct a synthetic dataset where each video consists of four types of semantics: \textit{simple static} (background color: blue or green), \textit{complex static} (foreground shape: pentagon or circle), \textit{simple dynamic} (background brightness variation: brightening or darkening), and \textit{complex dynamic} (foreground motion trajectory: rotation or linear motion). We further design two correlation scenarios: (i) simple static correlated with complex dynamic, and (ii) simple dynamic correlated with complex static. To decouple the parameters associated with static and dynamic semantics, we adopt a two-stream architecture where one stream processes the first video frame to capture static features and the other processes the optical flow to capture dynamic features. The outputs of the two networks are summed to form the video representation. The model is pre-trained with the InfoNCE loss, and we record both linear evaluation accuracy for each semantic type and the gradient norms \(\|\nabla_{\theta_A}\mathcal{L}\|\) and \(\|\nabla_{\theta_M}\mathcal{L}\|\). Samples of the constructed dataset are illustrated in Figure~\ref{fig:synth_examples}.}

{\paragraph{Results under Correlated and Independent Settings.}  We first examine Proposition~1, which predicts that under correlated settings, the optimization process will favor the easier semantic component while suppressing the harder one. In the case of simple static correlated with complex dynamic semantics, the gradient norm of the static stream remains consistently larger than that of the dynamic stream, while the latter barely changes during training. Correspondingly, the classification accuracy on simple static semantics quickly converges to nearly 1.0, whereas the accuracy on complex dynamic semantics shows almost no improvement. Conversely, when simple dynamic semantics are correlated with complex static semantics, the gradient norm of the dynamic stream dominates throughout training, while the static stream remains nearly unchanged. As a result, the accuracy on simple dynamic semantics steadily increases, but the accuracy on complex static semantics stagnates. These findings, shown in Figure~\ref{fig:synth_results}, confirm the prediction of Proposition~1. As a comparison, when no correlations are present (Figure~\ref{fig:acc_no_corr}), the accuracies of all four semantics rapidly approach 1.0, confirming that each semantic can be learned in isolation and that the failure under correlation is not due to intrinsic difficulty but to optimization bias.}

{\paragraph{Effect of Decoupled Loss.} Finally, we investigate Corollary~1, which suggests that separating the similarity losses for static and dynamic semantics can mitigate the suppression effect. In both correlated settings, applying the decoupled loss balances the gradient updates: while the simple semantics still achieve near-perfect accuracy, the previously suppressed complex semantics now exhibit steady improvements, as shown in Figures~\ref{fig:acc_1_dec} and~\ref{fig:acc_2_dec}. These results provide direct empirical support for Corollary~1 by demonstrating that the decoupled loss effectively alleviates the imbalance observed in Proposition~1.}

\subsection{Visualization Results}
\begin{figure}[htb]
    \centering
    \subfigure[V-SimCLR on Diving-48]{%
        \includegraphics[width=.9\linewidth]{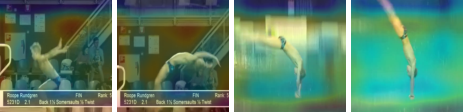}
        \label{fig:k400_simclr}
    }
    \subfigure[V-SimCLR + BOD-VCL on Diving-48]{%
        \includegraphics[width=.9\linewidth]{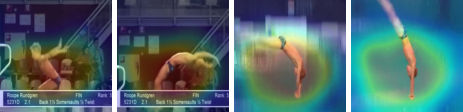}
        \label{fig:K400_ours}
    }
    \caption{{GradCAM visualizations of a video clip sampled from the Diving-48 dataset. (a) Visualization results using the V-SimCLR method, and (b) visualization results using the V-SimCLR + BOD-VCL method.}}
    \label{fig:gradcam}
\end{figure}

We utilize GradCAM \cite{selvaraju2017grad} to visualize the results of the V-SimCLR method and our proposed V-SimCLR + BOD-VCL approach. Specifically, we sample a {video clip} from the Diving-48 dataset, {showing a diver executing a dive}, process it through different pre-trained models {(Slow-R50 backbones)}, and obtain their visualization results, as shown in Figure \ref{fig:gradcam}. We observe that the model pre-trained with V-SimCLR tends to focus on background regions, whereas the model pre-trained with V-SimCLR + BOD-VCL consistently identifies the {athlete performing the dive}. This indicates that our method is not affected by confounding effects, thereby validating the effectiveness of our approach.

\subsection{Ablation Studies}
\label{subsec:ablation}
In this section, we conduct ablation experiments from the following aspects: (1) The performance under different pretraining epochs. (2) The impact of bi-level optimization. (3) The impact of different choices of bi-level optimization implementation, (4) The impact of different dimensions of Koopman Operator $M$, (5) The influence of hyperparameters.

\begin{figure}[tb]
    \centering
    \includegraphics[width=1\linewidth]{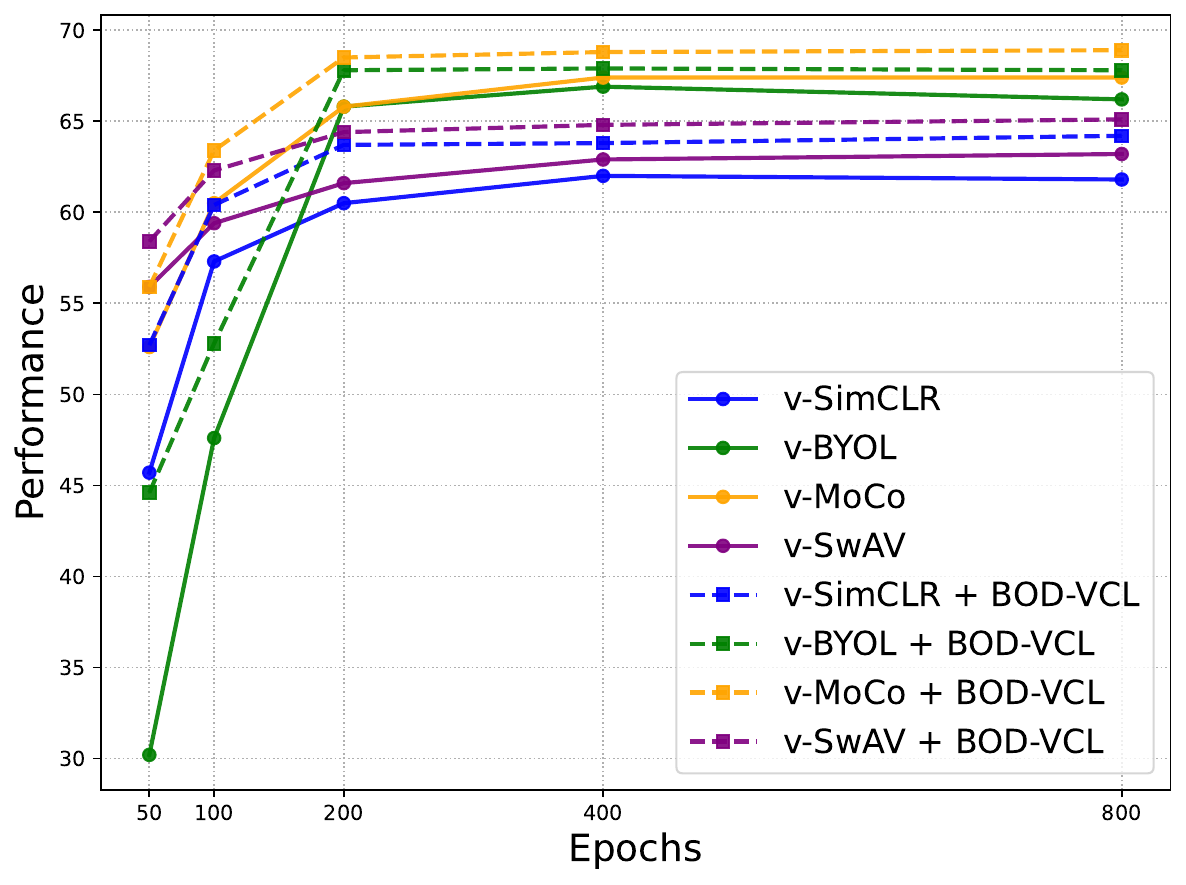}
    \caption{Linear evaluation on Kinetics 400 with different pre-training epochs. Solid lines represent the original V-CL methods. Dashed lines represent V-CL methods with BOD-VCL. For all results, the number of positive samples $\varrho=2$ and frame length $L=8$, stride $\delta=8$.}
    \label{fig:epochs}
\end{figure}

\paragraph{Ablation on pre-training epochs.}
We conduct an ablation study on the impact of longer pre-training epochs in Figure \ref{fig:epochs}, comparing the linear evaluation performance on the Kinetics-400 dataset. From Figure \ref{fig:epochs}, we can observe that as the number of pre-training epochs increases, the performance of V-CL + BOD-VCL consistently outperforms V-CL. While a longer training duration generally leads to improved performance, the results for 800 epochs show no significant gain over those for 200 epochs ($<1\%$). Moreover, our proposed V-CL + BOD-VCL converges faster than the standard V-CL methods, demonstrating the effectiveness of our proposed method.

\paragraph{Ablation on decoupled contrastive learning}
{
We further investigate the impact of incorporating $\mathcal{L}_{pred}$ into the training objective. 
Specifically, we replace the decoupled V-CL objective in Equation (\ref{eq:bi-level}) with the standard V-CL objective and examine whether adding $\mathcal{L}_{pred}$ alone is detrimental. Table \ref{tab:add_ablation} reports the linear evaluation performance on Kinetics-400 and the fine-tuning performance on UCF-101 and HMDB-51. We include V-SimCLR,  V-SwAV, V-BYOL, and V-MoCo for a more comprehensive comparison.  
\begin{table}[htb]
    \centering
    \caption{{Comparison results for linear evaluation on Kinetics-400 and fine-tuning on UCF-101 and HMDB-51, with or without $\mathcal{L}_{pred}$. ``+ $\mathcal{L}_{pred}$'' denotes directly adding the prediction loss to the standard contrastive objective, while ``+ BOD-VCL'' denotes our full bilevel decoupled formulation.}}
    \begin{tabular}{l|c|c|c}
    \toprule
       Method  & K400 & UCF-101 & HMDB-51 \\
    \midrule
        V-SwAV$_{\rho=2}$ & 61.6 & 87.3 & 68.3 \\
        \quad + $\mathcal{L}_{pred}$ & 62.7 & 88.4 & 69.1 \\
        \quad + BOD-VCL & \textbf{64.2} & \textbf{92.0} & \textbf{72.6} \\
    \midrule
        V-BYOL$_{\rho=2}$ & 65.8 & 90.2 & 70.6 \\
        \quad + $\mathcal{L}_{pred}$ & 66.3 & 91.1 & 71.2 \\
        \quad + BOD-VCL & \textbf{67.9} & \textbf{93.4} & \textbf{73.5} \\
    \midrule
        V-SimCLR$_{\rho=2}$ & 60.5 & 88.9 & 67.2 \\
        \quad + $\mathcal{L}_{pred}$ & 61.2 & 89.7 & 68.1 \\
        \quad + BOD-VCL & \textbf{63.8} & \textbf{90.3} & \textbf{70.4} \\
    \midrule
        V-MoCo$_{\rho=2}$ & 65.8 & 90.1 & 70.2 \\
        \quad + $\mathcal{L}_{pred}$ & 67.1 & 91.7 & 71.6 \\
        \quad + BOD-VCL & \textbf{68.6} & \textbf{92.8} & \textbf{73.8} \\
    \bottomrule
    \end{tabular}
    \label{tab:add_ablation}
\end{table}
From Table \ref{tab:add_ablation}, we observe that simply adding $\mathcal{L}_{pred}$ does \emph{not} degrade performance. 
On the contrary, across V-SwAV, V-BYOL, V-SimCLR, and V-MoCo, adding $\mathcal{L}_{pred}$ consistently brings slight improvements compared to their vanilla counterparts. This empirically confirms that minimizing $\mathcal{L}_{pred}$ does not discard useful information. Furthermore, our full bilevel decoupled formulation (BOD-VCL) significantly outperforms both the original contrastive objectives and their ``$+\mathcal{L}_{pred}$'' variants, validating the effectiveness of disentangling static and dynamic semantics. }


\paragraph{Ablation on Bi-level Optimization.}
We also conduct an ablation study by replacing the bi-level optimization problem in Equation (\ref{eq:bi-level}) with a single-level optimization problem. Where $\mathcal{L}_{pred}$, $\mathcal{L}_{A}$ and $\mathcal{L}_{M}$ are minimized simultaneously. The results of linear evaluation on Kinetics400 are demonstrated in Table \ref{tab:single_level}, where we denote the results with $^*$. By comparing the results in Table \ref{tab:single_level} with those in Table \ref{tab:k400}, we can observe that the results of V-CL$^*$ still outperform the corresponding V-CL methods, but are lower than the results achieved by V-CL + BOD-VCL. This phenomenon underscores the necessity of bi-level optimization and reveals the correctness of our analysis, as discussed in Section \ref{subsec:objective}.

\begin{table}[t]
    \centering
    \caption{Linear evaluation on Kinetics400 without using bi-level optimization. The results with $^*$ denotes minimizing the $\mathcal{L}_{pred}$, $\mathcal{L}_{A}$ and $\mathcal{L}_{M}$ simultaneously. The pre-training epoch is 200, with $\varrho=2,L=8,\delta=8$.}
    \label{tab:single_level}
    \resizebox{.9\linewidth}{!}{
    \begin{tabular}{c|c|c|c|c}
    \toprule
       Method  & V-SimCLR$^*$ & V-BYOL$^*$ & V-MoCo$^*$ & V-SwAV$^*$ \\
    \midrule
    Top-1 Acc. (\%) & 61.4 & 66.7 & 66.2 & 63.5 \\
    \bottomrule
    \end{tabular}
    }
\end{table}

\paragraph{Different implementations of bi-level optimization.}
Multilevel optimization (MLO) tackles nested optimization scenarios, where lower-level optimization problems constrain upper-level ones in a hierarchical manner. MLO serves as a unified mathematical framework for various applications, including meta-learning\cite{finn_model-agnostic_2017,rajeswaran_meta-learning_2019}, neural architecture search \cite{liu2018darts}, and reinforcement learning\cite{rajeswaran2020game}. Numerous optimization algorithms have been proposed for solving MLO, with gradient-based approaches being particularly popular. These approaches leverage best-response Jacobians, computed using methods like iterative differentiation (ITD) or approximate implicit differentiation (AID), applying the chain rule.
We implement our bi-level optimization algorithm using the \textit{BETTY} library \cite{choe_betty_2022}, designed for large-scale multilevel optimization (MLO). Specifically, we consider four MLO algorithms from \textit{BETTY}, namely ITD-RMAD \cite{finn_model-agnostic_2017}, AID-NMN \cite{lorraine_optimizing_2020}, AID-CG \cite{rajeswaran_meta-learning_2019}, and AID-FD \cite{liu2018darts}. The details about these algorithms are as follows:
\begin{itemize}
    \item \textbf{ITD-RMAD} \cite{finn_model-agnostic_2017} applies the implicit function theorem to the lower-level optimization problem and computes the gradients of the upper-level objective with respect to the upper-level parameters using reverse-mode automatic differentiation.
    \item \textbf{AID-NMN} \cite{lorraine_optimizing_2020} approximates the inverse of the Hessian matrix of the lower-level objective using a truncated Neumann series expansion and computes the gradients of the upper-level objective with respect to the upper-level parameters using forward-mode automatic differentiation.
    \item \textbf{AID-CG} \cite{rajeswaran_meta-learning_2019} solves a linear system involving the Hessian matrix of the lower-level objective using the conjugate gradient algorithm and computes the gradients of the upper-level objective with respect to the upper-level parameters using forward-mode automatic differentiation.
    \item \textbf{AID-FD} \cite{liu2018darts} approximates the inverse of the Hessian matrix of the lower-level objective using a finite difference approximation and computes the gradients of the upper-level objective with respect to the upper-level parameters using forward-mode automatic differentiation.
\end{itemize}

\begin{figure}[htb]
    \centering
    \includegraphics[width=1\linewidth]{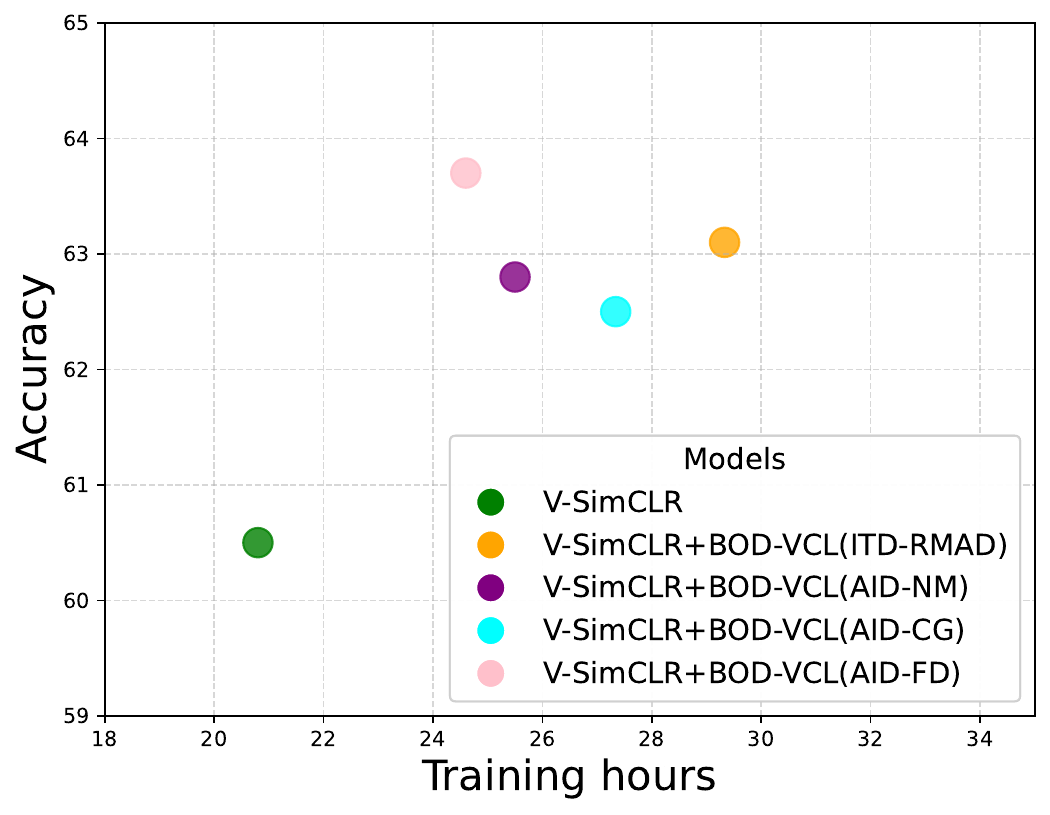}
    \caption{{Linear evaluation accuracy and training hours of V-SimCLR and V-SimCLR+BOD-VCL with various BLO implementations on the Kinetics400 dataset.}}
    \label{fig: BLO-implement}
\end{figure}

We conduct an ablation study on different implementations of bi-level optimization. For all results in this section, the experiments are conducted with the Slow-R50 as the feature extractor, the clip length $L=8$, sampling stride $\delta=8$, video resolution is set to $224\times224$, and the number of positive samples $\varrho=2$. The batch size is set to $64$. We evaluate four \textit{BETTY} algorithms, namely ITD-RMAD \cite{finn_model-agnostic_2017}, AID-NMN \cite{lorraine_optimizing_2020}, AID-CG \cite{rajeswaran_meta-learning_2019}, and AID-FD \cite{liu2018darts}. {Figure \ref{fig: BLO-implement} compares the time of total training duration for each algorithm and also the baseline, namely the V-SimCLR. }

As the results in Figure \ref{fig: BLO-implement} show, all methods with V-SimCLR+BOD-VCL outperform the original V-SimCLR method. Additionally, among the four MLO algorithms, the AID-FD has the shortest training duration while also having the best performance on the linear evaluation of Kinetics400. {In terms of GPU memory usage, V-SimCLR requires 298.8 GB, while V-SimCLR+BOD-VCL with ITD-RMAD, AID-NM, AID-CG, and AID-FD require 321.6 GB, 328.3 GB, 326.4 GB, and 307.2 GB, respectively.} Therefore, the default setting of the implementation of bi-level optimization is set to AID-FD.

\paragraph{Number of iterations in inner-loop.}
Our proposed BOD-VCL is a bi-level optimization procedure. The number of inner-loop update steps affects both the training duration and the performance. We present both the training time per iteration and the performance of linear evaluation on Kinetics 400, following the standard experimental setting in Subsection \ref{subsec:setup}. The results are illustrated in Table \ref{tab:innerloop}.

\begin{table}[htb]
    \centering
    \caption{The comparison results on different inner-loop steps. In the table, the time consumption of each training iteration and the linear evaluation accuracy on Kinetics 400 are provided.}
    \label{tab:innerloop}
    \begin{tabular}{c|c|c}
    \toprule
    Inner-loop steps  & Time/iter & Acc. (\%) \\
    \midrule
        1 &\bf 1.39s & 63.8 \\
        5 & 2.36s & 64.0 \\
        10 & 4.24s &\bf 64.0 \\
    \bottomrule
    \end{tabular}
\end{table}

As the results in Table \ref{tab:innerloop} show, we can observe that, as the number of inner-loop steps increases, the time consumption of the training iteration also significantly increases. Specifically, when each inner-loop is updated for 10 steps, the training iteration time is almost three times the time consumption of just one inner-loop step, while the linear evaluation accuracy on the Kinetics 400 dataset increases by only $0.2\%$. Therefore, for training efficiency, we use only one inner-loop step by default.

\paragraph{{Temporal distinctiveness of per-frame representations}}
{To further examine whether our implementation captures meaningful temporal dynamics, we assess the distinctiveness of frame-level representations on the UCF-101 validation set. Specifically, for each video, we pass a clip through a V-SimCLR + BOD-VCL pre-trained Slow-R50 encoder and extract per-frame features $\{h_t\}$ from the same forward pass by applying global average pooling over spatial dimensions while preserving the temporal dimension. We then compute the cosine similarity matrix among these features, where the $(i,j)$ entry is defined as
\begin{equation}
    \mathrm{Sim}(h_i,h_j) = \frac{h_i^\top h_j}{\|h_i\|\,\|h_j\|}.
\end{equation}
In addition, we report the average similarity as a function of temporal lag $k$, obtained by averaging the entries on the $k$-th super-diagonal of the similarity matrix. }

\begin{figure*}[t]
    \centering
    \subfigure[]{
    \includegraphics[width=0.48\textwidth]{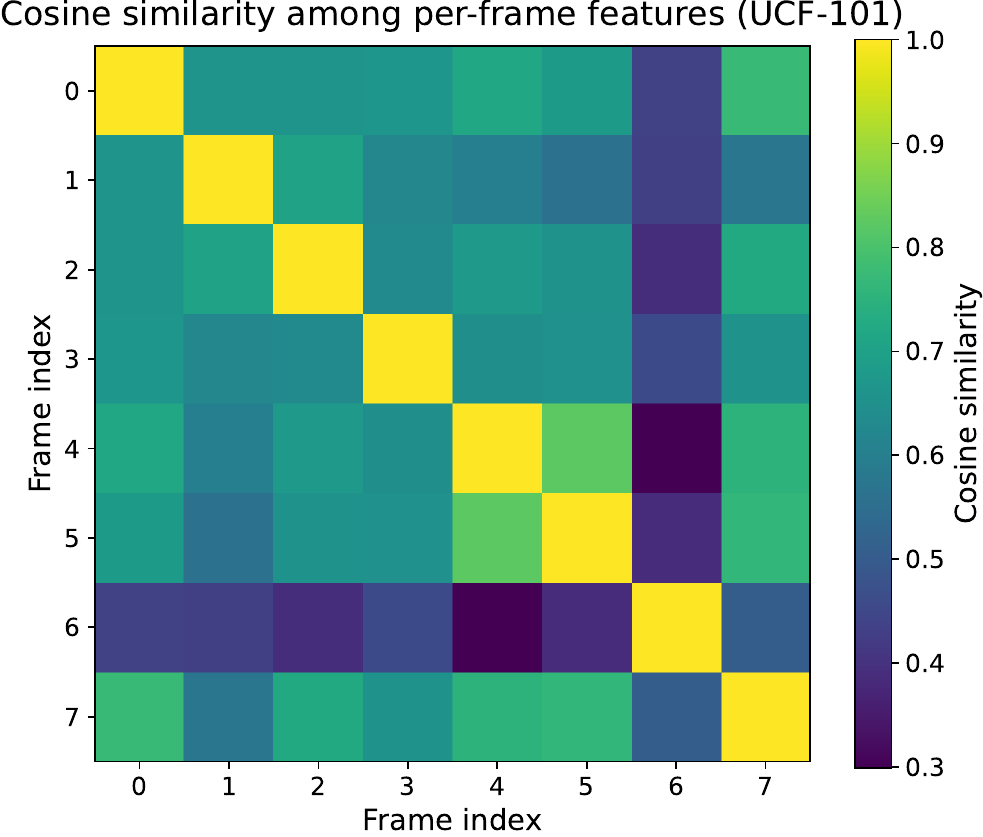}
    \label{fig:perframe_a}
    }
    \subfigure[]{
    \includegraphics[width=0.48\textwidth]{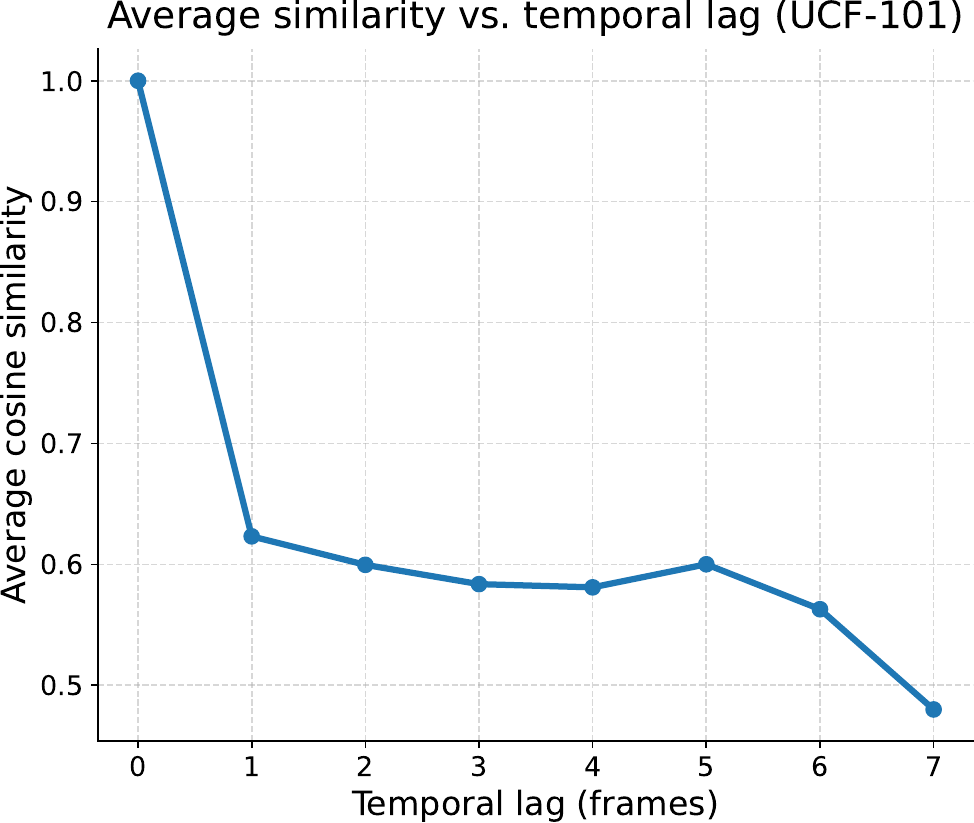}
    \label{fig:perframe_b}
    }
    \caption{{Temporal distinctiveness of per-frame features on the UCF-101 validation set. 
    (a) Cosine similarity matrix among per-frame features extracted from the same video. 
    (b) Average similarity as a function of temporal lag, aggregated over the validation set.} }
    \label{fig:per_frame_compare}
\end{figure*}
{The results in Figure~\ref{fig:per_frame_compare} demonstrate that the learned per-frame representations are temporally distinctive. Figure~\ref{fig:perframe_a} shows that the similarity matrix is diagonally dominant: adjacent frames yield high similarity, while similarity gradually decreases as the temporal gap increases. Figure~\ref{fig:perframe_b} further summarizes this trend across the dataset by showing that the average similarity consistently declines as the temporal lag grows. This confirms that the per-frame features are not identical across time steps, but instead encode temporally varying information. Such a property is consistent with prior observations in video action detection \cite{gu2018ava, li2020ava} and temporal localization \cite{zhang_actionformer_2022, lin2019bmn, xu2020gtad}, supporting that our implementation effectively captures temporal dynamics.}

{\paragraph{Strict version of BOD-VCL.} We further conducted an ablation experiment by implementing a ``strict version'' of BOD-VCL. In this setting, the computation of $\mathcal{L}_{ssl}$ remains unchanged, where the full 3D spatiotemporal convolutions are used for both forward and backward propagation. However, when optimizing $\mathcal{L}_{pred}$, we masked out the temporal dimension of the 3D kernels, retaining only the central slice, and fixed the temporal pooling kernel and stride to 1. This modification makes the forward process equivalent to a per-frame 2D convolution, ensuring that gradients flow exclusively to spatial parameters without accessing cross-frame information.}

\begin{table}[htb]
    \centering
    \caption{{Top-1 linear evaluation on Kinetics-400 and fine-tuning results on UCF-101 and HMDB-51. ``Strict'' denotes the ablation where $\mathcal{L}_{pred}$ is optimized with 2D-only convolutions (temporal kernels masked, temporal pooling stride fixed to 1), ensuring that gradients flow only to spatial parameters without accessing cross-frame information.}}
    \begin{tabular}{l|c|c|c}
    \toprule
       Method  & K400 & UCF-101 & HMDB-51 \\
    \midrule
        V-SwAV$_{\rho=2}$ & 61.6 & 87.3 & 68.3 \\
        \quad + BOD-VCL (Strict) & \textbf{64.2} & \textbf{92.1} & 72.3 \\
        \quad + BOD-VCL & \textbf{64.2} & 92.0 & \textbf{72.6} \\
    \midrule
        V-BYOL$_{\rho=2}$ & 65.8 & 90.2 & 70.6 \\
        \quad + BOD-VCL (Strict) & 67.5 & 93.3 & \textbf{73.6} \\
        \quad + BOD-VCL & \textbf{67.9} & \textbf{93.4} & 73.5 \\
    \midrule
        V-SimCLR$_{\rho=2}$ & 60.5 & 88.9 & 67.2 \\
        \quad + BOD-VCL (Strict) & 63.6 & \textbf{90.4} & 70.1 \\
        \quad + BOD-VCL & \textbf{63.8} & 90.3 & \textbf{70.4} \\
    \midrule
        V-MoCo$_{\rho=2}$ & 65.8 & 90.1 & 70.2 \\
        \quad + BOD-VCL (Strict) & \textbf{68.8} & 92.5 & 73.6 \\
        \quad + BOD-VCL & 68.6 & \textbf{92.8} & \textbf{73.8} \\
    \bottomrule
    \end{tabular}
    \label{tab:strict}
\end{table}
{The results in Table~\ref{tab:strict} show that compared to the full 3D $\mathcal{L}_{pred}$, the 2D-only strict version achieves very similar linear evaluation and fine-tuning performance across all benchmarks. This indicates the effectiveness of $\mathcal{L}_{pred}$ under the current implementation.}

\begin{figure}[tb]
    \centering
    \includegraphics[width=1\linewidth]{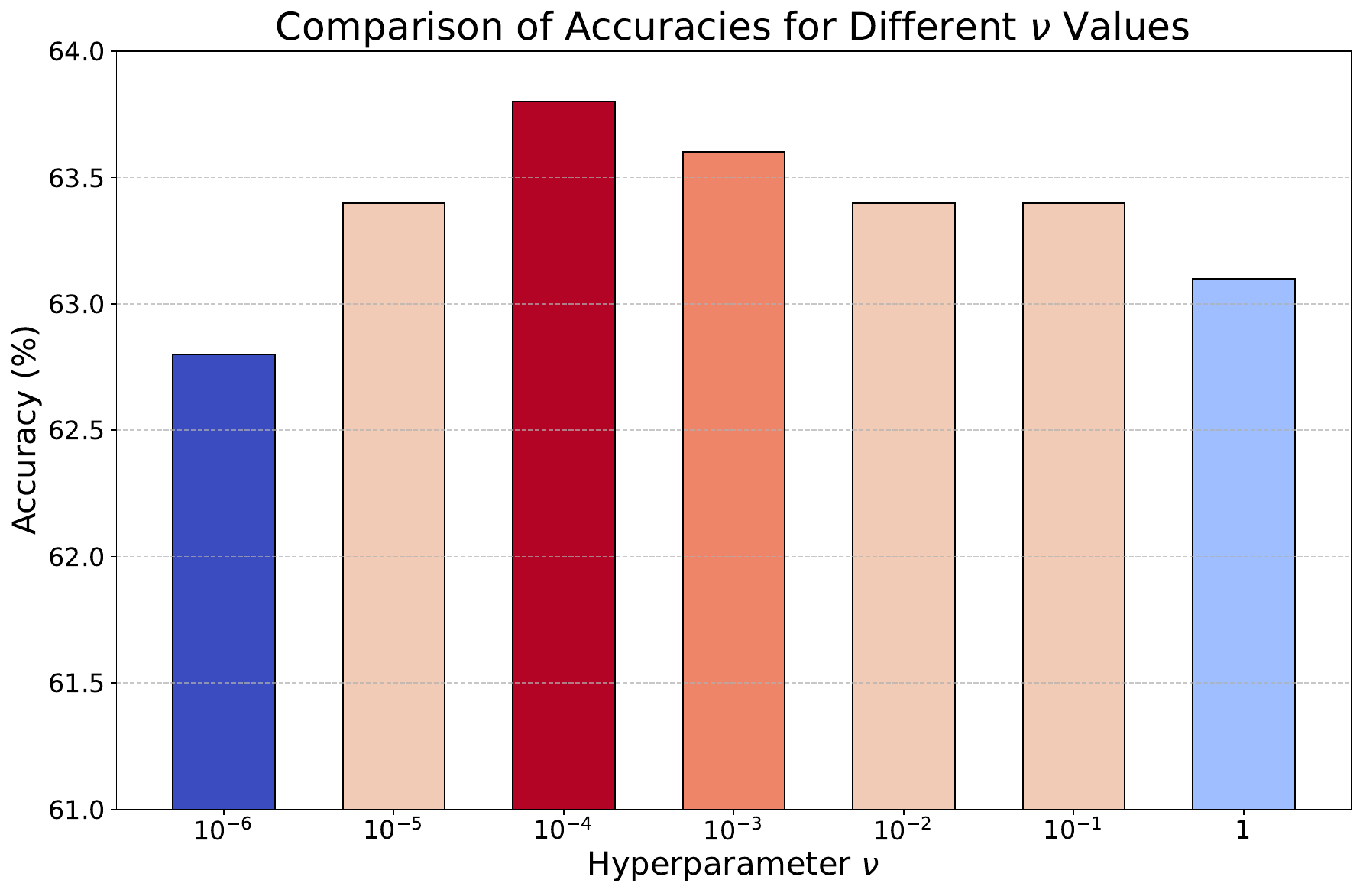}
    \caption{The comparison results on different choices of the hyperparameter $\nu$. All results are linear evaluation results with V-SimCLR + BOD-VCL on Kinetics400.}
    \label{fig:nu}
\end{figure}

\paragraph{Ablation study on $\nu$.}
As we mentioned in Subsection \ref{subsec:objective}. $\nu$ is a hyperparameter that determines the impact of the constraint in $\mathcal{L}_{pred}(\theta)$. We present the results of different choices of $\nu$ in Figure \ref{fig:nu}. All results are linear evaluation results on Kinetics400 based on V-SimCLR + BOD-VCL, with the default experimental setting in Subsection \ref{subsec:setup}, and number of positive samples $\varrho=2$.
From the results in Figure \ref{fig:nu}, we can observe that the linear evaluation performance initially increases and subsequently decreases, with the optimal choice of $\nu$ being approximately $10^{-4}$.

\paragraph{Ablation study on $\xi$.}
As we mentioned in Subsection \ref{subsec:stratify}. The hyperparameter $\xi$ is a threshold of the magnitude of the eigenvalues; eigenvalues with distance to $1+0j$ smaller than $\xi$ can be considered as related to static semantics. We present the results of different choices of $\xi$ in Table \ref{tab:xi}. All results are linear evaluation results on Kinetics400 based on V-SimCLR + BOD-VCL, with the default experimental setting in Subsection \ref{subsec:setup}, and number of positive samples $\varrho=2$.

\begin{table}[ht]
    \centering
    \caption{The comparison results on different choices of the hyperparameter $\xi$. All results are linear evaluation results with V-SimCLR + BOD-VCL on Kinetics400.}
    \label{tab:xi}
    \begin{tabular}{c|c|c|c|c|c}
    \toprule
       $\xi$ & $0.4$ & $0.3$ & $0.2$ & $0.1$ & $0.01$ \\
    \midrule
        Acc.(\%) & 62.3 & 62.6 & 63.1 &\bf 63.8 & 63.5 \\
    \bottomrule
    \end{tabular}
    
\end{table}

From the results in Table \ref{tab:xi}, we can observe that when the value of $\xi$ is closer to 0, the linear evaluation accuracy of the V-SimCLR + BOD-VCL also increases. However, when the $\xi$ is set too close to 0, the performance slightly decreases. We hypothesize that it is because in videos, most semantics, including the background, are slowly moved; therefore, to effectively separate the static semantics from the dynamic semantics, $\xi$ should not be set too close to 0.

\paragraph{Ablation study on $M$.}
As we mentioned in Subsection \ref{subsec:KMD}. $M$ is the dimension of the estimated Koopman operator $\hat{\mathcal{K}}$. Different choices of $M$ might influence how much dynamic semantics the Stratify module can obtain. We provide an ablation study on this hyperparameter in Table \ref{tab:eigen_M}. All results are linear evaluation results on Kinetics400 based on V-SimCLR + BOD-VCL with the default experimental setting in Subsection \ref{subsec:setup}.

\begin{table}[ht]
    \centering
    \caption{The comparison results on different choices of the number of eigenvectors $M$. The results are linear evaluation on Kinetics400 with V-SimCLR + BOD-VCL.}
    \label{tab:eigen_M}
    \begin{tabular}{c|c|c|c|c|c|c}
    \toprule
       $M$  & 16 & 32 & 64 & 128 & 256 & 512 \\
    \midrule
        Acc.(\%) & 62.3 & 62.5 & 62.8 &\bf 63.8 & 63.6 & 63.7 \\
    \bottomrule
    \end{tabular}
\end{table}

From the results in Table \ref{tab:eigen_M}, we can observe that as $M$ increases, the performance on Kinetic400 first increases and then stays stable afterward. This indicates that the optimal choice of $M$ should be $128$ because further increasing $M$ leads to more memory requirements.

\section{Conclusion}
In this work, we discover through experiments that V-CL methods can't effectively learn both static and dynamic features. Through theoretical analysis, we find that the undifferentiated manner of measuring of current V-CL methods may lead to sub-optimal learning of both static and dynamic features. To address this issue, we propose \textit{Bi-level Optimization with Decoupling for Video Contrastive Learning} (BOD-VCL) to learn static and dynamic in a decoupled manner. Our method can seamlessly integrate with existing V-CL approaches, demonstrating its effectiveness in capturing both static and dynamic semantics. Empirical results underscore the effectiveness of our method.

\section{{Limitation}}
{While the proposed framework demonstrates promising performance, several limitations should be acknowledged.  First, the method introduces a number of tunable hyperparameters. Although we have performed ablation studies to investigate their impact, the theoretical justification for the optimal choices remains to be fully established, and future work could aim to provide a deeper understanding of these parameters. Second, our study primarily focuses on contrastive video self-supervised learning. Extending the approach to multimodal pretraining scenarios, such as joint video–text representation learning, represents an important yet nontrivial direction for future research. We believe addressing these limitations would further enhance the generality and applicability of our framework.}

\section*{Data Availability Statement}

The benchmark datasets can be downloaded from the literature cited in Section \ref{subsec:setup}.

\section*{Conflict of Interest}
  
The authors declare no conflict of interest.

{
    \small
    \bibliographystyle{unsrt}
    \bibliography{main}
}

\clearpage
\setcounter{page}{1}
\appendix

\section*{Appendix}
The appendix provides supplementary material and additional details to support the main findings and methods. It is organized into several sections:
\begin{itemize}
    \item Section \ref{sec:pseudocode} provides the pseudo code of our proposed BOD-VCL.
    \item Section \ref{sec:notation} lists the definitions for all notations from the main text. 
    \item Section \ref{app:background_causal} provides background of causal inference.
    \item Section \ref{app:proof} provides proof for Proposition 1.
    \item Section \ref{app:proof_2} provides proof for Proposition 2.
\end{itemize}

\section{The Pseudo Code for BOD-VCL}
\label{sec:pseudocode}
The pseudo code of BOD-VCL is illustrated below. For the detailed implementation, please refer to the source code repository.\footnote{\href{https://github.com/ZeenSong/Video_contrastive}{https://github.com/ZeenSong/Video\_contrastive}}
\begin{algorithm}[htb]
\caption{Pseudo Code of the proposed BOD-VCL}
\label{alg:mamlsup}
\begin{algorithmic}[1]
{\footnotesize
\REQUIRE $f_\theta$ Networks
\STATE randomly initialize $f_\theta$.
\WHILE{not converge}
\STATE Sample batch of videos $\{X_i\}_{i=1}^N$.
\STATE Sample $\varrho$ clips $\{X_i^\rho\}_{\rho=1}^\varrho$ for each video $X_i$.
\FOR{lower level update steps}
\STATE Split each video into $X_{fore}$ and $X_{back}$.
\STATE Calculating per-frame feature of each video with $f_\theta$.
\STATE Get $\hat{\mathcal{K}}_\theta$ of each video with Equation (\ref{eq:K_target}).
\STATE Calculating $\mathcal{L}_{pred}(\theta)$ with Equation (\ref{eq:dynamic}).
\STATE Update $\theta^* \gets \theta - \gamma \nabla_{\theta} \mathcal{L}_{pred}(\theta)$.
\ENDFOR
\STATE Get $\hat{\mathcal{K}}_{\theta^*}$ of each video following steps $6\to 8$.
\STATE Get $\Phi,\Lambda$ by performing eigenvalue decomposition of for all $\hat{\mathcal{K}}_{\theta^*}$.
\STATE Initialize for each video: $\Phi_{\text{inv}}=[],\Phi_{\text{var}}=[]$.
\FOR{$m=0,1\dots ,M-1$} 
\IF{$\vert\lambda_m - 1\vert < \xi$}
\STATE Add $\varphi_m$ to $\Phi_{\text{inv}}$.
\ELSE
\STATE Add $\varphi_m$ to $\Phi_{\text{var}}$.
\ENDIF
\ENDFOR
\STATE Calculating $Z_A$ and $Z_M$ for each video sample with Equation (\ref{eq:separate}).
\STATE Calculate $\mathcal{L}_A(\theta^*(\theta))$ and $\mathcal{L}_M(\theta^*(\theta))$ with Equation (\ref{eq:decoupled_CL}).
\STATE Update $\theta \gets \theta - \gamma \nabla_{\theta}\left[ \mathcal{L}_{A}(\theta^*) + \mathcal{L}_{M}(\theta^*)\right]$.
\ENDWHILE
}
\end{algorithmic}
\end{algorithm}

\section{Table of Notations}
\label{sec:notation}
We list the definitions of all notations from the main text in Table \ref{tab:notation1} and \ref{tab:notation2}. Specifically, Table \ref{tab:notation1} contains the definition of notations from Section {3}, while Table \ref{tab:notation2} contains the definition of notations from Section {4}. 

\begin{table}[htb]
    \centering
    
    \caption{The definitions of all notations from Section {3}.}
    \resizebox{\linewidth}{!}{
    \begin{tabular}{c|c}
    \toprule
    Notations & Definition \\
    \midrule
    \multicolumn{2}{c}{Notations of Subsection {3.1}} \\
    \midrule
    $f$ & The spatiotemporal feature extractor. \\
       $\mathcal{D}=\{X_i\}_{i=1}^N$  & The unlabeled video dataset.  \\
       $N$ & The number of samples in the video dataset. \\
       $X_i=\{x_{i,t}\}_{t=1}^{T_i}$ & A video sample. \\
        $T_i$ & The total number of frames in $X_i$. \\
        $x_{i,t}\in\mathbb{R}^{H\times W \times C}$ & A single frame \\
        $H$ & The height of the frame. \\
        $W$ & The width of the frame. \\
        $C$ & The channel of the frame. \\
        $L$ & The length of the video clips. \\
        $\delta$ & The sampling stride when creating clips. \\
        $\mathcal{B}=\{X_i\}_{i=1}^{N_{bs}}$ & A mini-batch of video data. \\
        $N_{bs}$ & Number of videos in a mini-batch. \\
        $\varrho$ & The number of data augmentations. \\
        $\rho_{i,j}$ & the $j$-th augmentation of $X_i$. \\
        $f_p$ & a projection head. \\
        $Z_{i,j}=f_p(f(\rho_{i,j}(X_i)))$ & the embedding features. \\
        $\mathcal{Z}^-_{i}$ & The negative set of $X_{i}$. \\
        $\mathcal{Z}^+_{i,j}$ & The positive set of $Z_{i,j}$. \\
        $\ell_{i,j}^\rho$ & The objective for clip $\rho_j(X_i)$. \\
        $\alpha$ & The temperature hyper-parameter.\\
        $f_{\theta_m}$ & The target network. \\
        $\theta_m$ & The parameters of the target network. \\
        $\gamma$ & A hyper-parameter. \\
        $f_{pre}$ & The prediction head of V-BYOL. \\
        $\mathcal{D}_{train}$ & The training set for downstream evaluation. \\
        $N_{train}$ & Number of downstream training samples. \\
        $\mathcal{D}_{test}$ & The test set for downstream evaluation. \\
        $N_{test}$ & Number of downstream test samples. \\
        $C$ & Number of labels. \\
        $W$, $b$ & Weights and bias for linear classifier. \\
        
    \midrule
    \multicolumn{2}{c}{Notations of Section {3.3}} \\
    \midrule
    $\mathbf{X}$ & The video sample. \\
    $\mathbf{A}$ & The static semantics of a video. \\
    $\mathbf{M}$ & The dynamic semantics of a video. \\
    $\mathbf{S}$ & The similarity score. \\
    $g_A,g_M$ & Sub-networks for static and dynamic semantics. \\
    $\theta_A,\theta_M$ & The parameters of $g_A,g_M$. \\
    $\Delta(\theta_A,\theta_M)$ & The gradient norm.\\ 
    \bottomrule
    \end{tabular}
    }
    \label{tab:notation1}
\end{table}

\begin{table}[htb]
    \centering
    
    \caption{The definitions of all notations from Section {4}.}
    \resizebox{\linewidth}{!}{
    \begin{tabular}{c|c}
    \toprule
    Notations & Definition \\
    $x_t$ & The frame at time step $t$. \\
    $X$ & A video. \\
    $\mathcal{X}$ & The space of video frames. \\
    $F(x_t)$ & The transition function. \\
    $f^*$ & An ideal feature extractor. \\
    $\boldsymbol{z}_t$ & The semantic vector. \\
    $\boldsymbol{z}_{t}^{\text{inv}},\boldsymbol{z}_{t}^{\text{var}}$ & Invariant and variant components of of $\boldsymbol{z}_t$. \\
    $\mathcal{U}$ & A infinite-dimensional space. \\
    $\mathcal{G}:\mathcal{X} \rightarrow \mathcal{U}$ & A nonlinear mapping from $\mathcal{X}$ to $\mathcal{U}$. \\
    $\mathcal{K}$ & The Koopman operator. \\
    $f_\theta$ & The feature extractor. \\
    $X_{fore}$ & The first $T-1$ frames from $X$. \\
    $X_{back}$ & The last $T-1$ frames from $X$. \\
    $F_b$ & The representation of the first $T-1$ frames of $X$. \\
    $F_f$ & The representation of subsequent $T-1$ frames $X$. \\
    $\hat{\mathcal{K}}_\theta$ & The estimated Koopman operator. \\
    $\nu$ & The Lagrange multiplier. \\
    $\Phi=[\varphi_1,\dots,\varphi_M] \in \mathbb{C}^{M\times M}$ & The eigenvectors of $\hat{\mathcal{K}}_\theta$.\\
    $\Lambda=\mathop{diag}([\lambda_1,\dots,\lambda_M]) \in \mathbb{C}^{M\times M}$ & The diagonal matrix of eigenvalues of $\hat{\mathcal{K}}_\theta$ \\
    $\mathbb{C}$ & The set of all complex numbers. \\
    $M$ & The dimension of $\mathcal{K}$. \\
    $Re$ & The real part of a complex number\\
    $Im$ & The imaginary part of a complex number \\
    $\xi$ & An hyperparameter \\
    $\Lambda_{\text{inv}}$ & Eigenvalues with distance to $1+0j$ smaller than $\xi$. \\
    $\Lambda_{\text{var}}$ & Eigenvalues with distance to $1+0j$ larger than $\xi$. \\
    $Z_A$ & The static semantics of a video. \\
    $Z_M$ & The dynamic semantics of a video. \\
    $\gamma$ & The learning rate. \\ 
        \bottomrule
    \end{tabular}
    }
    \label{tab:notation2}
\end{table}

\section{Background in Causal Inference.}
\label{app:background_causal}
Causal effects are typically described as how the output variable changes when the input variable is altered while excluding the influence of other variables \cite{rosenbaumCentralRolePropensity1983, rubinCausalInferenceUsing2005, pearl2009causality, glymour_causal_2016}. Randomized controlled trials (RCTs) are considered the gold standard for studying causal effects because, in a well-designed RCT, all factors that influence the outcome variable are either held constant or vary randomly, except for the input variable. Consequently, any observed change in the outcome variable can be attributed solely to the input variable \cite{fisher1966design, glymour_causal_2016}.

Causal inference aims to estimate causal effects by analyzing the probability distributions of observed data \cite{pearl2009causality, glymour_causal_2016}. In the realm of causal inference, the Structural Causal Model (SCM) is extensively utilized to articulate the causal relationships among variables. An SCM is formally defined as follows:
\begin{definition}
    (Structural Causal Model \cite{peters2017elements}) A structural causal model $\mathcal{C} := (S, P_\epsilon)$ comprises:
    \begin{enumerate}
        \item A set $S$ of $d$ structural assignments:
        \begin{equation}
            X_j := f_j(\mathrm{PA}_j, \epsilon_j), \quad j = 1, \dots, d,
        \end{equation}
        where $\mathrm{PA}_j \subseteq \{X_1, \dots, X_d\} \setminus \{X_j\}$ denotes the parents of $X_j$, and $f_j$ is the deterministic function that generates the value of $X_j$ based on its parents and an exogenous noise term $\epsilon_j$.
        
        \item A joint distribution $P_\epsilon = P(\epsilon_1) \cdots P(\epsilon_d)$ over the independent exogenous noise variables $\epsilon_j$.
    \end{enumerate}
\end{definition}

An SCM can be represented graphically using a Directed Acyclic Graph (DAG) \cite{glymour_causal_2016}. In this DAG $\mathcal{G}$, each variable $X_j$ is depicted as a vertex, and directed edges are drawn from each parent in $\mathrm{PA}_j$ to $X_j$. The SCM also specifies the observational distribution $P(\mathbf{X})$ over the set of variables $\mathbf{X} = \{X_1, \dots, X_d\}$. The joint distribution of all variables in the SCM is given by the product of their conditional distributions based on the DAG structure:
\begin{equation}
    P(X_1, \dots, X_d) = \prod_{j=1}^d P(X_j \mid \mathrm{PA}_j).
\end{equation}
When a structural assignment is replaced with a deterministic value, such as $X_j := x$, this operation is known as a $do$-intervention \cite{pearl2009causality}. Intervention actively modifies the joint distribution by setting a variable to a specific value, overriding the structural equation that would have otherwise determined that variable’s value. Consequently, performing a $do$-intervention on $X_j$ leads to a post-intervention distribution denoted by $P(X_1, \dots, X_d \mid do(X_j = x))$, which can be viewed as the resulting distribution after an RCT.  Here, $do(\cdot)$ signifies the intervention operation. The post-intervention distribution can be expressed as:
\begin{equation}
\begin{split}
    &P(X_1, \dots, X_d \mid do(X_j = x))\\
     &= 
    \begin{cases}
        \displaystyle \prod_{k \neq j} P(X_k \mid \mathrm{PA}_k) & \text{if } X_j = x, \\
        0 & \text{otherwise}.
    \end{cases}
\end{split}
\end{equation}
This formulation ensures that $X_j$ is held constant at $x$, while the distribution of the other variables adjusts accordingly based on their conditional dependencies.

\begin{definition}
    \textnormal{(Path)}. A path consists of three components including the Chain Structure: $A\rightarrow B \rightarrow C$, the Collider Structure: $A\rightarrow B \leftarrow C$ and the Fork Structure: $A\leftarrow B \rightarrow C$.
\end{definition}
\begin{definition}
    \textnormal{($d$-separation)}. A path $p$ is blocked by a set of Nodes $Z$ if and only if:
    \begin{itemize}
        \item $p$ contains a chain of nodes $A\rightarrow B \rightarrow C$ or a fork $A\leftarrow B \rightarrow C$ such that the middle node $B$ is in $Z$ (i.i., $B$ is conditioned on), or
        \item $p$ contains a collider $A\rightarrow B \leftarrow C$ such that the collision node $B$ is not in $Z$, and no descendant of $B$ is in $Z$. 
    \end{itemize}
    If $Z$ blocks every path between two nodes $X$ and $Y$, then $X$ and $Y$ are $d$-separated.
\end{definition}

\section{Proof for Proposition 1}
\label{app:proof}

\begin{proposition}[Incomplete Semantic Learning with Unified Loss]
Assume that for each \(j \in \{1, \dots, k\}\), the pair \((a^{(j)}, m^{(j)})\) is confoundedly correlated, meaning that \(a^{(j)}\) and \(m^{(j)}\) provide redundant information about the similarity structure \(S\). Suppose further that, during training (i.e., along the gradient descent iterations), for each \(j\), one component of the pair \((a^{(j)}, m^{(j)})\) is consistently easier to learn than its counterpart, as defined in Definition~\ref{def:1}. Specifically, for each \(j\), either \(\Delta_{a^{(j)}}(\theta_A, \theta_M) > \Delta_{m^{(j)}}(\theta_A, \theta_M)\) or \(\Delta_{m^{(j)}}(\theta_A, \theta_M) > \Delta_{a^{(j)}}(\theta_A, \theta_M)\) holds consistently throughout training.

Then, when the loss function \(\mathcal{L}\) measures similarity using both \(A\) and \(M\) via the representation \(Z = g_A(X; \theta_A) + g_M(X; \theta_M)\), the learned representation converges as \(t \to \infty\) to a solution where, for each \(j\), the contribution from the harder semantic component vanishes. That is, for each \(j\), either
\begin{equation}
    \begin{split}
        &\lim_{t \to \infty} g_M^{(j)}(X; \theta_M^{(t)}) \approx g_M^{(j)}(X; \theta_M^{(0)}) \quad \text{or} \\
        &\lim_{t \to \infty} g_A^{(j)}(X; \theta_A^{(t)}) \approx g_A^{(j)}(X; \theta_A^{(0)}),
    \end{split}
\end{equation}
depending on which component is harder to learn. Consequently, rather than jointly encoding both \(a^{(j)}\) and \(m^{(j)}\) for each semantic dimension \(j\), the final representation effectively encodes only the easier semantic component.
\end{proposition}

\begin{proof}
We prove the proposition by showing that redundancy between \(a^{(j)}\) and \(m^{(j)}\), combined with the consistent ease of learning for one component, causes the gradient with respect to the harder component to vanish, leading to its contribution remaining near the initial value.

\textbf{Step 1. Redundancy and Optimal Representation.} \\
Since \(a^{(j)}\) and \(m^{(j)}\) are confoundedly correlated, they provide redundant information about the similarity structure \(S\). Formally, for each \(j\),
\[
I(a^{(j)}; S) \approx I(a^{(j)}, m^{(j)}; S),
\]
implying that \(a^{(j)}\) alone captures nearly all the mutual information that \(a^{(j)}\) and \(m^{(j)}\) jointly have with \(S\). Therefore, there exists an optimal representation \(Z^*\) that minimizes the loss \(\mathcal{L}\) such that \(Z^{*(j)}\) can be achieved by encoding only one of the components, say \(a^{(j)}\), without needing \(m^{(j)}\). In other words, there exists a minimizer where \(g_M^{(j)}(X; \theta_M^*) = 0\) if \(a^{(j)}\) is easier to learn, and similarly for \(g_A^{(j)}\) if \(m^{(j)}\) is easier.

To formalize this, consider the true optimal representation \(Z^* = \arg\min_Z \mathcal{L}(Z)\). Due to the redundancy, we can decompose \(Z^*\) as \(Z^* = Z_A^* + Z_M^*\), where \(Z_A^*\) encodes the information from \(A\) and \(Z_M^*\) from \(M\). However, because of the confounded correlation, there exists a minimizer where \(Z_M^{*(j)} = 0\) for each \(j\) if \(a^{(j)}\) suffices to capture the necessary information, and vice versa.

\textbf{Step 2. Gradient Descent Dynamics and Ease of Learning.} \\
Let \((\theta_A^{(t)}, \theta_M^{(t)})\) denote the parameters at iteration \(t\) of gradient descent. By the chain rule, the gradients are
\[
\nabla_{\theta_{a^{(j)}}}\mathcal{L} = \frac{\partial \mathcal{L}}{\partial Z} \cdot \frac{\partial g_A(X)}{\partial \theta_{a^{(j)}}}, \quad \nabla_{\theta_{m^{(j)}}}\mathcal{L} = \frac{\partial \mathcal{L}}{\partial Z} \cdot \frac{\partial g_M(X)}{\partial \theta_{m^{(j)}}}.
\]
The ease of learning is defined by the gradient norms:
\[
\Delta_{a^{(j)}} = \left\| \nabla_{\theta_{a^{(j)}}}\mathcal{L} \right\|, \quad \Delta_{m^{(j)}} = \left\| \nabla_{\theta_{m^{(j)}}}\mathcal{L} \right\|.
\]
Assume that for each \(j\), \(\Delta_{a^{(j)}} > \Delta_{m^{(j)}}\) consistently throughout training, meaning \(a^{(j)}\) is easier to learn than \(m^{(j)}\). Consequently, the updates to \(\theta_{a^{(j)}}\) are larger than those to \(\theta_{m^{(j)}}\), causing \(g_A^{(j)}(X; \theta_A^{(t)})\) to adapt more quickly towards capturing the necessary information in \(Z^*\).

As \(g_A^{(j)}\) approaches \(Z^{*(j)}\), the loss \(\mathcal{L}\) decreases, and the gradient \(\frac{\partial \mathcal{L}}{\partial Z}\) diminishes. Since \(\nabla_{\theta_{m^{(j)}}}\mathcal{L} = \frac{\partial \mathcal{L}}{\partial Z} \cdot \frac{\partial g_M}{\partial \theta_{m^{(j)}}}\), the gradient with respect to \(\theta_{m^{(j)}}\) becomes smaller as \(\frac{\partial \mathcal{L}}{\partial Z} \to 0\), even if \(\frac{\partial g_M}{\partial \theta_{m^{(j)}}}\) remains non-zero. Thus, the updates to \(\theta_{m^{(j)}}\) become negligible:
\[
\theta_{m^{(j)}}^{(t+1)} \approx \theta_{m^{(j)}}^{(t)},
\]
implying that \(g_M^{(j)}(X; \theta_M^{(t)})\) remains close to its initial value, say \(g_M^{(j)}(X; \theta_M^{(0)}) = 0\), assuming standard initialization.

\textbf{Step 3. Convergence to an Incomplete Representation.} \\
Under standard assumptions of smoothness and convergence of gradient descent, the parameters \((\theta_A^{(t)}, \theta_M^{(t)})\) converge to a local minimizer \((\theta_A^*, \theta_M^*)\) as \(t \to \infty\). Given the dynamics above, for each \(j\), if \(a^{(j)}\) is easier to learn, then
\[
\lim_{t \to \infty} g_M^{(j)}(X; \theta_M^{(t)}) \approx g_M^{(j)}(X; \theta_M^{(0)}),
\]
meaning the contribution from the harder component \(m^{(j)}\) vanishes in the final representation. Similarly, if \(m^{(j)}\) is easier, then \(g_A^{(j)}\) remains near its initial value. Therefore, the unified loss leads the network to encode only the easier semantic component for each \(j\), failing to jointly encode both \(a^{(j)}\) and \(m^{(j)}\).
\end{proof}

\section{Proof for Corollary 1}
\label{app:proof_2}
\begin{corollary}[Complete Representation with Decoupled Losses]
Under the same structural assumptions as in Proposition~\ref{prop:1}, suppose that the loss function is decoupled so that similarity is measured independently on the static and dynamic features. The overall loss as
\begin{equation}
    \mathcal{L}_{\text{sep}}(\theta_A,\theta_M) = \mathcal{L}_A + \mathcal{L}_M,
\end{equation}
where \(\mathcal{L}_A\) and \(\mathcal{L}_M\) are arbitrary continuously differentiable functions that measure similarity solely on the static and dynamic representations, respectively. Then, gradient descent will optimize $\theta_A$ and $\theta_M$ independently. Consequently, for each $j$, both
\[
\Delta_{a^{(j)}}^{(A)} = \Bigl\|\nabla_{\theta_{a^{(j)}}}\mathcal{L}_A\Bigr\| \quad \text{and} \quad \Delta_{m^{(j)}}^{(M)} = \Bigl\|\nabla_{\theta_{m^{(j)}}}\mathcal{L}_M\Bigr\|
\]
will be preserved without interference from each other. Consequently, the final representation effectively encodes both the easier semantic and the hard semantic.
\end{corollary}

\begin{proof}
We begin by defining the decoupled loss functions:
\[
\mathcal{L}_A = \text{Loss}\bigl(g_A(A;\theta_A)\bigr) \quad \text{and} \quad \mathcal{L}_M = \text{Loss}\bigl(g_M(M;\theta_M)\bigr),
\]
so that the overall loss is 
\[
\mathcal{L}_{\text{sep}} = \mathcal{L}_A + \mathcal{L}_M.
\]

\textbf{Step 1. Independence of Gradients.} \\
By construction, $\mathcal{L}_A$ depends solely on the static features $g_A(A;\theta_A)$, and is independent of the dynamic features (and hence of $\theta_M$). Similarly, $\mathcal{L}_M$ depends solely on $g_M(M;\theta_M)$ and is independent of $\theta_A$. Therefore, the gradients satisfy
\[
\nabla_{\theta_A}\mathcal{L}_{\text{sep}} = \nabla_{\theta_A}\mathcal{L}_A \quad \text{and} \quad \nabla_{\theta_M}\mathcal{L}_{\text{sep}} = \nabla_{\theta_M}\mathcal{L}_M.
\]
This implies that the parameter updates for $\theta_A$ and $\theta_M$ are decoupled:
\[
\theta_A^{(t+1)} = \theta_A^{(t)} - \eta\,\nabla_{\theta_A}\mathcal{L}_A(\theta_A^{(t)}),
\]
\[
\theta_M^{(t+1)} = \theta_M^{(t)} - \eta\,\nabla_{\theta_M}\mathcal{L}_M(\theta_M^{(t)}).
\]

\textbf{Step 2. Preservation of Individual Gradient Signals.} \\
Since the static and dynamic losses are optimized independently, the gradient signal for each semantic component is determined solely by its own loss. That is, for each semantic dimension \(j\), we have
\[
\Delta_{a^{(j)}}^{(A)} = \left\|\nabla_{\theta_{a^{(j)}}}\mathcal{L}_A\right\| \quad \text{and} \quad \Delta_{m^{(j)}}^{(M)} = \left\|\nabla_{\theta_{m^{(j)}}}\mathcal{L}_M\right\|.
\]
Because there is no cross-coupling, a large gradient in one branch does not suppress the gradient in the other. Consequently, both $\Delta_{a^{(j)}}^{(A)}$ and $\Delta_{m^{(j)}}^{(M)}$ are maintained and will drive their respective parameters to minimize $\mathcal{L}_A$ and $\mathcal{L}_M$ independently.

\textbf{Step 3. Joint Encoding of All Semantics.} \\
Since both losses are minimized independently, the optimal parameters $(\theta_A^*,\theta_M^*)$ satisfy
\[
\theta_A^* = \arg\min_{\theta_A} \mathcal{L}_A \quad \text{and} \quad \theta_M^* = \arg\min_{\theta_M} \mathcal{L}_M.
\]
Thus, the final feature is
\[
Z = g_A(A;\theta_A^*) + g_M(M;\theta_M^*).
\]
Because both $g_A(A;\theta_A^*)$ and $g_M(M;\theta_M^*)$ are independently optimized to encode their respective semantic modalities, the final representation $Z$ jointly encodes the full set of semantic components,
\[
\{a^{(1)},\dots,a^{(k)},\, m^{(1)},\dots, m^{(k)}\}.
\]

This completes the proof.
\end{proof}

\section{{Discussion of Information Loss}}

{\begin{definition}[Information loss]
Given an encoder $f_\theta$ and the decomposition defined by Equation~(\ref{eq:separate}), let
\begin{equation}
\begin{split}
    Z(X) \triangleq \frac{1}{T}\sum_{t=1}^T f_\theta(x_t),\\
Z_A(X) \triangleq \frac{1}{T}\sum_{t=1}^T \Phi \Lambda_{\text{inv}} \Phi^{-1} f_\theta(x_t), \\
Z_M(X) \triangleq \frac{1}{T}\sum_{t=1}^T \Phi \Lambda_{\text{var}} \Phi^{-1} f_\theta(x_t).
\end{split}
\end{equation}
We say that information loss occurs if there exist two different videos $X\neq X'$ such that
\begin{equation}
    \big(Z_A(X), Z_M(X)\big) = \big(Z_A(X'), Z_M(X')\big).
\end{equation}
\end{definition}}

{\begin{proposition}[No information loss under orthogonality]
Assume Equation (\ref{eq:separate}) with $\hat{\mathcal K}_\theta = \Phi \Lambda \Phi^{-1}$ and $\Lambda=\Lambda_{\text{inv}}+\Lambda_{\text{var}}$. Define the linear operators
\begin{equation}
    T_A \triangleq \Phi \Lambda_{\text{inv}} \Phi^{-1}, \qquad
T_M \triangleq \Phi \Lambda_{\text{var}} \Phi^{-1}.
\end{equation}
If the orthogonality regularizer in Equation~(\ref{eq:dynamic}) enforces $\hat{\mathcal K}_\theta^\top \hat{\mathcal K}_\theta \approx I$, ensuring that $\hat{\mathcal K}_\theta$ is full rank and the decomposition is nondegenerate, then
\begin{equation}
    \begin{split}
        Z_A(X) &= T_A Z(X),\ Z_M(X) = T_M Z(X), \\  Z(X) &= Z_A(X) + Z_M(X).
    \end{split}
\end{equation}
Consequently, the mapping $Z(X)\mapsto\big(Z_A(X),Z_M(X)\big)$ is bijective. In particular, if $Z(X)\neq Z(X')$ then $\big(Z_A(X),Z_M(X)\big)\neq\big(Z_A(X'),Z_M(X')\big)$, hence no information loss occurs.
\begin{proof}
By linearity,
\begin{equation}
    \begin{split}
        Z_A(X)&=\frac{1}{T}\sum_{t=1}^T T_A f_\theta(x_t)\\ &= T_A \Big(\frac{1}{T}\sum_{t=1}^T f_\theta(x_t)\Big)\\
        &=T_A Z(X),\\
Z_M(X)&=T_M Z(X).
    \end{split}
\end{equation}
Since $\Lambda=\Lambda_{\text{inv}}+\Lambda_{\text{var}}$, we have
\begin{equation}
    T_A+T_M=\Phi(\Lambda_{\text{inv}}+\Lambda_{\text{var}})\Phi^{-1}=I,
\end{equation}
which implies $Z(X)=Z_A(X)+Z_M(X)$. Therefore, the mapping $Z(X)\mapsto (Z_A(X),Z_M(X))$ is injective. Moreover, given both $Z_A(X)$ and $Z_M(X)$, one can directly reconstruct $Z(X)$ by summing them, i.e., $Z(X)=Z_A(X)+Z_M(X)$. This shows that the correspondence between $Z(X)$ and $(Z_A(X),Z_M(X))$ is bijective. The orthogonality constraint $\hat{\mathcal K}_\theta^\top \hat{\mathcal K}_\theta \approx I$ ensures that $\hat{\mathcal K}_\theta$ is full rank and the decomposition is well conditioned, which prevents degenerate cases that could otherwise collapse distinct features.    
\end{proof}
\end{proposition}}

\end{document}